\documentclass[]{fairmeta}

\title{Unified Vision–Language Modeling\\ via Concept Space Alignment}

\author[1,2,*]{Yifu Qiu}
\author[2]{Paul-Ambroise Duquenne}
\author[2]{Holger Schwenk}

\affiliation[1]{University of Edinburgh}
\affiliation[2]{FAIR at Meta}

\contribution[*]{This work was done during an internship of Yifu Qiu at FAIR at Meta}

\usepackage{amsmath,amsfonts,bm}

\def\eqref#1{equation~\ref{#1}}

\def\1{\bm{1}}

\DeclareMathAlphabet{\mathsfit}{\encodingdefault}{\sfdefault}{m}{sl}
\SetMathAlphabet{\mathsfit}{bold}{\encodingdefault}{\sfdefault}{bx}{n}

\usepackage{hyperref}
\usepackage{url}
\usepackage{natbib}
\usepackage{comment,xspace,color}

\usepackage[utf8]{inputenc} %
\usepackage[T1]{fontenc}    %
\usepackage{hyperref}       %
\usepackage{url}            %
\usepackage{booktabs}       %
\usepackage{amsfonts}       %
\usepackage{nicefrac}       %
\usepackage{microtype}      %
\usepackage{multirow}   %
\usepackage{booktabs}   %
\usepackage{graphicx}   %
\usepackage{siunitx} %
\usepackage{tcolorbox}
\usepackage{array}
\usepackage{rotating}  %
\usepackage{adjustbox}
\usepackage{booktabs}
\usepackage{siunitx}
\usepackage{wrapfig}
\usepackage{graphicx}
\usepackage{subcaption}
\usepackage{tabularx}
\usepackage{colortbl}

\usepackage{makecell}

\definecolor{deepmindblue}{HTML}{e6f0ff} %
\definecolor{badred}{HTML}{d32f2f}       %
\definecolor{goodgreen}{HTML}{388e3c}    %

\usepackage{makecell}

\newcommand{\sonar}{\textsc{Sonar}\xspace}
\newcommand{\sonartwo}{\textsc{OmniSONAR}\xspace}
\newcommand{\PE}{\textsc{Perception~Encoder}\xspace}
\newcommand{\vsonar}{\textsc{\mbox{v-Sonar}}\xspace}
\newcommand{\lcm}{\textsc{LCM}\xspace}
\newcommand{\PEvideo}{\textsc{PE-Video}\xspace}
\newcommand{\vatex}{\textsc{Vatex}\xspace}
\newcommand{\dream}{\textsc{Dream-1k}\xspace}
\newcommand{\videoxum}{\textsc{VideoXum}\xspace}

\newcommand{\vlcm}{\textsc{v-LCM}\xspace}
\newcommand{\bleu}{\textsc{Bleu}\xspace}
\newcommand{\rouge}{\textsc{Rouge}\xspace}

\abstract{ We introduce \vsonar, a vision–language embedding space extended from the text-only embedding space \sonar %
\citep{sonar2big}, which supports 1500 text languages and 177 speech languages.
To construct \vsonar, we propose a post-hoc alignment pipeline that maps the representations of an existing vision encoder into the \sonar space.
We thoroughly evaluate \vsonar and show that its embeddings achieve competitive performance on text-to-video retrieval.
Equipped with the \sonartwo text decoder, \vsonar further surpasses state-of-the-art vision–language models on video captioning tasks, including \dream (\bleu 23.9 vs. 19.6) and \PEvideo (\bleu 39.0 vs. 30.0).

Leveraging \vsonar, we first demonstrate that the Large Concept Model (\lcm; \citealt{barrault2024largeConceptModelLCM}) operating in \sonar and trained with English text only, can perform both single- and multi-visual concept understanding in a zero-shot manner.
Finally, we introduce \vlcm, which extends the \lcm with vision–language instruction tuning. \vlcm encodes vision and language inputs into an unified sequence of latent embeddings via \vsonar and \sonar, and it is trained with the same latent diffusion objective for next-embedding prediction as in \lcm's text-only pre-training.
Experiments on a large-scale multilingual and -modal instruction–tuning data mixture highlight the potential of \vlcm: \vlcm matches state-of-the-art vision-language models on tasks covering image/video captioning and question answering, while significantly outperforming them across 61 rich- to low-resource languages out of all 62 tested languages.
 }

\date{\today}
\correspondence{Holger Schwenk at \email{schwenk@meta.com}}

\begin{document}

\maketitle

\section{Introduction}
Language- and modality-agnostic embedding spaces have emerged as a powerful paradigm for multilingual and --modal representation learning  \citep{Artetxe:2019:tacl_massive_ml,labse,sentenceT5:2021:arxiv,duquenne2023sonar,multlinE5:2024:arxiv,M3embed:2024:arxiv}.
Such spaces have achieved state-of-the-art performance across a wide range of applications, e.g., bitext mining \citep{schwenk:2019:arxiv_ccmatrix,nllb2022,ramesh-etal-2022-samanantar}, and speech–text mining \citep{NEURIPS2021_8466f9ac-sonar-speech-mining}. 
Beyond these, embedding spaces with the encoder–decoder architecture such as SONAR \citep{duquenne2023sonar} have further enabled generative modeling directly in the latent embedding space. The Large Concept Model (\lcm; \citealt{barrault2024largeConceptModelLCM}) extends this direction by showing that diffusion-based language modeling can operate directly in the language-agnostic latent space, i.e., over continuous embeddings rather than discrete tokens. Despite these advances, existing embedding spaces remain restricted to text and speech, limiting their potential for vision–language tasks.

In this work, we introduce \vsonar, which extends \sonartwo \citep{sonar2big} to the image and video modality. To the best of our knowledge, this makes \sonartwo the most universal embedding space covering four modalities\footnote{\sonartwo supports text in 1.5k languages, speech in 177 languages and the added image and video modalities.} and up to 1500 languages.
We use teacher-student training \citep{Reimers,Duquenne:2021:neurips,laser3} to align the representations of a state-of-the-art vision encoder, \PE \citep{bolya2025perceptionencoder}, with \sonar{}’s semantic space in a post-hoc manner. The alignment follows a coarse-to-fine curriculum, over three stages of vision captioning data: (1) large-scale image–caption pairs (12M) for coarse grounding, (2) synthetic video–caption pairs (2M) for temporal adaptation, and (3) high-quality human-annotated video captions (200K) for fine-grained alignment.
We evaluate \vsonar extensively. On zero-shot video retrieval, it achieves Recall@1 of 73.03 on \PEvideo, largely surpassing SigLIP2-g-opt (63.91). On zero-shot video captioning, it outperforms state-of-the-art vision–language models, improving \bleu by +18, +4.3 on \PEvideo, \dream, respectively, over the Perception Language Model \citep{cho2025perceptionlm}.

By aligning \vsonar to \sonar, we show that the latent diffusion language model operating in \sonar, \lcm \citep{barrault2024largeConceptModelLCM} trained with English textual corpus, can zero-shot process the visual embeddings encoded by \vsonar.
In the single-concept understanding task, i.e., video captioning, \lcm only lags behind the existing VLMs with limited margins across \PEvideo, \dream, and \vatex. Similarly, \lcm remains competitive for multi-concept reasoning task, i.e., long video summarization as evaluated on \videoxum. 

From the view of vision-language modeling, \lcm serves as a new paradigm which unifies vision and language modality to the modality-agnostic latent space shared by \sonar and \vsonar, and directly predict the next embedding with the latent diffusion objective. 
Therefore, we further introduce a vision-language instruction fine-tuned \lcm as an exploration to maximize its utility in various downstream vision-language tasks, named \vlcm.
\vlcm encodes multimodal data (images, videos, and text) with \vsonar and \sonar, and it is trained with the same latent diffusion strategy, following the original two-tower \lcm framework \citep{barrault2024largeConceptModelLCM} in its textual pre-training.

We evaluate \vlcm on the multilingual and -modal instruction-tuning dataset, M3IT \citep{li2023m3it}, which spans 8 task categories, supports both image and video modalities, and covers 80 languages. \vlcm achieves competitive performance with other vision-language models such as InternVL \citep{chen2024internvl}, Qwen-VL \citep{wang2024qwen2vl,bai2025qwen2_5_vl} and Perception LM on image/video captioning, visual question answering, and other generation tasks. Notably, in M3IT's multilingual evaluation, \vlcm outperforms other VLMs in 61 languages out of 62 tested languages, ranging from high-resource to low-resource setting.
The contributions of this work are:
\begin{itemize}
    \item We introduce \vsonar, the first extension of a language- and modality-agnostic embedding space (\sonar) to image and video, via a post-hoc coarse-to-fine alignment strategy.
    \item We demonstrate that \vsonar achieves state-of-the-art zero-shot performance on video retrieval and captioning, and generalizes robustly to multilingual settings.
    \item We show that the \lcm, originally trained on text-only data, can effectively operate on \vsonar embeddings for zero-shot single- and multi-concept vision understanding tasks.
    \item We extend LCM into a latent diffusion vision–language model (\vlcm) by unifying vision and language in the shared latent space of \vsonar and \sonar. On M3IT, \vlcm matches state-of-the-art VLMs in captioning and question answering while outperforming them in 61 non-English languages.
\end{itemize}

\section{\vsonar}
\label{sec:vsonar}

We begin by introducing \vsonar, a vision–language embedding space constructed by post-hoc aligning a state-of-the-art vision encoder, \PE, with the multilingual textual embedding space \sonar. We select the \PE as the base encoder for two key reasons: (1) it achieves state-of-the-art performance across both image and video modalities \citep{bolya2025perceptionencoder}, and (2) it has been pre-trained in conjunction with a lightweight text encoder, which facilitates much easier post-hoc alignment with \sonar. This design choice distinguishes \PE from alternative vision encoders such as v-JEPA \citep{bardes2023vJEPA,assran2025vJEPA2} and DINO \citep{oquab2023dinov2,simeoni2025dinov3}, which primarily prioritize visual feature learning without explicit consideration of textual alignment.

\begin{figure}
    \centering
    \includegraphics[width=0.8\linewidth]{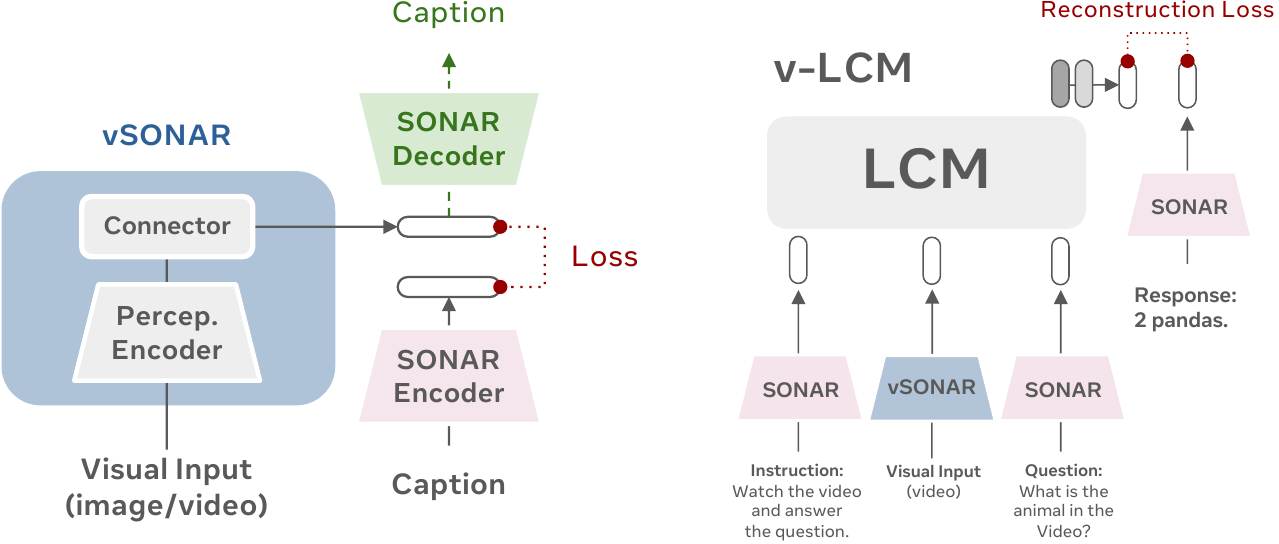}
    \caption{Left: Illustration of \vsonar. Right:  training \vlcm with vision-language instruction tuning.}
    \label{fig:vsonar-architecture}
\end{figure}

\paragraph{Architecture}
The architecture of \vsonar is illustrated in the left panel of \autoref{fig:vsonar-architecture}.
Given the input image or video, \PE (PE) will first encode each frame separately. Then, we stack a lightweight projector on top of PE to adapt the encoder’s representations into the \sonar space.
The projector first injects positional embeddings to the embeddings of all frames, thus encoding temporal order information, followed by a single temporal attention layer that enables frame-level interactions.
Finally, an attention layer then aggregates the frame embeddings into a single video-level representation, which serves as the final embedding for downstream tasks. See \autoref{appendix:implementation-details} for implementation details.

\paragraph{Alignment from Vision to Language.}

We use the captioning data for the \PE to align with \sonar, with assumption that the visual inputs and caption should have the same semantic meaning, thus the high-level representations should be as close as possible in the latent modality-agnostic space.

Therefore, given a set of $N$ paired visual inputs and captions $\mathcal{D} = \{(V_i, T_i)\}_{i=1}^N$, where $V_i$ is an image or video and $T_i$ is its corresponding caption, we seek to learn a mapping such that the visual embedding $\mathbf{z}_v = f_\theta(V_i)$ and the textual embedding $\mathbf{z}_t = g(T_i)$ share the same semantic space, where $f_\theta$ denotes the trainable vision encoder and $g$ is the frozen \sonar text encoder. To enforce semantic alignment, we minimize the discrepancy between visual and textual embeddings in the \sonar space using Mean Squared Error (MSE) loss:
\begin{equation}
\mathcal{L}_{\text{align}} = \frac{1}{N} \sum_{i=1}^N \| f_\theta(V_i) - g(T_i) \|_2^2.
\end{equation}
Following the teacher-student training \citep{Reimers,duquenne2023sonar}, \sonar is frozen and we only update the parameters in the lightweight projector, and the vision encoder. We also experimented with an additional contrastive loss \citep{oord2018representation,radford2021learningCLIP} but found no significant gains; details and results are in \autoref{appendix:contrastive-loss}.

\begin{wraptable}{r}{0.4\textwidth}
    \begin{tabular}{l|c|c}
      model & \textsc{xsim} $\downarrow$& \textsc{xsim++} $\downarrow$ \\
      \hline
      \sonar{1} & 1.37 & 15.27 \\
      \sonartwo & 0.65 & 6.14  \\ %
    \end{tabular}
    \caption{Similarity search over 200 languages in \textsc{Flores}.}
    \label{tab:sonar}
\end{wraptable}

We design a coarse-to-fine curriculum to progressively adapt the vision encoder to more complex semantics. The alignment proceeds through three stages. In Stage 1, we initialize alignment using 12M large-scale image--caption pairs from the PLM data pipeline \citep{cho2025perceptionlm} which consists of Segment-Anything \citep{kirillov2023segmentAnything} and OpenImages \citep{kuznetsova2020openImages}. This stage establishes a basic mapping between visual and textual embeddings. Then, we introduce 2M pairs from PLM's synthetic video captioning data from YouTube1B corpus \citep{cho2025perceptionlm}. This step adapts the vision encoder to temporal dynamics while maintaining semantic consistency with \sonar. Finally, we refine the alignment using 200K high-quality human-checked video--caption pairs sourced from \PEvideo \citep{bolya2025perceptionencoder}. 

We use two versions of the \sonar encoder: \sonar{}1 is the published and open-sourced version \citep{duquenne2023sonar}.
This is the version supported by the \lcm.
We had early access to an improved version, named \sonartwo, which was trained on more data and adds three stages of contrastive training and self distillation \citep{sonar2big}.
As summarized in \autoref{tab:sonar}, \sonartwo substantially outperforms \sonar{1} on the proxy metric of multilingual similarity search. The metric \textsc{xsim++} includes hard negatives \citep{chen:2023_acl:xsimpp}.
We provide an ablation of the two \sonar versions for vision captioning tasks in \autoref{appendix:sonar1-vs-sonar2}.

\subsection{\vlcm}
\label{sec:lcm-vl}

The Large Concept Model (\lcm; \citealt{barrault2024largeConceptModelLCM}) is a latent diffusion language model operating directly in the \sonar embedding space. 
It follows an auto-regressive paradigm, predicting the next sentence embedding conditioned on preceding clean embeddings. 
For the textual modality, all embeddings are encoded and decoded by the fixed \sonar encoder and decoder.
To model the conditional distribution of the next embedding, \lcm employs a diffusion objective: given a clean embedding $x^0 \in \mathbb{R}^d$, the forward process progressively perturbs it with Gaussian noise under a variance-preserving schedule \citep{karras2022elucidating}:  
\begin{equation}
q(x_t \mid x^0) = \mathcal{N}\!\left(x_t; \alpha_t x^0, \, \sigma_t^2 \mathbf{I}\right), 
\quad x_t = \alpha_t x^0 + \sigma_t \epsilon, \;\; \epsilon \sim \mathcal{N}(0, I),
\end{equation}
where $(\alpha_t, \sigma_t)$ are determined by a monotonically decreasing log-SNR schedule $\lambda_t = \log(\alpha_t^2 / \sigma_t^2)$.  
The reverse process is parameterized by a denoiser $\mu_\theta(x_t, t, c)$, conditioned on the context embeddings $c$, with Gaussian transitions:  
\begin{equation}
p_\theta(x_{t-1} \mid x_t, c) = 
\mathcal{N}\!\left(x_{t-1}; \mu_\theta(x_t, t, c), \, \sigma_t^2 \mathbf{I}\right).
\end{equation}
Training minimizes a reconstruction loss on the original clean embedding:
\begin{equation}
\mathcal{L}(\theta) = \mathbb{E}_{t, x^0, \epsilon}\;
\bigl\| x^0 - \mu_\theta(\alpha_t x^0 + \sigma_t \epsilon, t, c) \bigr\|_2.
\end{equation}
We use the two-tower variant of \lcm, which separates the contextualizer (encoding the preceding embeddings) from the denoiser (iteratively reconstructing the next embedding). 

From the perspective of vision–language modeling, \lcm represents a new paradigm that fuses information from visual and textual modalities within a modality-agnostic latent space prior, rather than discrete visual and textual tokens \citep{team2024chameleon}. This enables autoregressive generation to be performed entirely in the latent space. Building on this principle, we further introduce \vlcm, an extension of \lcm trained through vision–language instruction fine-tuning to enhance its utility across a broad range of downstream vision-language tasks. In \vlcm, visual inputs (images and videos) are encoded into the \sonar latent space using \vsonar, while textual instructions and prompts are encoded with \sonar. The resulting visual and textual embeddings are concatenated into a single sequence, which is then processed under the same latent diffusion framework as in \lcm{}’s original text-only training, predicting the next embedding in the sequence.

\begin{table}[t!]
\centering
\footnotesize
\setlength{\tabcolsep}{2pt}
\renewcommand{\arraystretch}{1.15}
\begin{tabular}{c|l|ccccc|cc|cc}
\toprule
 & \textbf{Method} & \textbf{R@1$^{\uparrow}$}  & \textbf{R@5$^{\uparrow}$} & \textbf{R@10$^{\uparrow}$} & \textbf{MRR$^{\uparrow}$} & \textbf{AC$^{\uparrow}$} & \textbf{V. Trace$^{\uparrow}$} & \textbf{V. logdet$^{\uparrow}$} & \textbf{T. Trace$^{\uparrow}$} & \textbf{T. logdet$^{\uparrow}$} \\
\midrule
\multirow{3}{*}{\rotatebox{90}{\textbf{PE-Vid}}} 
& SigLIP2-G-OPT & 47.55 & 71.47 & 79.41 & 58.47 & 0.396 & 0.393 & \num{-1.71e4} & 0.582 & \num{-1.72e4} \\
& PECoreG        & 63.91 & 85.98 & 91.61 & 73.77 & 0.476 & 0.479 & \num{-1.35e4} & 0.686 & \num{-1.35e4} \\
& \vsonar           & \textbf{73.03} & \textbf{89.75} & \textbf{93.81} & \textbf{80.50} & \textbf{0.519} & \textbf{0.700} & \num{-1.17e4} & \textbf{2.216} & \textbf{\num{-7.98e3}} \\
\midrule
\multirow{3}{*}{\rotatebox{90}{\textbf{DREAM}}} 
& SigLIP2-G-OPT & 61.50 & 83.50 & 89.10 & 71.50 & 0.263 & 0.401 & \num{-1.79e4} & 0.662 & \num{-1.75e4} \\
& PECoreG        & \textbf{72.10} & \textbf{89.80} & \textbf{93.60} & \textbf{79.90} & 0.307 & 0.495 & \num{-1.40e4} & 0.639 & \num{-1.39e4} \\
& \vsonar           & 63.30 & 84.10 & 89.00 & 72.46 &\textbf{ 0.410} & \textbf{0.559} & \num{-1.20e4} & \textbf{2.523} & \textbf{\num{-8.53e3}} \\
\midrule
\multirow{3}{*}{\rotatebox{90}{\textbf{\vatex}}} 
& SigLIP2-G-OPT & 27.52 & 57.70 & 70.06 & 41.27 & 0.289 & 0.352 & \num{-1.74e4} & 0.660 & \num{-1.72e4} \\
& PECoreG        & 18.90 & 42.42 & 54.72 & 30.42 & 0.379 & 0.480 & \num{-1.36e4} & 0.508 & \num{-1.37e4} \\
& \vsonar           & \textbf{40.75} & \textbf{68.63} & \textbf{78.88} & \textbf{53.59} & \textbf{0.427} & \textbf{0.558} & \num{-1.19e4} & \textbf{1.660} & \textbf{\num{-8.26e3}} \\
\bottomrule
\end{tabular}
\caption{
Zero-shot Retrieval performance on \PEvideo, \dream and \vatex. 
We report the Recall rate at 1/5/10 and MRR scores.
We also report the analytical metrics for the embedding space, including 1) trace reflects overall variance, and 2) log determinant (logdet) approximates volume in the space. Best values for each columns are \textbf{bolded}.
}
\label{tab:zero-shot-text-to-video-retrieval}
\end{table}

\section{Experiments}
\label{sec:experiments}

We first verify the effectiveness of aligning the vision encoder to \sonar, by evaluating text-video retrieval and captioning using \vsonar{Omni} aligned with the \sonartwo text encoder, and provide several ablations.
We then switch to the zero-shot evaluation for \lcm, and evaluation of \vlcm on M3IT \citep{li2023m3it} which requires the use of \vsonar{1}, as the \lcm had been trained on \sonar{1}.

\subsection{Concept Space Alignment using \vsonar{Omni}}

\paragraph{Text-video Retrieval}
We treat \vsonar as a paired vision–text encoder and begin by evaluating its zero-shot performance on text-to-video retrieval, following the setup in \citet{bolya2025perceptionencoder}. We compare \vsonar against two strong baselines: the state-of-the-art \textsc{SigLIP2} vision encoder \citep{tschannen2025siglip2} and the \PE \citep{bolya2025perceptionencoder}, from which \vsonar is derived. Evaluations are conducted on three widely used video captioning benchmarks: \PEvideo (15K pairs of captioning data) \citep{bolya2025perceptionencoder}, \vatex (5K pairs of captioning data) \citep{wang2019vatex}, and \dream (1K pairs of captioning data) \citep{wang2024-dream-1k}, following the protocol in \citet{cho2025perceptionlm}. In addition to standard retrieval metrics such as Recall@$1/5/10$, we introduce three complementary measures to analyse embedding space properties: (1) Alignment Consistency (AC): the rank correlation between vision and text similarity scores, reflecting cross-modal alignment quality. (2) Trace: the trace of the covariance matrix of vision and text embeddings, indicating the spread of representations. (3) Log-determinant (logdet): the logarithm of the determinant of the covariance matrix, interpreted as the volume of the embedding ellipsoid.

\begin{table}[]
\setlength{\tabcolsep}{8pt}
\renewcommand{\arraystretch}{1.1}
\centering
\small
\begin{tabular}{lrrrrrrr}
\toprule
\textbf{Model} & \textbf{\bleu} & \textbf{R-1} & \textbf{R-2} & \textbf{R-L} & \textbf{BS-P} & \textbf{BS-R} & \textbf{BS-F} \\
\midrule
\multicolumn{8}{l}{\textbf{\PEvideo}} \\
InternVL2.5-1B               & 19.4 & 32.1 &  9.0 & 23.4 & 31.2 & 27.3 & 29.3 \\
InternVL2-1B                 & 24.1 & 35.8 & 10.7 & 25.5 & 30.8 & 32.1 & 31.5 \\
Qwen2-VL-2B-Instruct         & {29.9} & {41.7} & {18.8} & {31.2} & 34.8 & {40.0} & {37.3} \\
Qwen2.5-VL-3B-Instruct       & 30.0 & 41.3 & 16.1 & 28.9 & 30.2 & 38.6 & 34.4 \\
PLM-1B                       & 21.5 & 37.6 & 11.9 & 26.6 & 35.8 & 26.2 & 31.0 \\
PLM-3B                       & 21.1 & 37.5 & 11.7 & 26.4 & 36.6 & 26.1 & 31.3 \\
\vsonar w/ \sonartwo Decoder & \textbf{39.0} & \textbf{50.1} & \textbf{23.3} & \textbf{38.0} & \textbf{44.4} & \textbf{41.6} & \textbf{43.0} \\
\midrule
\multicolumn{8}{l}{\textbf{\dream}} \\
InternVL2.5-1B               & 10.2 & 21.5 &  3.7 & 15.6 & \textbf{26.1} & 11.2 & 18.6 \\
InternVL2-1B                 & 14.6 & 25.0 &  4.3 & 17.2 & 23.8 & 15.2 & 19.5 \\
Qwen2-VL-2B-Instruct         & 19.7 & 27.1 &  5.2 & 18.5 & 12.9 & 14.8 & 13.9 \\
Qwen2.5-VL-3B-Instruct       & 16.1 & 23.9 &  4.4 & 15.9 &  1.6 & 15.6 &  8.6 \\
PLM-1B                       & 18.5 & 27.0 &  6.4 & 19.3 & 14.5 & 16.8 & 15.5 \\
PLM-3B                       & 19.6 & 28.6 &  6.7 & 20.4 & 19.9 & 18.1 & 19.0 \\
\vsonar w/ \sonartwo Decoder & \textbf{23.9} & \textbf{32.7} & \textbf{8.4} & \textbf{22.7} & 19.7 & \textbf{21.6} & \textbf{20.7} \\
\midrule
\multicolumn{8}{l}{\textbf{\vatex}} \\
InternVL2.5-1B               & 41.5 & 23.3 &  4.4 & 19.2 & 37.6 & 45.2 & 40.2 \\
InternVL2-1B                 & \textbf{47.8} & \textbf{27.3} & \textbf{6.4} & \textbf{22.4} & \textbf{36.9} & \textbf{50.4} & \textbf{42.4} \\
Qwen2-VL-2B-Instruct         & 32.1 & 19.8 &  6.0 & 16.4 & 18.7 & 46.3 & 30.8 \\
Qwen2.5-VL-3B-Instruct       & 29.4 & 18.3 &  5.1 & 15.0 & 12.1 & 47.1 & 27.6 \\
PLM-1B                       & 33.4 & 21.8 &  5.7 & 19.1 & 15.1 & 48.1 & 29.6 \\
PLM-3B                       & 34.0 & 22.1 &  5.9 & 19.3 & 16.9 & 48.6 & 30.8 \\
\vsonar w/ \sonartwo Decoder  & 26.7 & 17.2 &  5.0 & 14.8 & 13.6 & 42.8 & 26.5 \\ \midrule
\multicolumn{8}{l}{\textbf{\vatex{}-zh}} \\
InternVL2-1B-Instruct         & 22.3 & 14.1 &  2.9 & 11.8 & 8.7 & 18.2 & 12.6 \\
InternVL2.5-1B-Instruct       & \textbf{33.2} & 22.5 &  4.4 & 18.8 & 19.1 & 31.5 & 24.2 \\
\vsonar w/ \sonartwo Decoder  & 30.6 & \textbf{32.1} &  \textbf{8.53} & \textbf{26.9} & \textbf{23.2} & \textbf{48.5} & \textbf{33.7} \\
\bottomrule
\end{tabular}
\caption{Video captioning performance across \PEvideo, \dream and VATEX (English and Chinese). Metrics include \bleu, \rouge (R-1, R-2, R-L), and BERTScore (BS-P, BS-R, BS-F).}
\label{tab:zero_shot_captioning}
\end{table}

\autoref{tab:zero-shot-text-to-video-retrieval} summarizes the results and embedding space statistics. On the three datasets, \PEvideo, \dream and \vatex, \vsonar significantly outperforms SigLIP2, achieving improvements of 9.12, 1.8 and 13.23 points in Recall@1, respectively, demonstrating the effectiveness of our approach for retrieval tasks. Compared to the original \PE, \vsonar significantly improves on \PEvideo and \vatex (9.12 and 21.85 score at Recall@1), though it loses 8.8 score at Recall@1 in \dream, indicating that our curriculum alignment strategy preserves strong retrieval capability. These results confirm that a vision encoder can be successfully aligned with a purely text-trained embedding space (\sonar) in a post-hoc manner. Finally, our embedding space analysis reveals that \vsonar maintains a more expanded distribution. Moreover, by freezing the original \sonar space, \vsonar achieves the largest textual embedding dispersion, as evidenced by the highest trace and logdet values among all compared models.

\paragraph{Text-video Captioning}

Different with the traditional vision encoder, such as SigLIP 2 or \PE, aligning \vsonar to \sonar embedding space allows us to leverage the \sonar decoder to directly verbalize the encoded vector of \vsonar. Hence, we conduct the zero-shot evaluation on video captioning for \vsonar, and  compare it with few state-of-the-art vision-language models (VLMs) including InternVL-2/2.5 \citep{chen2024internvl}, Qwen-VL 2/2.5 \citep{wang2024qwen2vl, bai2025qwen2_5_vl} and Perception Language Models \citep{cho2025perceptionlm}. We compare with VLMs at the scale between 1B to 3B for a fair comparison, as the \sonar decoder is at 1.5B and \vsonar is at 1.9B. We evaluate the models with lexical metrics including \bleu and \rouge scores, and semantic metrics including BERTScore-Precision/Recall/F1, following \citep{zhang2025pretrainedImageTextModelsVideoCaptioner}. 

We illustrate the results in \autoref{tab:zero_shot_captioning}. For detailed captioning benchmarks such as \PEvideo and \dream, we observe that \vsonar paired with the \sonar decoder can achieve a state-of-the-art performance. In particular, \vsonar improves the second best model, Qwen2.5-VL-3B-Instruct, by 9 points in \bleu. The only exception is \vatex where the captions are relatively short as one sentence, \vsonar lags behind InternVL2; however, this is expected as we align \vsonar with \sonar mostly with the detailed caption data. And we observe \vsonar is still comparable with PLM and Qwen-VL series. We use \vatex{}-Chinese validation set for the multilingual evaluation, and we mostly compare \vsonar with InternVL, as QwenVL is reported to leverage \vatex Chinese split during training \citep{wang2024qwen2vl, bai2025qwen2_5_vl}, and PLM-1/3B fail to support the fluent generation in Chinese. We find that in \vatex Chinese split, \vsonar still outperforms the InternVL series, indicating the advantage in multilingual evaluation.

\begin{figure}[h]
    \centering
    \includegraphics[width=1\linewidth]{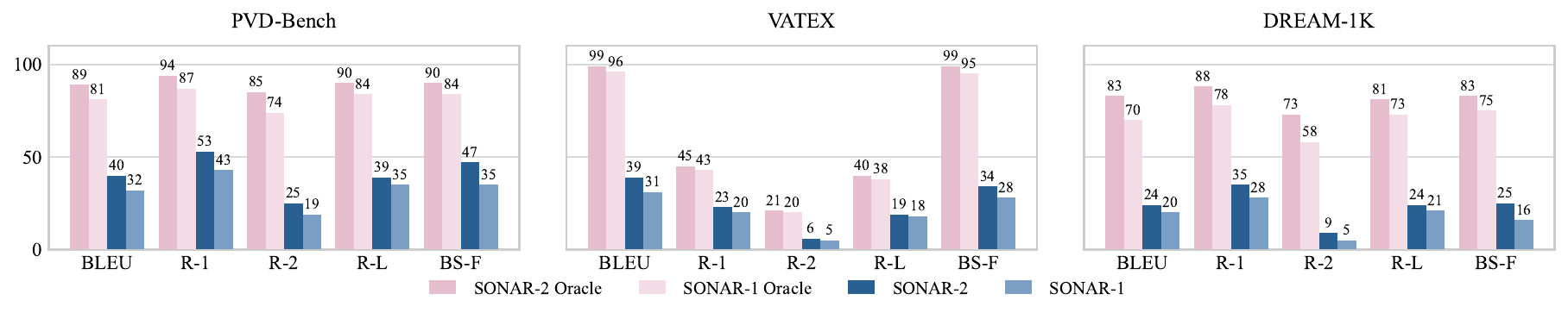}
    \caption{Comparison for \vsonar trained with SONAR version 1 and 2 (\sonartwo) embedding space on three captioning datasets.}
    \label{fig:sonar1-vs-sonar2}
\end{figure}

\paragraph{\vsonar 1 vs \vsonar{Omni}.}
We present the comparison between \sonar \citep{duquenne2023sonar} and \sonartwo in \autoref{fig:sonar1-vs-sonar2}. We report both \sonar space's oracle performance (we encode the reference caption with \sonar encoder, and decoded with \sonar decoder). And we report the zero-shot performance with \vsonar (we encode the video with \vsonar and decode with \sonar decoder). \sonar oracle serves as an estimation of the upper-bound performance that \vsonar can achieve for leveraging \sonar decoder.

We observe that both \sonar versions have a strong oracle performance, indicating \sonar's encoding and decoding from textual space into its representation space is quite lossless. Specifically, in PE-Video, \vatex and \dream, \sonartwo can achieve \bleu scores of 81, 96 and 70. Comparing the zero-shot performance for \sonar and \sonartwo, we see \sonar{1} is worse by a considerable margin. We hypothesize that \sonar is harder to align since its space is reported to be collapsed. Our analysis for \sonar and \sonartwo also supports this observation: in PVD, \sonar and \sonartwo have the embeddings norm at 0.264 and 1.69, and covariance trace at 0.049 and 1.83, respectively. The comparison in retrieval tasks are in \autoref{appendix:sonar1-vs-sonar2}.

\begin{table}[t]
\centering
\setlength{\tabcolsep}{3pt}
\renewcommand{\arraystretch}{1.1}
\begin{tabular}{c|l|cc|ccccccc}
\toprule
 & \textbf{} & \textbf{MSE} & \textbf{Cos. Sim.} & \textbf{\bleu} & \textbf{R-1} & \textbf{R-2} & \textbf{R-L} & \textbf{BS-P} & \textbf{BS-R} & \textbf{BS} \\
\midrule
\multirow{6}{*}{\rotatebox{90}{\textbf{Architecture}}} 
& Linear Proj. & $1.45\!\times\!10^{-3}$ & 0.694  & 38.0 & 49.7 & 21.6 & 36.7 & 47.2 & 40.1 & 43.7 \\
& Full PE & $1.54\!\times\!10^{-3}$ & 0.672 & 37.1 & 48.5 & 21.3 & 36.5 & 46.9 & 38.8 & 42.9 \\
& + Async. LR & $1.43\!\times\!10^{-3}$ & 0.700 & 39.7 & 51.3 & 23.3 & 37.7 & 48.1 & 42.1 & 45.1 \\
& + Norm. Init. & $1.39\!\times\!10^{-3}$ & 0.708 & 39.8 & 51.8 & 24.0 & 38.5 & 49.4 & 42.2 & 45.8 \\
& + Attn. Pooling & $1.39\!\times\!10^{-3}$ & 0.708 & 39.8 & 51.9 & 24.0 & 38.5 & 49.7 & 42.4 & 46.0 \\
& + Temporal Attn. & $1.39\!\times\!10^{-3}$ & 0.708 & 39.8 & 51.9 & 24.0 & 38.5 & 49.7 & 42.4 & 46.1 \\
\midrule
\multirow{3}{*}{\rotatebox{90}{\textbf{Pipeline}}} 
& Full Pipeline & {$1.36\!\times\!10^{-3}$} & {0.716} & {40.1} & {52.6} & {24.9} & {39.2} & {50.8} & {43.2} & {47.0} \\
& \textit{w/o} SV & $1.39\!\times\!10^{-3}$ & 0.710 & 39.6 & 51.9 & 24.1 & 38.6 & 50.0 & 42.4 & 46.2 \\
& \textit{w/o} IC \& SV & $1.39\!\times\!10^{-3}$ & 0.708 & 39.8 & 51.9 & 24.0 & 38.5 & 49.7 & 42.4 & 46.1 \\
\bottomrule
\end{tabular}
\caption{
Ablation study in model architecture and the three-stage training {pipeline}. SV: our second stage curriculum with the synthetic video captioning data. IC: our first stage curriculum with image captioning data.
}
\label{tab:ablation}
\end{table}

\subsection{Ablation Study}

We conduct an ablation study for model architecture design, and our proposed training pipeline on the \PEvideo test set (\autoref{tab:ablation}). 

\paragraph{Model Architecture} We ablate architectural choices for the projector network. As a baseline, we evaluate linear projection (Linear Proj.), where the \PE is frozen and only a linear layer is trained, and full-model fine-tuning (Full PE), where the encoder is updated jointly. Linear projection performs better, indicating that the encoder’s contrastive pre-training already yields strong semantic alignment, while full fine-tuning is hindered by unstable gradients from the randomly initialized projector. To mitigate this, we adopt strategies that incrementally improve downstream performance, including asynchronous learning rates for the projector and encoder, initialization trick, attention-based aggregation strategy for video frames' features, and temporal attention layer.

\paragraph{Data Mixture} We then ablate the second stage synthetic video captioning and first-stage image captioning curriculum on our pipeline. We observe that both stages contribute positively to the downstream performance in captioning performance on PE-Video. Removing the 12M image captioning pairs and 2M video captioning pairs reduce the \bleu with 0.3 and 0.2, respectively.

\subsection{Zero-shot Processing \vsonar Embeddings by Large Concept Model}

Since the \lcm \citep{barrault2024largeConceptModelLCM} operates directly on \sonar{1}, it should seamlessly transfer its ability and understand the visual concepts in \vsonar aligned with \sonar{1}. We examine \vsonar with \lcm gradually from  single to multiple vision concept understanding tasks, where the \lcm accepts the instruction encoded by \sonar, with the vision embeddings from \vsonar, and predicts the target embedding. Note that in both experiments, we do not fine-tune the \lcm, with neither any video data nor captioning data. Thus, the \lcm is only trained in English textual corpus including its pre-training and instruction fine-tuning as in \citet{barrault2024largeConceptModelLCM}. We compare \lcm{}'s performance with VLMs at 7/8-B scale for the InternVL series \citep{chen2024internvl}, Qwen-VL \citep{bai2025qwen2_5_vl, team2024qwen2} and PLM \citep{cho2025perceptionlm}.

\paragraph{Single Vision Concept Understanding: Video Captioning}

\sisetup{detect-all, table-format=2.1, table-number-alignment=center}

\begin{table}[t]
\centering
\small
\setlength{\tabcolsep}{4pt}
\resizebox{\linewidth}{!}{
\begin{tabular}{lc|cccccccc|cccc|ccc}
\toprule
\multicolumn{2}{c}{\multirow{3}{*}{}} & \multicolumn{8}{|c}{\textbf{Video Captioning / Summarization}}                                                                                              & \multicolumn{4}{|c|}{\textbf{M3IT Image}}                                   & \multicolumn{3}{c}{\textbf{M3IT Video}}                     \\
\cmidrule{3-17}
\multicolumn{2}{c}{}                                & \multicolumn{2}{|c}{\rotatebox{90}{\scriptsize \textbf{PE-Video}}} & \multicolumn{2}{c}{\rotatebox{90}{\scriptsize \textbf{\dream}}} & \multicolumn{2}{c}{\rotatebox{90}{\scriptsize \textbf{\vatex}}} & \multicolumn{2}{c|}{\rotatebox{90}{\scriptsize \textbf{\videoxum}}} & \rotatebox{90}{\scriptsize \textbf{COCO}} & \rotatebox{90}{\scriptsize \textbf{VIQUAE}} & \rotatebox{90}{\scriptsize \textbf{VisualMRC}} & \rotatebox{90}{\scriptsize \textbf{ScienceQA}} & \rotatebox{90}{ \scriptsize \textbf{ActivNetQA}} & \rotatebox{90}{\scriptsize \textbf{MSRVTT-QA}} & \rotatebox{90}{\scriptsize \textbf{IVQA}} \\
\cmidrule(lr){3-4} \cmidrule(lr){5-6} \cmidrule(lr){7-8} \cmidrule(lr){9-10} \cmidrule(lr){11-11} \cmidrule(lr){12-12} \cmidrule(lr){13-13} \cmidrule(lr){14-14} \cmidrule(lr){15-15} \cmidrule(lr){16-16} \cmidrule(lr){17-17} 
&                                & R-L               & BS                & R-L               & BS                & R-L              & BS              & R-L               & BS                & R-L           & R-L             & R-L                & Acc.               & R-L                    & R-L                & R-L           \\

\midrule

\multirow{3}{*}{\textbf{InterVL2}}        & 1B      & 25.5              & 31.5              & 17.2              & 19.5              & 22.4             & 42.4            & 15.3              & 17.7              & 12.6          & 24.0            & 30.6               & 53.9               & 40.6                   & 27.6               & 39.5          \\
                                          & 4B      & 15.0              & 18.4              & 12.2              & 14.7              & 19.2             & 42.4            & 15.6              & 17.4              & 17.5          & 20.0            & 35.3               & 89.6               & 27.5                   & 24.5               & 31.8          \\
                                          & 8B      & 18.6              & 23.4              & 16.4              & 19.4              & 16.2             & 42.4            & 29.1              & 26.1              & 21.0          & 21.6            & 42.9               & 87.2               & 29.7                   & 27.1               & 38.4          \\ \midrule
\multirow{3}{*}{\textbf{InternVL-2.5}}    & 1B      & 23.4              & 29.3              & 15.6              & 18.6              & 19.2             & 40.3            & 17.1              & 23.2              & 13.2          & 10.8            & 27.3               & 69.0               & 16.6                   & 11.7               & 19.3          \\
                                          & 4B      & 14.6              & 18.0              & 13.3              & 14.8              & 17.3             & 36.2            & 18.1              & 23.0              & 15.1          & 23.1            & 45.2               & 86.4               & 26.8                   & 21.8               & 24.9          \\
                                          & 8B      & 21.4              & 26.0              & 17.0              & 17.0              & 20.6             & 42.4            & 24.9              & 20.5              & 16.8          & 17.3            & 42.8               & \textbf{93.1}      & 20.9                   & 16.9               & 22.7          \\ \midrule
\multirow{2}{*}{\textbf{Qwen2-VL}}        & 2B      & 31.2              & \textbf{37.3}     & 18.5              & 13.9              & 16.4             & 30.8            & 23.6              & 29.8              & 24.9          & \textbf{50.2}   & 56.1               & 54.5               & 53.7                   & 39.6               & 49.4          \\
                                          & 7B      & 26.9              & 32.6              & 19.8              & 18.1              & 28.5    & \textbf{51.6}   & 26.0              & 32.4              & 23.7          & 49.7            & \textbf{57.4}      & 70.4               & 41.9                   & 22.7               & 39.1          \\ \midrule
\multirow{2}{*}{\textbf{Qwen2.5-VL}}      & 3B      & \textbf{28.9}     & 34.4              & 15.9              & 8.6               & 15.0             & 27.6            & 26.0              & 32.9              & 25.1          & 48.3            & 55.7               & 55.0               & 52.1                   & 41.6               & 48.5          \\
                                          & 7B      & 22.2              & 25.9              & 15.7              & 10.5              & 27.5             & 50.8            & 24.1              & 28.9              & 18.5          & 34.5            & 45.0               & 61.6               & 46.0                   & 41.4               & 54.2          \\ \midrule
\multirow{3}{*}{\textbf{Percep. LM}}      & 1B      & 26.6              & 31.0              & 19.3              & 15.5              & 19.1             & 29.6            & 21.8              & 33.2              & 27.5          & 30.8            & 45.5               & 73.6               & 27.8                   & 14.5               & 39.1          \\
                                          & 3B      & 26.4              & 31.3              & 20.4              & 19.0              & 19.3             & 30.8            & \textbf{27.0}     & \textbf{36.4}     & 34.3          & 23.7            & 51.1               & 89.8               & 28.0                   & 19.4               & 26.1          \\
                                          & 8B      & 27.4              & 31.9              & \textbf{20.8}     & \textbf{19.7}     & 19.0             & 30.8            & 26.2              & 33.7              & 36.3          & 31.0            & 50.0               & 87.7               & 40.5                   & 25.3               & 41.4          \\ \midrule
\multirow{2}{*}{\textbf{LCM}}             & \lcm     & 25.5              & 27.9              & 18.5              & 16.6              & 23.8             & 30.8            & 21.5              & 22.1              & 18.0          & 34.3            & 33.5               & 44.7               & 51.7                   & 36.0               & 48.9          \\
                                          & \vlcm   & 27.4              & 30.0              & 19.8              & 19.2              & \textbf{28.8}             & 48.7            & 20.6              & 25.3              & \textbf{38.8} & 39.4            & 34.1               & 76.2               & \textbf{63.6}          & \textbf{48.7}      & \textbf{63.9} \\ \bottomrule
\end{tabular}
}
\caption{Main results on vision-language tasks in M3IT and the previous video benchmarks (\PEvideo, \dream, \vatex, VideoXum). }
\label{tab:vsft-result}
\end{table}

We report the results of \lcm on video captioning in \autoref{tab:vsft-result}. Compared to the strongest baseline, the zero-shot \lcm lags behind by 1.15/4.44/4.76 \bleu scores on \PEvideo, \dream and \vatex, respectively. Among the models, PLM-8B delivers the strongest overall performance. The relatively narrow performance gap between \lcm and competitive VLMs suggests that the \lcm is able to understand the single vision concept, despite never being trained with video data.

\paragraph{Multiple Vision Concept Understanding: Long Video Summarization}

We next evaluate \lcm in a setting requiring understanding multiple visual embeddings. Long videos are uniformly segmented into snippets with 8 frames per each, with each snippet encoded by \vsonar as a separate video embedding.
Since the \lcm shows strong performance in document summarization \citep{barrault2024largeConceptModelLCM} for multiple \sonar embeddings; we hypothesize that it should be capable of performing zero-shot summarization over sets of video embeddings from \vsonar. For this evaluation, we use \videoxum \citep{lin2023videoxum}, which contains videos of one to five minutes, uniformly split into snippets of 8 frames each.

We report the \videoxum results in \autoref{tab:vsft-result}. Again, PLM achieves the strongest performance among competitive VLMs at the same scale. \vsonar + \lcm achieves 22.1 score at BertScore-F1, trailing the best-performing PLM-8B (33.7) but is slightly higher than InternVL-2.5-8B at 20.5. These findings indicate that, even without exposure to any video data during training, \lcm demonstrate non-trivial understanding of multiple \vsonar embeddings for long videos.

\paragraph{Reasoning in \vsonar}
\begin{wrapfigure}{r}{0.4\linewidth}
    \centering
    \vspace{-20pt} %
    \includegraphics[width=0.8\linewidth]{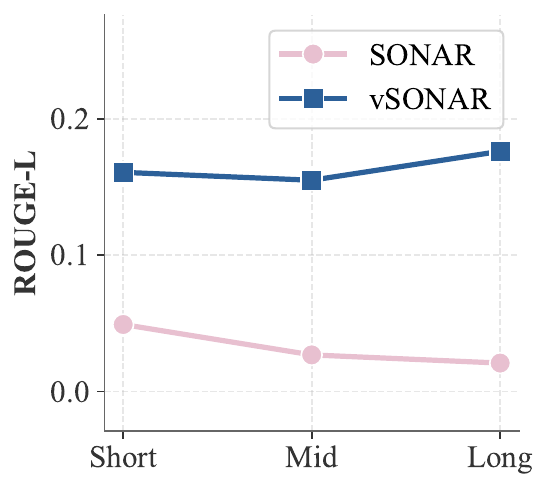}
    \caption{Operating in \vsonar space, the \lcm performs better than only accepting the textual \sonar inputs. We compare \lcm-7B-IFT in \videoxum with \rouge-L scores across short, mid, and long categories of video inputs.}
    \label{fig:sonar_vsonar_ablation}
    \vspace{-10pt}
\end{wrapfigure}
We then investigate whether \lcm truly leverages the latent representations in \vsonar for multimodal reasoning. To test this, we compare two settings: (1) encoding video clips directly into \vsonar embeddings, which are then fed into \lcm for zero-shot summarization; and (2) decoding video embeddings into captions using the \sonar decoder, re-encoding them with \sonar, and providing these \sonar embeddings to \lcm. Our hypothesis is that \vsonar embeddings retain richer visual features than their textual equivalents in \sonar, and thus should yield stronger performance if \lcm relies on visual representations.

We group videos into short (<90s), mid-length (90–150s), and long (>150s) categories, and report \rouge-L scores in \autoref{fig:sonar_vsonar_ablation}. Across all categories, \lcm with \vsonar consistently outperforms it with \sonar. Notably, while \sonar performance declines with increasing video length, \vsonar remains stable, highlighting its robustness. These results support our hypothesis that the \lcm reasons directly in the visual embedding space provided by \vsonar containing richer visual information than \sonar representations of textual input.

\paragraph{\vlcm Reasoning with Visual Details.} 
\begin{wraptable}{r}{0.45\textwidth}
\vspace{10pt} %
\centering
\begin{tabular}{lll}
\toprule
\textbf{VCR}  & \textbf{F1}    & \textbf{Sim.} \\ \midrule
LCM           & 0.385          & 0.258                        \\
\textbf{vLCM} & \textbf{0.671} & \textbf{0.529}               \\ \midrule
PLM-8B        & 0.441          & 0.432                        \\
Qwen-2.5-7B   & 0.275          & 0.402                        \\
Qwen-2-7B     & 0.502          & 0.513                        \\
InterVL2.5-8B & 0.155          & 0.158                        \\
InterVL2-8B   & 0.325          & 0.340                  \\ \bottomrule
\end{tabular}
\caption{
VCR results comparing token-level F1 and semantic similarity.
}
\label{tab:vcr-result}
\end{wraptable}

To assess whether our alignment from \PE to \sonar preserves the visual details critical for downstream reasoning, we further evaluate visual commonsense reasoning (VCR) performance on M3IT (\autoref{tab:vcr-result}). In this benchmark, models must reason about objects using commonsense knowledge while interpreting their associated bounding boxes. Compared to semantic-level alignment alone, VCR demands substantially richer visual grounding and layout awareness (see \autoref{fig:qualitative_vcr_2}). We report token-level F1 scores against the ground truth rationales, alongside semantic similarity to the reference explanations. The strong performance of \vlcm indicates that, despite being trained solely on semantic-level captions, \vlcm effectively leverage layout grounding and spatial relationship preserved by \vsonar in reasoning tasks.

\subsection{\vlcm}

\paragraph{In-task Performance} We next evaluate \vlcm, which is supervised fine-tuned on M3IT \citep{li2023m3it} to better capture and reason over visual concepts. 
We mainly rely on M3IT \citep{li2023m3it} as it supports a variety of tasks, and the wide coverage for up to 80 languages ranging from high- to low-resource languages.
The evaluation covers 7 datasets spanning 5 tasks defined in M3IT: (1) image captioning (COCO), (2) visual QA (VIQUAE), (3) document image QA (VisualMRC), (4) video captioning (MSRVTT), and (5) question answering (IVQA, MSRVTT-QA, ActivityNetQA). In addition, we report \vlcm{}’s performance on the video captioning and long video summarization benchmarks introduced in the previous section. Following the evaluation protocol in \citet{li2023m3it}, we use \mbox{\rouge-L} for generative tasks (e.g., captioning and open-ended QA) and accuracy for multiple-choice QA.

\autoref{tab:vsft-result} compares \lcm with strong open-source vision–language models. We observe that \vlcm substantially outperforms the zero-shot \lcm across most benchmarks. For example, \vlcm achieves 63.9 R-L on IVQA and 63.6 R-L on ActivityNetQA, surpassing 48.9 and 51.7 for \lcm, representing clear gains from training with vision instruction-tuning data. While \vlcm lags behind the best-performing models on some benchmarks such as VisualMRC, VIQUAE, and ScienceQA, it achieves state-of-the-art results on video question answering tasks, including IVQA, ActivityNetQA, and MSRVTT-QA. Meanwhile, performance on our previous video captioning and summarization datasets remains competitive: \vlcm attains 27.4 R-L on PE-Video and 19.8 R-L on DREAM-1K, trailing the best model by only 1.5 and 1 \rouge scores, highlighting the generalization of \lcm to unseen datasets during training. 

\begin{figure}
    \centering
    \includegraphics[width=\linewidth]{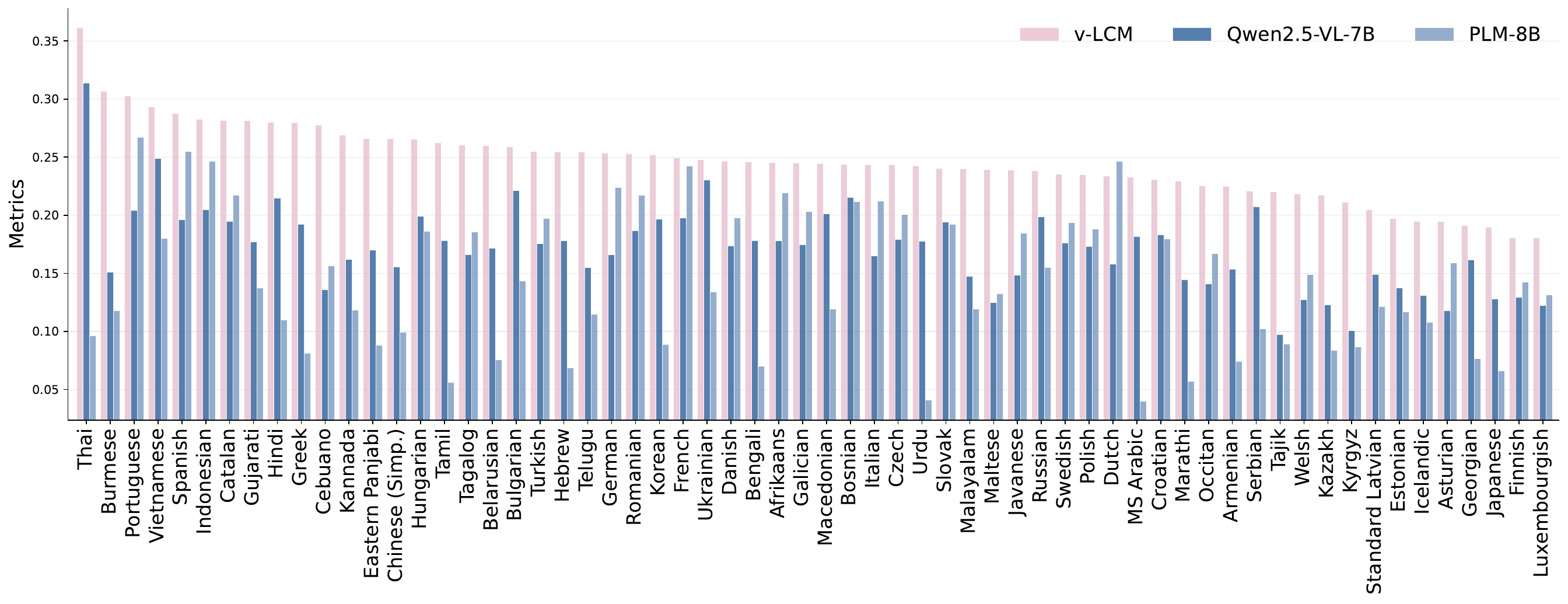}
    \caption{Performance in 62 languages for \vlcm, Qwen2.5-VL-7B and PLM-8B on M3IT testing set for MSRVTT, MSRVTT-QA, ImageNet, VQA-V2, VIST and OKVQA. We report the \rouge-L scores averaged from all datasets. Detailed results for each dataset can be found in the \autoref{appendix:detailed-m3it-multilinguaul}.}
    \label{fig:multilingual-m3it-eval}
\end{figure}

\vspace{-0.3em}
\paragraph{Multilinguality}
We further conduct a multilingual evaluation of \vlcm on the M3IT benchmark across 62 languages,\footnote{The intersection of all languages supported by \sonar, M3IT, and multilingual \rouge.} leveraging the fact that \vlcm operates entirely in the latent space of \sonar and \vsonar, and can therefore decode outputs to any languages supported by \sonar. Evaluation spans five various tasks including image classification (ImageNet), image question answering (VQA-V2, OKVQA), video question answering (MSRVTT-QA), video captioning (MSRVTT) and narrative generation (VIST), covering a spectrum from high-resource languages (e.g., Chinese), mid-resource languages (e.g., Japanese) to low-resource languages (e.g., Javanese). We use the \rouge-L implementation from \citep{shohan2024xl-multilingualROUGE} for the multilingual evaluation.

As shown in \autoref{fig:multilingual-m3it-eval}, \vlcm consistently outperforms Qwen2.5-VL-7B and PLM-8B across 61 of 62 languages, with Dutch being the only exception. While improvements in some high-resource languages are modest (e.g. French), the gains become substantial in mid- and low-resource settings, including Burmese, Tajik and Telugu. Notably, for languages such as Urdu, modern Arabic and Tamil, which is unsupported by PLM-8B based on LLaMA-3.2 \citep{touvron2023llama,dubey2024llama3}, \vlcm successfully generates meaningful outputs, whereas competing models fail entirely.

\section{Related Works}
\label{sec:relatd-work}
A central paradigm in multimodal learning is to align vision and language representations into a shared embedding space. Early approaches such as CLIP \citep{radford2021learningCLIP} and ALIGN \citep{jia2021scalingAlign} established large-scale contrastive learning between paired images and captions, enabling zero-shot transfer to downstream tasks. Subsequent works extended this idea to video–language pretraining \citep{lei2021less,xu2021videoclip,wang2022allinone}. 
More recent efforts focus on aligning pretrained encoders into an unified space: Perception Encoder \citep{bolya2025perceptionencoder} projects diverse perceptual modalities into a shared latent space, while scaling data and architectures in models, such as Florence \citep{yuan2021florence} and SigLIP2 \citep{tschannen2025siglip2}, further improve alignment quality. 
Recent work also shows that using large language models as text encoders enhances vision–language alignment \citep{stone2025learningCompositionLLMCLIP}, and post-hoc alignment strategies have been proposed as lightweight alternatives to joint training \citep{brokowski-ssca,yang2025languageAlignmentFixedTextEncoder}.

Parallel advances in multilingual text embedding models, such as \textsc{LASER} \citep{Artetxe:2019:tacl_massive_ml,laser3}, \textsc{LaBSE} \citep{labse}, and \sonar \citep{duquenne2023sonar,sonar2big}, demonstrate the effectiveness of language-agnostic embedding spaces across hundreds of languages. 
Modular approaches have further explored language-specialized components to reduce interference in universal embedding spaces \citep{huang2024modularSentenceEncoderLanugageSpecialization}. These universal text embeddings provide an attractive target for aligning vision encoders, as they inherit cross-lingual generalization without requiring multimodal data in every language. Prior work has explored similar strategies in speech-to-text alignment \citep{chung2018unsupervised,Duquenne:2021:neurips,laperriere2024dual,du2024cosyvoice}, but large-scale alignment of visual embeddings into such universal text spaces remains underexplored.

\section{Conclusion}
\label{sec:conclusion}

We introduce \vsonar by extending the \sonar embedding space with the image and video modality.
To the best of our knowledge, this makes \sonar the most universal embedding space covering four modalities (text, speech, image and video) and up to 200 languages.
We propose a three-stage training approach to map a pooled representation based on the \PE to the semantic \sonar representation.
We achieve very competitive results for text-to-video retrieval and video captioning.
The Large Concept Model (\lcm; \citealt{barrault2024largeConceptModelLCM}) is a recent approach to perform reasoning at a higher semantic conceptual level, namely \sonar.
Encoded by \vsonar, we show that the \lcm can zero-shot process image or video embeddings without the need of training data in these modalities.
We further introduce \vlcm with the multimodal instruction fine-tuning, which matches state-of-the-art VLMs, while significantly outperforming them across 61 rich- to low-resource languages.

\bibliographystyle{assets/plainnat}
\bibliography{iclr2025_conference}

@STRING{acl = "ACL"}

@STRING{tacl = "TACL"}

@STRING{emnlp = "EMNLP"}

@misc{sonar2big,
  title = {{SONAR2}: Omnilingual Sentence Embeddings for Text and Speech},
  author = {{Omnilingual Embeddings Team} and Pere-Lluís Huguet Cabot and João Maria Janeiro and Yen Meng and Ioannis Tsiamas† and Vivek Iyer and Jaehyeong Jo and Guillem Ramírez and Belen Alastruey and Loic Barrault and David Dale and Kevin Heffernan and Artyom Kozhevnikov and Alex Mourachko and Christophe Ropers and Holger Schwenk and Yu-An Chung and Marta R. Costa-Jussa and Paul-Ambroise Duquenne},
  pages = {1--64},
  year = 2026,
  note = {to be published},
}

@inproceedings{chen:2023_acl:xsimpp,
    title = "x{SIM}++: An Improved Proxy to Bitext Mining Performance for Low-Resource Languages",
    author = "Chen, Mingda  and
      Heffernan, Kevin  and
      {\c{C}}elebi, Onur  and
      Mourachko, Alexandre  and
      Schwenk, Holger",
    editor = "Rogers, Anna  and
      Boyd-Graber, Jordan  and
      Okazaki, Naoaki",
    booktitle = ACL,
    year = "2023",
    url = "https://aclanthology.org/2023.acl-short.10/",
    pages = "101--109",
}

@article{ramesh-etal-2022-samanantar,
    title = "Samanantar: The Largest Publicly Available Parallel Corpora Collection for 11 {I}ndic Languages",
    author = "Ramesh, Gowtham  and
      Doddapaneni, Sumanth  and
      Bheemaraj, Aravinth  and
      Jobanputra, Mayank  and
      AK, Raghavan  and
      Sharma, Ajitesh  and
      Sahoo, Sujit  and
      Diddee, Harshita  and
      J, Mahalakshmi  and
      Kakwani, Divyanshu  and
      Kumar, Navneet  and
      Pradeep, Aswin  and
      Nagaraj, Srihari  and
      Deepak, Kumar  and
      Raghavan, Vivek  and
      Kunchukuttan, Anoop  and
      Kumar, Pratyush  and
      Khapra, Mitesh Shantadevi",
    journal = "TACL",
    volume = "10",
    year = "2022",
    url = "https://aclanthology.org/2022.tacl-1.9/",
    pages = "145--162",
}

@article{lin2023videoxum,
  title={Videoxum: Cross-modal visual and textural summarization of videos},
  author={Lin, Jingyang and Hua, Hang and Chen, Ming and Li, Yikang and Hsiao, Jenhao and Ho, Chiuman and Luo, Jiebo},
  journal={IEEE Transactions on Multimedia},
  volume={26},
  pages={5548--5560},
  year={2023},
  publisher={IEEE}
}

@misc{nllb2022,
  author =        {{NLLB Team} and Costa-jussà, Marta R. and
                   Cross, James and Çelebi, Onur and Elbayad, Maha and
                   Heafield, Kenneth and Heffernan, Kevin and
                   Kalbassi, Elahe and Lam, Janice and Licht, Daniel and
                   Maillard, Jean and Sun, Anna and Wang, Skyler and
                   Wenzek, Guillaume and Youngblood, Al and Akula, Bapi and
                   Barrault, Loic and Mejia-Gonzalez, Gabriel and
                   Hansanti, Prangthip and Hoffman, John and
                   Jarrett, Semarley and Sadagopan, Kaushik Ram and
                   Rowe, Dirk and Spruit, Shannon and Tran, Chau and
                   Andrews, Pierre and Ayan, Necip Fazil and
                   Bhosale, Shruti and Edunov, Sergey and Fan, Angela and
                   Gao, Cynthia and Goswami, Vedanuj and
                   Guzmán, Francisco and Koehn, Philipp and
                   Mourachko, Alexandre and Ropers, Christophe and
                   Saleem, Safiyyah and Schwenk, Holger and Wang, Jeff},
  title =         {No Language Left Behind: Scaling Human-Centered
                   Machine Translation},
  year =          {2022},
  url =           {https://arxiv.org/abs/2207.04672},
}

@inproceedings{Duquenne:2021:neurips,
  author =        {Duquenne, Paul-Ambroise and Gong, Hongyu and
                   Schwenk, Holger},
  booktitle =     {Advances in Neural Information Processing Systems},
  editor =        {M. Ranzato and A. Beygelzimer and Y. Dauphin and
                   P.S. Liang and J. Wortman Vaughan},
  pages =         {15748--15761},
  publisher =     {Curran Associates, Inc.},
  title =         {Multimodal and Multilingual Embeddings for
                   Large-Scale Speech Mining},
  volume =        {34},
  year =          {2021},
  url =           {https://proceedings.neurips.cc/paper_files/paper/2021/file/
                  8466f9ace6a9acbe71f75762ffc890f1-Paper.pdf},
}

@InProceedings{Reimers,
  title = {Making Monolingual Sentence Embeddings Multilingual using Knowledge Distillation},
  author = {Nils Reimers and Iryna Gurevych},
  year = 2020,
  booktitle = emnlp,
  pages = {4512--4525},
}

@article{labse,
  title={Language-agnostic bert sentence embedding},
  author={Feng, Fangxiaoyu and Yang, Yinfei and Cer, Daniel and Arivazhagan, Naveen and Wang, Wei},
  journal={arXiv preprint arXiv:2007.01852},
  year={2020}
}

@article{Artetxe:2019:tacl_massive_ml,
  title = {Massively Multilingual Sentence Embeddings for Zero-Shot Cross-Lingual Transfer and Beyond},
  author = {Mikel Artetxe and Holger Schwenk},
  journal = tacl,
  vol = "7",
  year = {2019},
  pages = "597--610",
}

@InProceedings{schwenk:2019:arxiv_ccmatrix,
  title={{CCMatrix}: Mining billions of high-quality parallel sentences on the web},
  author={Schwenk, Holger and Wenzek, Guillaume and Edunov, Sergey and Grave, Edouard and Joulin, Armand and Fan, Angela},
  booktitle ={\url{https://arxiv.org/abs/1911.04944}},
  year={2019}
}

@article{laser3,
  title={Bitext Mining Using Distilled Sentence Representations for Low-Resource Languages},
  author={Heffernan, Kevin and {\c{C}}elebi, Onur and Schwenk, Holger},
  journal={arXiv preprint arXiv:2205.12654},
  year={2022}
}

@misc{sentenceT5:2021:arxiv,
      title={Sentence-T5: Scalable Sentence Encoders from Pre-trained Text-to-Text Models}, 
      author={Jianmo Ni and Gustavo Hernández Ábrego and Noah Constant and Ji Ma and Keith B. Hall and Daniel Cer and Yinfei Yang},
      year={2021},
      eprint={2108.08877},
      archivePrefix={arXiv},
      primaryClass={cs.CL},
      url={https://arxiv.org/abs/2108.08877}, 
}

@misc{multlinE5:2024:arxiv,
      title={Multilingual E5 Text Embeddings: A Technical Report}, 
      author={Liang Wang and Nan Yang and Xiaolong Huang and Linjun Yang and Rangan Majumder and Furu Wei},
      year={2024},
      eprint={2402.05672},
      archivePrefix={arXiv},
      primaryClass={cs.CL},
      url={https://arxiv.org/abs/2402.05672}, 
}

@misc{M3embed:2024:arxiv,
      title={M3-Embedding: Multi-Linguality, Multi-Functionality, Multi-Granularity
Text Embeddings Through Self-Knowledge Distillation},
      author={Jianlv Chen and Shitao Xiao and Peitian Zhang and KunLuo and Defu Lian and Zheng Liu},
      year={2024},
      eprint={2402.03216
      archivePrefix={arXiv},
      primaryClass={cs.CL},
      url={https://arxiv.org/pdf/2402.03216}, 
}}

@article{barrault2024largeConceptModelLCM,
  title={Large concept models: Language modeling in a sentence representation space},
  author={{LCM team} and The Barrault, Lo{\"\i}c and Duquenne, Paul-Ambroise and Elbayad, Maha and Kozhevnikov, Artyom and Alastruey, Belen and Andrews, Pierre and Coria, Mariano and Couairon, Guillaume and Costa-juss{\`a}, Marta R and Dale, David and others},
  journal={arXiv preprint arXiv:2412.08821},
  year={2024}
}

@article{duquenne2023sonar,
  title={SONAR: sentence-level multimodal and language-agnostic representations},
  author={Duquenne, Paul-Ambroise and Schwenk, Holger and Sagot, Beno{\^\i}t},
  journal={arXiv preprint arXiv:2308.11466},
  year={2023}
}

@article{bolya2025perceptionencoder,
  title={Perception encoder: The best visual embeddings are not at the output of the network},
  author={Bolya, Daniel and Huang, Po-Yao and Sun, Peize and Cho, Jang Hyun and Madotto, Andrea and Wei, Chen and Ma, Tengyu and Zhi, Jiale and Rajasegaran, Jathushan and Rasheed, Hanoona and others},
  journal={arXiv preprint arXiv:2504.13181},
  year={2025}
}

@article{cho2025perceptionlm,
  title={Perceptionlm: Open-access data and models for detailed visual understanding},
  author={Cho, Jang Hyun and Madotto, Andrea and Mavroudi, Effrosyni and Afouras, Triantafyllos and Nagarajan, Tushar and Maaz, Muhammad and Song, Yale and Ma, Tengyu and Hu, Shuming and Jain, Suyog and others},
  journal={arXiv preprint arXiv:2504.13180},
  year={2025}
}

@article{team2024chameleon,
  title={Chameleon: Mixed-modal early-fusion foundation models},
  author={{Chameleon Team}},
  journal={arXiv preprint arXiv:2405.09818},
  year={2024}
}

@inproceedings{chen2024internvl,
  title={Internvl: Scaling up vision foundation models and aligning for generic visual-linguistic tasks},
  author={Chen, Zhe and Wu, Jiannan and Wang, Wenhai and Su, Weijie and Chen, Guo and Xing, Sen and Zhong, Muyan and Zhang, Qinglong and Zhu, Xizhou and Lu, Lewei and others},
  booktitle={Proceedings of the IEEE/CVF conference on computer vision and pattern recognition},
  pages={24185--24198},
  year={2024}
}

@article{bai2025qwen2_5_vl,
  title={Qwen2. 5-vl technical report},
  author={Bai, Shuai and Chen, Keqin and Liu, Xuejing and Wang, Jialin and Ge, Wenbin and Song, Sibo and Dang, Kai and Wang, Peng and Wang, Shijie and Tang, Jun and others},
  journal={arXiv preprint arXiv:2502.13923},
  year={2025}
}

@article{wang2024qwen2vl,
  title={Qwen2-vl: Enhancing vision-language model's perception of the world at any resolution},
  author={Wang, Peng and Bai, Shuai and Tan, Sinan and Wang, Shijie and Fan, Zhihao and Bai, Jinze and Chen, Keqin and Liu, Xuejing and Wang, Jialin and Ge, Wenbin and others},
  journal={arXiv preprint arXiv:2409.12191},
  year={2024}
}

@article{touvron2023llama,
  title={Llama: Open and efficient foundation language models},
  author={Touvron, Hugo and Lavril, Thibaut and Izacard, Gautier and Martinet, Xavier and Lachaux, Marie-Anne and Lacroix, Timoth{\'e}e and Rozi{\`e}re, Baptiste and Goyal, Naman and Hambro, Eric and Azhar, Faisal and others},
  journal={arXiv preprint arXiv:2302.13971},
  year={2023}
}

@article{team2024qwen2,
  title={Qwen2 technical report},
  author={Team, Qwen},
  journal={arXiv preprint arXiv:2407.10671},
  year={2024}
}

@article{tschannen2025siglip2,
  title={Siglip 2: Multilingual vision-language encoders with improved semantic understanding, localization, and dense features},
  author={Tschannen, Michael and Gritsenko, Alexey and Wang, Xiao and Naeem, Muhammad Ferjad and Alabdulmohsin, Ibrahim and Parthasarathy, Nikhil and Evans, Talfan and Beyer, Lucas and Xia, Ye and Mustafa, Basil and others},
  journal={arXiv preprint arXiv:2502.14786},
  year={2025}
}

@inproceedings{wang2019vatex,
  title={Vatex: A large-scale, high-quality multilingual dataset for video-and-language research},
  author={Wang, Xin and Wu, Jiawei and Chen, Junkun and Li, Lei and Wang, Yuan-Fang and Wang, William Yang},
  booktitle={Proceedings of the IEEE/CVF international conference on computer vision},
  pages={4581--4591},
  year={2019}
}

@article{wang2024-dream-1k,
  title={Tarsier: Recipes for training and evaluating large video description models},
  author={Wang, Jiawei and Yuan, Liping and Zhang, Yuchen and Sun, Haomiao},
  journal={arXiv preprint arXiv:2407.00634},
  year={2024}
}

@inproceedings{zhang2025pretrainedImageTextModelsVideoCaptioner,
  title={Pretrained Image-Text Models are Secretly Video Captioners},
  author={Zhang, Chunhui and Jian, Yiren and Ouyang, Zhongyu and Vosoughi, Soroush},
  booktitle={Proceedings of the 2025 Conference of the Nations of the Americas Chapter of the Association for Computational Linguistics: Human Language Technologies (Volume 2: Short Papers)},
  pages={292--305},
  year={2025}
}

@inproceedings{NEURIPS2021_8466f9ac-sonar-speech-mining,
 author = {Duquenne, Paul-Ambroise and Gong, Hongyu and Schwenk, Holger},
 booktitle = {Advances in Neural Information Processing Systems},
 editor = {M. Ranzato and A. Beygelzimer and Y. Dauphin and P.S. Liang and J. Wortman Vaughan},
 pages = {15748--15761},
 publisher = {Curran Associates, Inc.},
 title = {Multimodal and Multilingual Embeddings for Large-Scale Speech Mining},
 url = {https://proceedings.neurips.cc/paper_files/paper/2021/file/8466f9ace6a9acbe71f75762ffc890f1-Paper.pdf},
 volume = {34},
 year = {2021}
}

@inproceedings{radford2021learningCLIP,
  title={Learning transferable visual models from natural language supervision},
  author={Radford, Alec and Kim, Jong Wook and Hallacy, Chris and Ramesh, Aditya and Goh, Gabriel and Agarwal, Sandhini and Sastry, Girish and Askell, Amanda and Mishkin, Pamela and Clark, Jack and others},
  booktitle={International conference on machine learning},
  pages={8748--8763},
  year={2021},
  organization={PmLR}
}

@inproceedings{jia2021scalingAlign,
  title={Scaling up visual and vision-language representation learning with noisy text supervision},
  author={Jia, Chao and Yang, Yinfei and Xia, Ye and Chen, Yi-Ting and Parekh, Zarana and Pham, Hieu and Le, Quoc and Sung, Yun-Hsuan and Li, Zhen and Duerig, Tom},
  booktitle={International conference on machine learning},
  pages={4904--4916},
  year={2021},
  organization={PMLR}
}

@inproceedings{lei2021less,
  title={Less is More: ClipBERT for Video-and-Language Learning via Sparse Sampling},
  author={Lei, Jie and Berg, Tamara L. and Bansal, Mohit},
  booktitle={CVPR},
  year={2021}
}

@inproceedings{xu2021videoclip,
  title={VideoCLIP: Contrastive Pre-training for Zero-shot Video-Text Understanding},
  author={Xu, Hu and Ghosh, Gargi and Huang, Po-Yao and Zhang, Dahuang and Ainslie, Joshua and others},
  booktitle={EMNLP},
  year={2021}
}

@inproceedings{wang2022allinone,
  title={All-in-One: Exploring Unified Video-Language Pre-training},
  author={Wang, Minghang and Ma, Zeqiu and Xie, Yuanjun and Li, Xuewen and others},
  booktitle={CVPR},
  year={2022}
}

@inproceedings{yuan2021florence,
  title={Florence: A New Foundation Model for Computer Vision},
  author={Yuan, Lu and Chen, Dongdong and Chen, Yi-Ling and Codella, Noel and Dai, Xiyang and and others},
  booktitle={CVPR},
  year={2021}
}

@article{chung2018unsupervised,
  title={Unsupervised cross-modal alignment of speech and text embedding spaces},
  author={Chung, Yu-An and Weng, Wei-Hung and Tong, Schrasing and Glass, James},
  journal={Advances in neural information processing systems},
  volume={31},
  year={2018}
}

@article{li2023m3it,
  title={M3IT: A Large-Scale Dataset towards Multi-Modal Multilingual Instruction Tuning},
  author={Li, Lei and Yin, Yuwei and Li, Shicheng and Chen, Liang and Wang, Peiyi and Ren, Shuhuai and Li, Mukai and Yang, Yazheng and Xu, Jingjing and Sun, Xu and others},
  journal={arXiv preprint arXiv:2306.04387},
  year={2023}
}

@article{assran2025vJEPA2,
  title={V-jepa 2: Self-supervised video models enable understanding, prediction and planning},
  author={Assran, Mido and Bardes, Adrien and Fan, David and Garrido, Quentin and Howes, Russell and Muckley, Matthew and Rizvi, Ammar and Roberts, Claire and Sinha, Koustuv and Zholus, Artem and others},
  journal={arXiv preprint arXiv:2506.09985},
  year={2025}
}

@article{bardes2023vJEPA,
  title={V-jepa: Latent video prediction for visual representation learning},
  author={Bardes, Adrien and Garrido, Quentin and Ponce, Jean and Chen, Xinlei and Rabbat, Michael and LeCun, Yann and Assran, Mido and Ballas, Nicolas},
  year={2023},
  journal = {submitted to ICLR'25}
}

@article{simeoni2025dinov3,
  title={Dinov3},
  author={Sim{\'e}oni, Oriane and Vo, Huy V and Seitzer, Maximilian and Baldassarre, Federico and Oquab, Maxime and Jose, Cijo and Khalidov, Vasil and Szafraniec, Marc and Yi, Seungeun and Ramamonjisoa, Micha{\"e}l and others},
  journal={arXiv preprint arXiv:2508.10104},
  year={2025}
}

@article{oquab2023dinov2,
  title={Dinov2: Learning robust visual features without supervision},
  author={Oquab, Maxime and Darcet, Timoth{\'e}e and Moutakanni, Th{\'e}o and Vo, Huy and Szafraniec, Marc and Khalidov, Vasil and Fernandez, Pierre and Haziza, Daniel and Massa, Francisco and El-Nouby, Alaaeldin and others},
  journal={arXiv preprint arXiv:2304.07193},
  year={2023}
}

@article{karras2022elucidating,
  title={Elucidating the design space of diffusion-based generative models},
  author={Karras, Tero and Aittala, Miika and Aila, Timo and Laine, Samuli},
  journal={Advances in neural information processing systems},
  volume={35},
  pages={26565--26577},
  year={2022}
}

@article{kuznetsova2020openImages,
  title={The open images dataset v4: Unified image classification, object detection, and visual relationship detection at scale},
  author={Kuznetsova, Alina and Rom, Hassan and Alldrin, Neil and Uijlings, Jasper and Krasin, Ivan and Pont-Tuset, Jordi and Kamali, Shahab and Popov, Stefan and Malloci, Matteo and Kolesnikov, Alexander and others},
  journal={International journal of computer vision},
  volume={128},
  number={7},
  pages={1956--1981},
  year={2020},
  publisher={Springer}
}

@inproceedings{kirillov2023segmentAnything,
  title={Segment anything},
  author={Kirillov, Alexander and Mintun, Eric and Ravi, Nikhila and Mao, Hanzi and Rolland, Chloe and Gustafson, Laura and Xiao, Tete and Whitehead, Spencer and Berg, Alexander C and Lo, Wan-Yen and others},
  booktitle={Proceedings of the IEEE/CVF international conference on computer vision},
  pages={4015--4026},
  year={2023}
}

@inproceedings{shohan2024xl-multilingualROUGE,
  title={XL-HeadTags: Leveraging Multimodal Retrieval Augmentation for the Multilingual Generation of News Headlines and Tags},
  author={Shohan, Faisal Tareque and Nayeem, Mir Tafseer and Islam, Samsul and Akash, Abu Ubaida and Joty, Shafiq},
  booktitle={ACL (Findings)},
  year={2024}
}

@article{dubey2024llama3,
  title={The llama 3 herd of models},
  author={Dubey, Abhimanyu and Jauhri, Abhinav and Pandey, Abhinav and Kadian, Abhishek and Al-Dahle, Ahmad and Letman, Aiesha and Mathur, Akhil and Schelten, Alan and Yang, Amy and Fan, Angela and others},
  journal={arXiv e-prints},
  pages={arXiv--2407},
  year={2024}
}

@article{yang2025languageAlignmentFixedTextEncoder,
  title={Language-Image Alignment with Fixed Text Encoders},
  author={Yang, Jingfeng and Wu, Ziyang and Zhao, Yue and Ma, Yi},
  journal={arXiv preprint arXiv:2506.04209},
  year={2025}
}

@inproceedings{brokowski-ssca,
  title={SSCA: SigLIP-2 Sonar Concept Alignment},
  author={Brokowski, Trevor and Sallinen, Alexandre and Hartley, Mary-Anne},
  booktitle={Second Workshop on Visual Concepts},
  year = {2025}
}

@inproceedings{stone2025learningCompositionLLMCLIP,
  title={Learning visual composition through improved semantic guidance},
  author={Stone, Austin and Soltau, Hagen and Geirhos, Robert and Yi, Xi and Xia, Ye and Cao, Bingyi and Chen, Kaifeng and Ogale, Abhijit and Shlens, Jonathon},
  booktitle={Proceedings of the Computer Vision and Pattern Recognition Conference},
  pages={3740--3750},
  year={2025}
}

@article{huang2024modularSentenceEncoderLanugageSpecialization,
  title={Modular sentence encoders: Separating language specialization from cross-lingual alignment},
  author={Huang, Yongxin and Wang, Kexin and Glava{\v{s}}, Goran and Gurevych, Iryna},
  journal={arXiv preprint arXiv:2407.14878},
  year={2024}
}

@article{du2024cosyvoice,
  title={Cosyvoice: A scalable multilingual zero-shot text-to-speech synthesizer based on supervised semantic tokens},
  author={Du, Zhihao and Chen, Qian and Zhang, Shiliang and Hu, Kai and Lu, Heng and Yang, Yexin and Hu, Hangrui and Zheng, Siqi and Gu, Yue and Ma, Ziyang and others},
  journal={arXiv preprint arXiv:2407.05407},
  year={2024}
}

@article{laperriere2024dual,
  title={A dual task learning approach to fine-tune a multilingual semantic speech encoder for Spoken Language Understanding},
  author={Laperri{\`e}re, Ga{\"e}lle and Ghannay, Sahar and Jabaian, Bassam and Est{\`e}ve, Yannick},
  journal={arXiv preprint arXiv:2406.12141},
  year={2024}
}

@article{oord2018representation,
  title={Representation learning with contrastive predictive coding},
  author={Oord, Aaron van den and Li, Yazhe and Vinyals, Oriol},
  journal={arXiv preprint arXiv:1807.03748},
  year={2018}
}

\newpage

\clearpage
\newpage
\beginappendix

\section{Contrastive Loss for Aligning \PE and \sonar}
\label{appendix:contrastive-loss}
We have also explored the use of a contrastive loss in addition to the MSE loss for aligning the \PE to \sonar. Specifically, given a mini-batch of $B$ paired samples $\{(V_i, T_i)\}_{i=1}^B$, we aim to not only minimize the distance between matched pairs $(f_\theta(V_i), g(T_i))$ but also push apart mismatched pairs.
We define the contrastive loss as:
\begin{equation}
\mathcal{L}_{\text{con}} = - \frac{1}{B} \sum_{i=1}^B \log \frac{\exp \big( \text{sim}(f_\theta(V_i), g(T_i)) / \tau \big)}{\sum_{j=1}^B \exp \big( \text{sim}(f_\theta(V_i), g(T_j)) / \tau \big)},
\end{equation}
where $\text{sim}(\cdot, \cdot)$ denotes cosine similarity and $\tau$ is a temperature parameter.
The final loss is then a weighted combination of the MSE alignment loss and the contrastive loss:
\begin{equation}
\mathcal{L} = \mathcal{L}_{\text{align}} + \lambda \mathcal{L}_{\text{con}},
\end{equation}
where $\lambda$ controls the strength of the contrastive term.
However, in our preliminary experiments, adding the contrastive component did not yield a significant improvement over the MSE-only objective (\autoref{tab:contrastive-ablation}) in captioning performance, while it leads to gains in retrieval performance. However, since our downstream usage of \vsonar for \vlcm is closer to generation task, we choose the MSE-only loss as the final loss.

\begin{table}[h!]
\centering
\begin{tabular}{lccccc}
\toprule
 & \multicolumn{3}{c}{\textbf{Captioning}} & \multicolumn{2}{c}{\textbf{Retrieval}} \\
\cmidrule(lr){2-4} \cmidrule(lr){5-6}
 & \bleu & R-L & BS-F1 & R@1 & MRR \\
\midrule
MSE-only      & \textbf{38.9} & \textbf{37.8} & \textbf{44.9} & 49.0 & 60.3 \\
MSE + Contrastive & 38.6 & 37.5 & 44.5 & \textbf{52.4} & \textbf{63.7} \\
\bottomrule
\end{tabular}
\caption{Ablation study on using only MSE loss vs. adding a contrastive loss. Results are reported on the PE-Video benchmark for captioning and retrieval with a single MLP layer as the connector in \vsonar.}
\label{tab:contrastive-ablation}
\end{table}

To further explain the trade-off between retrieval and captioning performance in \autoref{tab:contrastive-ablation}, we hypotheses that contrastive training can move vision vectors off the exact SONAR manifold. It only enforces relative ordering via cosine margins, and produces embeddings whose norms or local covariance differ from those the SONAR decoder were trained on. That small manifold shift can degrade generative reconstruction even if retrieval improves.

We analyze the statistics of \vsonar embeddings in \autoref{tab:contrastive-statistics-analysis}. It is a clear observation that \vsonar trained with the contrastive loss has a more separate embeddings, evidenced by its higher norm, trace and volume. It shows that contrastive loss pushes the vSONAR embeddings to be a more expanded distribution compared to the MSE-only (higher values in Norm, Covariance trace, and Volume). However, their poor alignment consistencies in both MSE and cosine similarity, where we calculate the correlation between vision and text similarity ranking for each sample, suggests contrastive loss breaks the local covariance structure and may break the alignment with SONAR manifold.

\begin{table}[h!]
\centering
\begin{tabular}{llllll}
\toprule
                & \textbf{V. Norm}       & \textbf{V. Trace}      & \textbf{V. Volume}            & \textbf{AC (Cosine)}   & \textbf{AC (MSE) }     \\ \midrule
MSE             & 1.22          & 0.48          & -10007.16         & \textbf{0.41} & \textbf{0.32} \\
MSE+Contrastive & \textbf{1.30} & \textbf{1.74} & \textbf{-8107.94} & 0.31          & 0.13   \\
\bottomrule
\end{tabular}
\caption{Additional statistics for \vsonar representations on PE-Video. We report the norm (V.Norm), trace (V.Trace), Volume (V.Volume), alignment consistency measured in Cosine Similarity (AC Cosine) and MSE (AC MSE) for video embeddings predicted by \vsonar.}
\label{tab:contrastive-statistics-analysis}
\end{table}

\section{Dataset Statistics}
\label{appendix:dataset-stat}

\autoref{tab:dataset_stats} summarizes the datasets used in our three-stage training pipeline for alignment. The PLM-Image datasets (SA1B and OpenImages) provide large-scale image–caption pairs, which are particularly valuable for improving grounding and linguistic richness. The PLM-Video-Auto-YT1B dataset contributes video–text pairs with an average duration of 22.75 seconds, enabling the model to capture temporal dynamics in multimodal content. Finally, the PE-Video dataset provides carefully curated human-annotated video–caption pairs with moderate length, serving as a higher-quality supervision signal in later stages. Together, these datasets balance scale and quality, ensuring both broad coverage and precise alignment across modalities.

\begin{table}[h]
\centering
\resizebox{\linewidth}{!}{
\begin{tabular}{lcccc}
\toprule
\textbf{Dataset} & \textbf{\#Samples} & \textbf{Duration (s)} & \textbf{Caption Length (sent.)} & \textbf{Caption Length (words)} \\
\midrule
PLM-Image-Auto-SA1B         & 7.99M  & --    & 10.7 & 181.8 \\
PLM-Image-Auto-OpenImages   & 1.37M  & --    & 7.9  & 132.1 \\
PLM-Video-Auto-YT1B         & 2.14M  & 22.8 & 2.3  & 95.5  \\
PE-Video                    & 118K   & 16.7  & 4.4  & 51.4  \\
\bottomrule
\end{tabular}
}
\caption{Statistics of the datasets used in 3-stage training for alignment. We report the number of samples, average video duration (if applicable), and average caption length in sentences and words.}
\label{tab:dataset_stats}
\end{table}

\section{Implementations}
\label{appendix:implementation-details}
\paragraph{\vsonar Architecture} 
We build our model on top of the Perception Encoder Vision Transformer backbone \texttt{PE-Core-G14-448}\footnote{https://huggingface.co/facebook/PE-Core-G14-448}. The encoder processes RGB images at a resolution of 448×448 pixels, splitting them into 14×14 patches and yielding 1024 patches per frame. The vision tower consists of 50 transformer layers, each with a hidden width of 1024, 16 attention heads, and a 4096-dimensional feed-forward network, resulting in approximately 1.9B parameters. For video inputs, we uniformly sample 8 frames and extract frame-level embeddings of 1536 dimensions from the encoder, which are subsequently projected into a 1024-dimensional SONAR embedding space. To bridge perception features with the target space, we attach a lightweight connector, where weights are optionally initialized from a Gaussian distribution $(\mu = 0, \sigma = 1e-5)$ with zero biases for stability. To capture temporal dynamics, the connector augments encoder outputs with sinusoidal positional encodings and applies a temporal multi-head self-attention module (8 heads, dropout 0.1) across frames, combined with residual connections. The resulting sequence is aggregated using attention-based pooling, where a learnable CLS token attends over the frame embeddings via an 8-head attention module, though we also evaluate mean and max pooling variants. The final pooled representation (1536 dimensions) is then mapped to the 1024-dimensional SONAR space by a linear MLP layer.

\paragraph{\vsonar Training Details}
To stabilize training, the projector is initialized from a zero-mean Gaussian distribution with a small variance ($1e$–5), which mitigates gradient explosion when mapping from the high-dimensional \PE features to the target embedding space.
We employ a two-phase training recipe: in the first 2{,}000 steps, \PE is frozen while only the projector is optimized, allowing the projector to adapt without perturbing the pre-trained encoder.
Subsequently, both the projector and \PE are jointly optimized, using asynchronous learning rates: a higher rate ($1e$–4) for the projector to enable rapid adaptation, and a lower rate ($1e$–5) for \PE to preserve pre-trained knowledge.

We train the model using a three-stage curriculum. 
{Stage 1 (image captioning)} runs for 15 epochs with a batch size of 512, 
a base learning rate of $1\times10^{-5}$, and a connector learning rate of $1\times10^{-4}$, 
with 4000 warmup steps applied to the connector. 
{Stage 2 (synthetic video captioning data)} runs for 10 epochs with an effective batch size of 128, 
a learning rate of $1\times10^{-5}$, and a connector learning rate of $1\times10^{-4}$ with 2000 warmup steps. 
{Stage 3 (manually verified video captioning data)} adopts the same settings as Stage 2. 
Across all stages, we optimize with AdamW, cosine learning rate decay, and a 500-step linear warmup schedule. 
We fix the random seed to 42 and evaluate on 2000 validation samples per stage. 
Training is distributed with Fully Sharded Data Parallel (FSDP) across 64 Nvidia A100-80G GPUs using \texttt{bfloat16} precision, 
gradient accumulation for memory efficiency, and early stopping with a patience of 3 epochs, 
checkpointing the best validation model.

\paragraph{\vlcm Training Details}

For training the \vlcm, we adopt the \lcm two-tower architecture \citep{barrault2024largeConceptModelLCM} with the diffusion-based next-sentence fine-tuning objective. The optimizer is AdamW with $\epsilon=10^{-6}$, weight decay of $0.01$, gradient clipping at $25.0$, and learning rate $3\times10^{-5}$ scheduled with cosine decay, warmed up over the first $300$ steps, and annealed to a final learning rate of $10^{-6}$. Training is run for a maximum of $10{,}000$ steps with batch sizes dynamically determined up to $7168$ latent embeddings, using gradient accumulation set to $1$. Checkpoints are saved every $1000$ steps, and we select the best performance according to the validation performance. 
The criterion incorporates a conditional guidance probability of $0.15$ as used in \lcm, and loss is reduced with a summation loss function. Training uses Fully Sharded Data Parallel (FSDP) with bf16 precision for efficiency. Data loading is set to uniformly sampled from all training set in M3IT, length-ordered batching without packing. Experiments are conducted on $1$ node with $8$ A100 GPUs (80GB).

\section{\sonar vs \sonartwo}
\label{appendix:sonar1-vs-sonar2}

We also report the comparison in the retrieval performance in \autoref{tab:sonar1-vs-sonar2-retrieval}. While \sonartwo generally holds an advantage, SONAR1 remains highly competitive (e.g., R@1 of 64.9 on PVD-Bench compared to PECoreG).

\begin{table}[h]
\centering
\begin{tabular}{llllll}
\toprule
                           &        & \textbf{R@1} & \textbf{R@5} & \textbf{R@10} & \textbf{MRR} \\ \midrule
\multirow{2}{*}{\textbf{PVD-Bench}} & SONAR1 & 0.649        & 0.843        & 0.895         & 0.737        \\
                           & \sonartwo & 0.730        & 0.898        & 0.938         & 0.805        \\ \midrule
\multirow{2}{*}{\textbf{DREAM-1K}}  & SONAR1 & 0.536        & 0.759        & 0.838         & 0.638        \\
                           & \sonartwo & 0.633        & 0.841        & 0.890         & 0.725        \\ \midrule
\multirow{2}{*}{\textbf{VATEX}}     & SONAR1 & 0.119        & 0.263        & 0.350         & 0.195        \\
                           & \sonartwo & 0.408        & 0.686        & 0.789         & 0.536        \\ \bottomrule
\end{tabular}
\caption{Comparison for \vsonar trained with SONAR version 1 and 2 embedding space on three text-to-video retrieval datasets.}
\label{tab:sonar1-vs-sonar2-retrieval}
\end{table}

\section{Detailed Multilingual Evaluation}
\label{appendix:detailed-m3it-multilinguaul}

\begin{figure}
    \centering
    \includegraphics[width=0.9\linewidth]{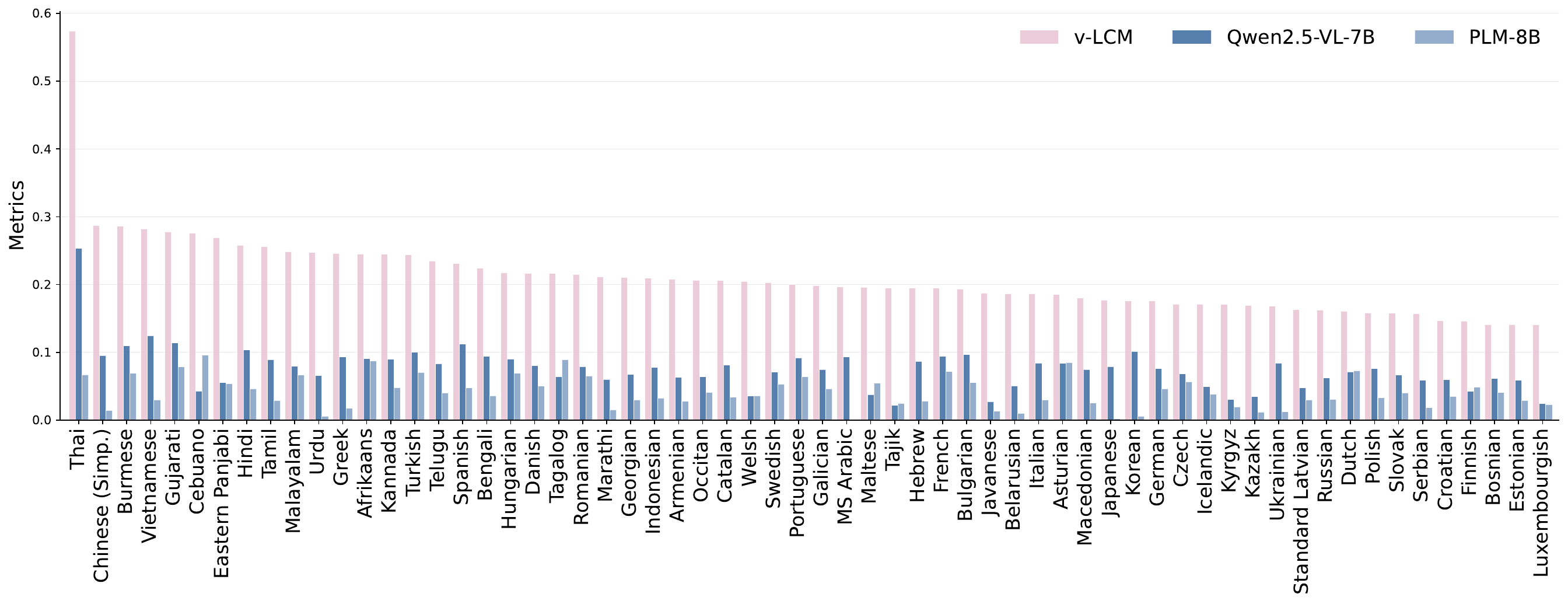}
    \caption{M3IT evaluation on 61 languages for MSRVTT.}
    \label{fig:multilin2_msrvtt}
\end{figure}

\begin{figure}
    \centering
    \includegraphics[width=0.9\linewidth]{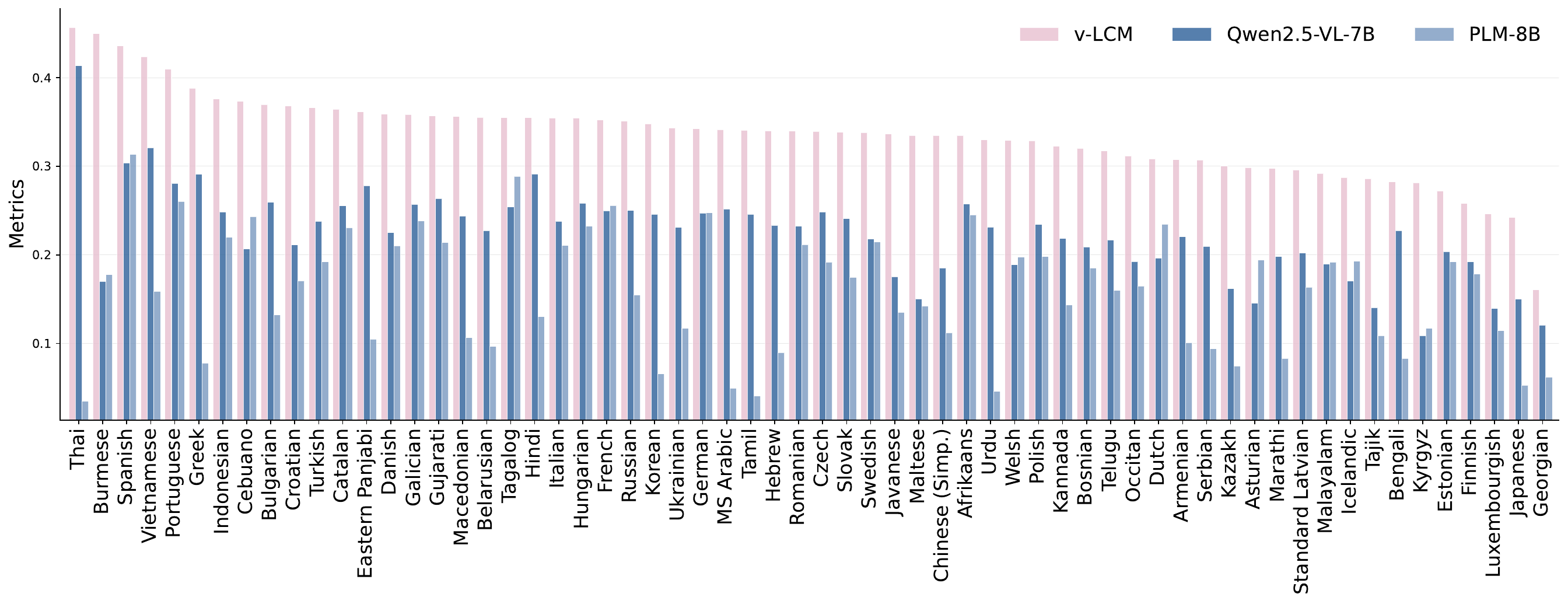}
    \caption{M3IT evaluation on 61 languages for MSRVTT-QA.}
    \label{fig:multlin3_msrvtt_qa}
\end{figure}

\begin{figure}
    \centering
    \includegraphics[width=0.9\linewidth]{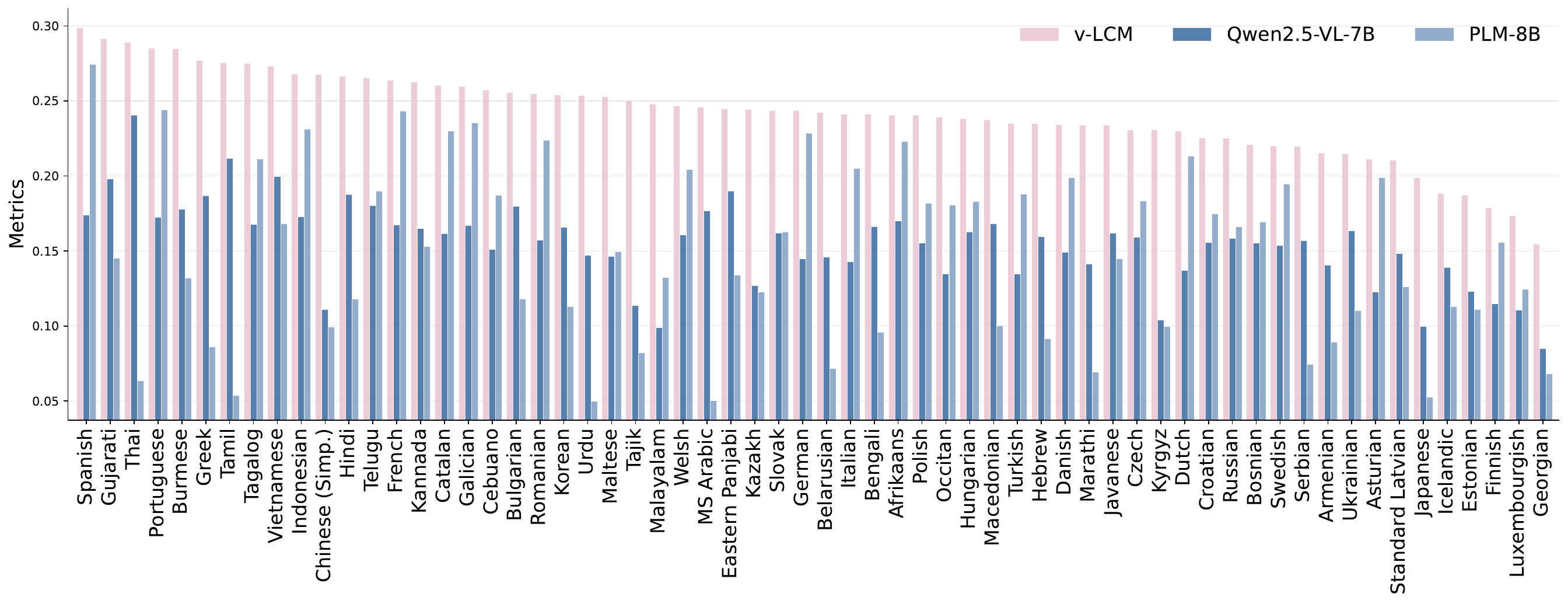}
    \caption{M3IT evaluation on 61 languages for OKVQA.}
    \label{fig:multlin4_okvqa}
\end{figure}

\begin{figure}
    \centering
    \includegraphics[width=0.9\linewidth]{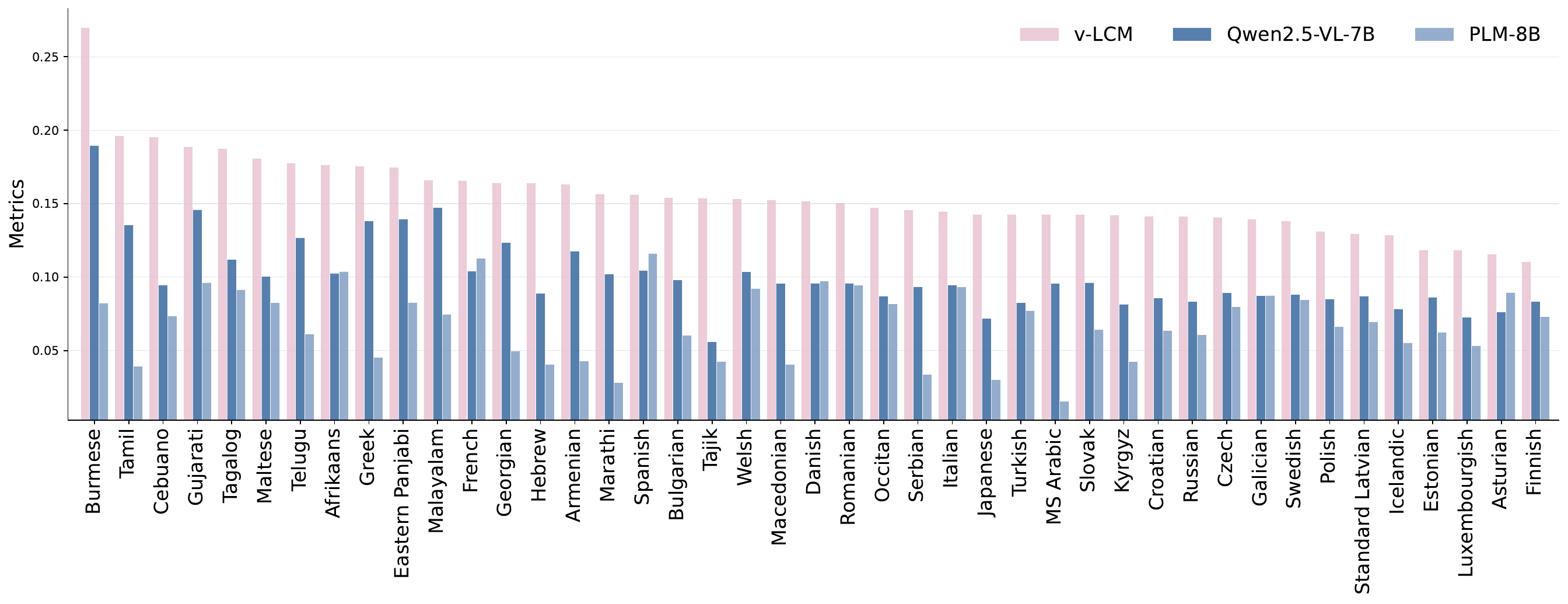}
    \caption{M3IT evaluation on 61 languages for VIST.}
    \label{fig:multlin5_vist}
\end{figure}

\begin{figure}
    \centering
    \includegraphics[width=0.9\linewidth]{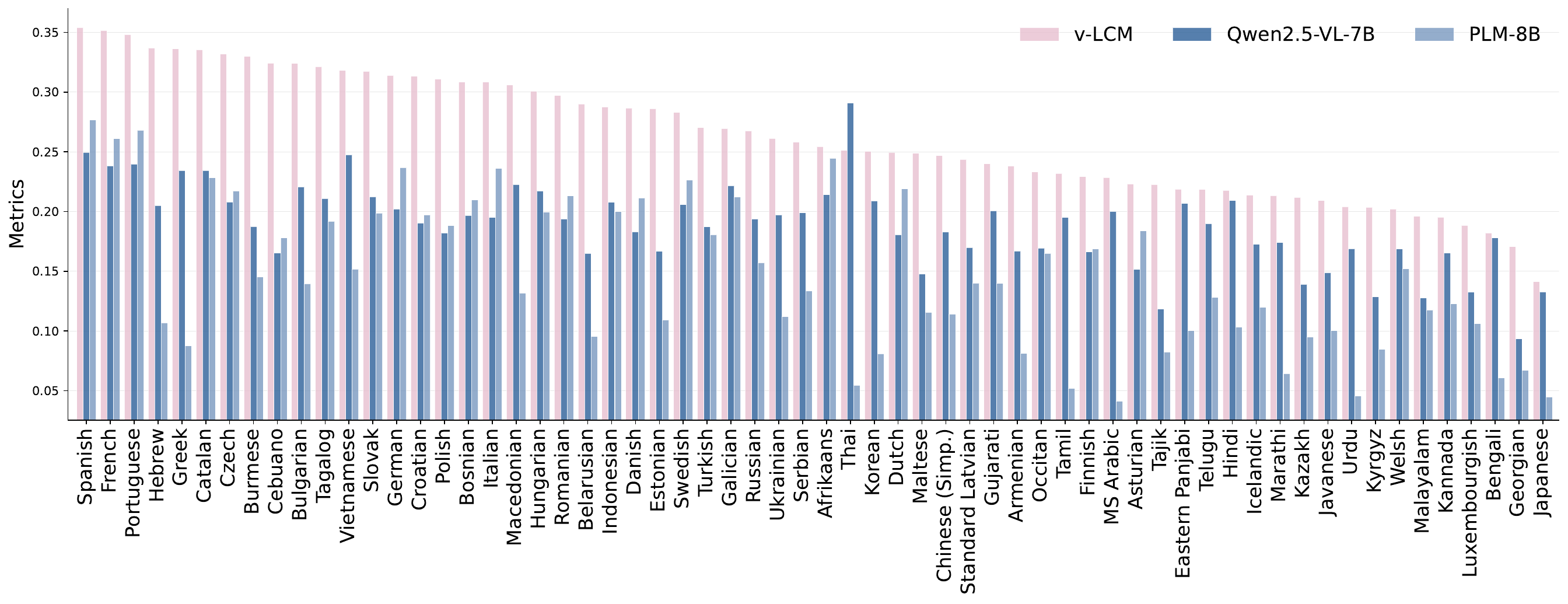}
    \caption{M3IT evaluation on 61 languages for VQA-V2.}
    \label{fig:multlin6_vqa}
\end{figure}

\begin{figure}
    \centering
    \includegraphics[width=0.9\linewidth]{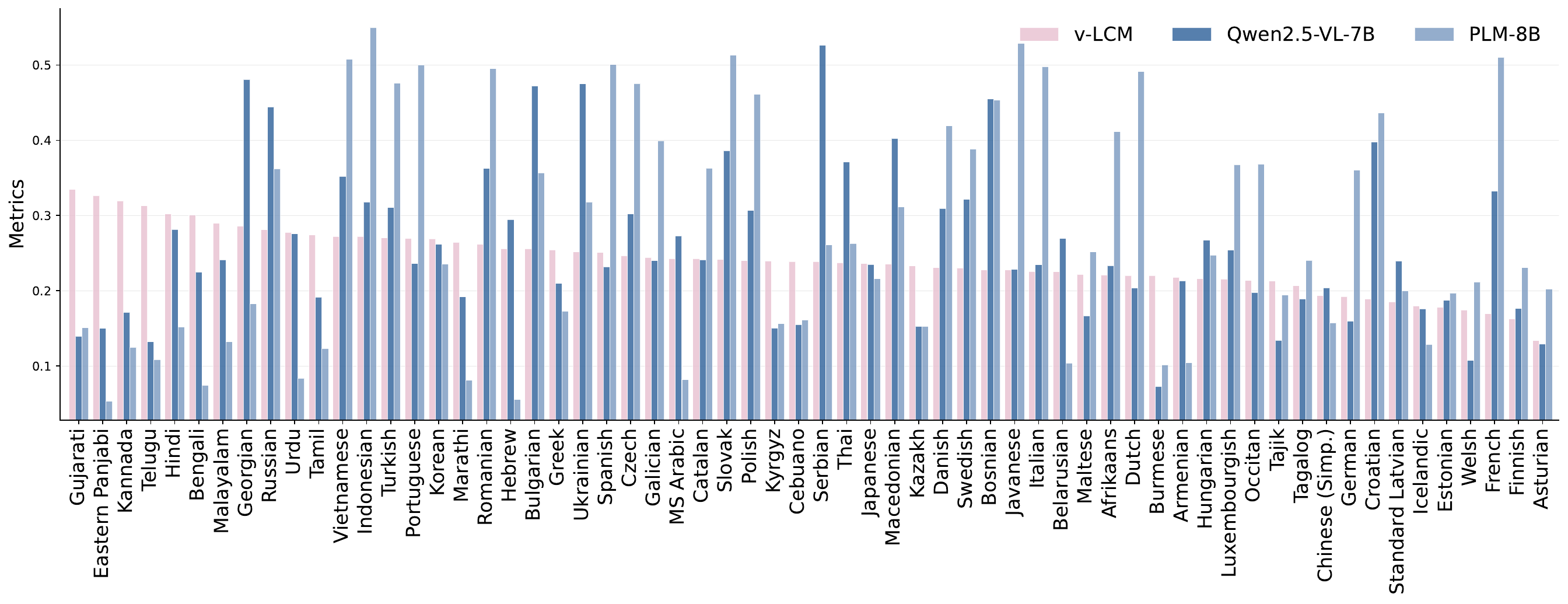}
    \caption{M3IT evaluation on 61 languages for image captioning.}
    \label{fig:multlin1_imagenet}
\end{figure}

We present the results of our multilingual evaluation across all supported datasets in this section. Specifically, we test all languages covered by \sonar and M3IT, reporting \rouge scores as the primary metric. \autoref{fig:multlin1_imagenet} shows the results for the image captioning task on ImageNet, while \autoref{fig:multilin2_msrvtt} and \autoref{fig:multlin3_msrvtt_qa} report results for video captioning and video QA on MSR-VTT, respectively. \autoref{fig:multlin4_okvqa} presents results on OKVQA, \autoref{fig:multlin5_vist} illustrates performance on story generation in VIST, and \autoref{fig:multlin6_vqa} shows results for image QA on VQA-v2. With the exception of ImageNet, likely due to its widespread use and extensive coverage in existing VLMs, our model consistently outperforms baselines across all tested languages, with the only exception being Thai in VQA-v2.

\section{Visualization for \vsonar's Latent Space}

\begin{figure*}[h]  
    \centering
    \begin{subfigure}[b]{0.48\textwidth}
        \centering
        \includegraphics[width=\textwidth]{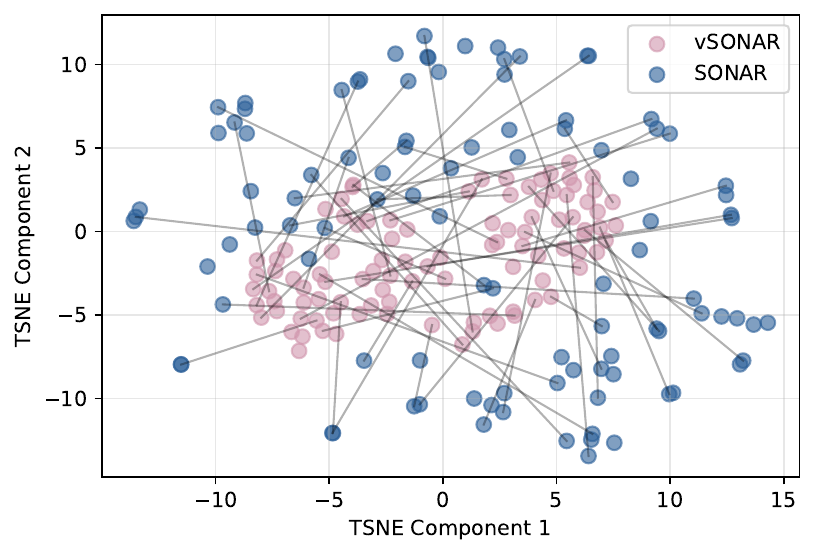}
        \caption{No alignment.}
        \label{fig:subfig1}
    \end{subfigure}
    \hfill
    \begin{subfigure}[b]{0.48\textwidth}
        \centering
        \includegraphics[width=\textwidth]{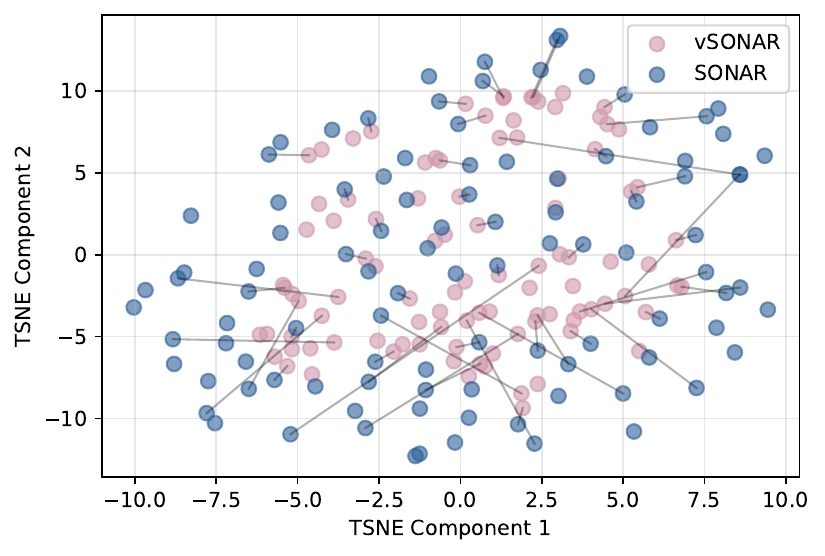}
        \caption{After stage 1.}
        \label{fig:subfig2}
    \end{subfigure}

    \begin{subfigure}[b]{0.48\textwidth}
        \centering
        \includegraphics[width=\textwidth]{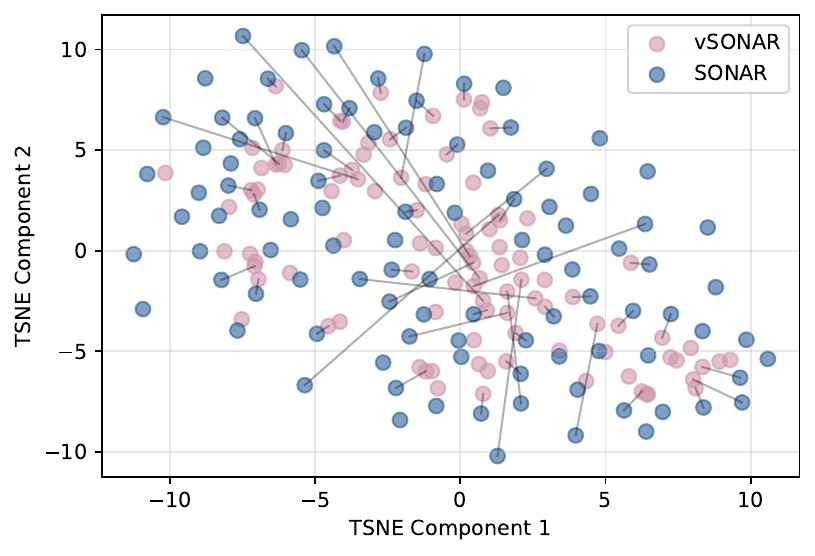}
        \caption{After stage 2.}
        \label{fig:subfig3}
    \end{subfigure}
    \hfill
    \begin{subfigure}[b]{0.48\textwidth}
        \centering
        \includegraphics[width=\textwidth]{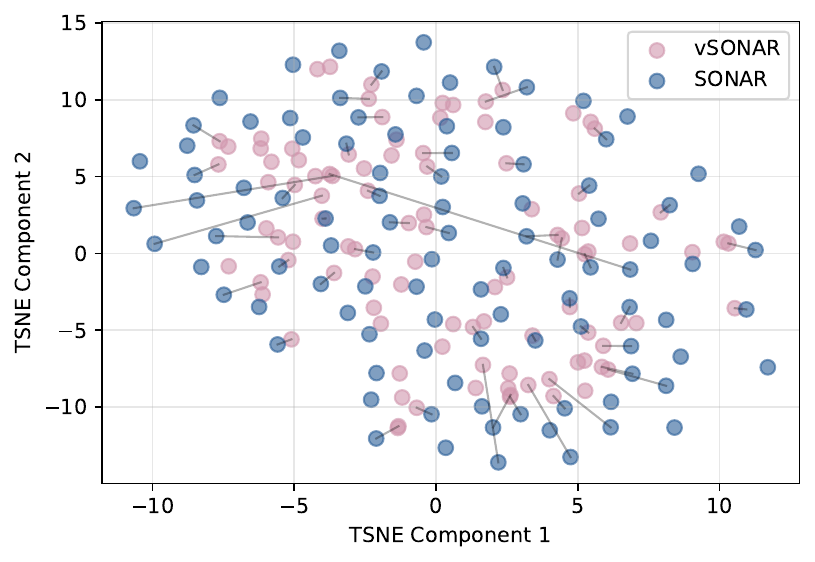}
        \caption{After stage 3.}
        \label{fig:subfig4}
    \end{subfigure}
    \caption{Visualization with t-SNE for \sonar and \vsonar embeddings after each stage of curriculum. \vsonar encodes the video, and \sonar encodes the caption. We randomly sample 200 samples from PE-Video's testing set for t-SNE, and explicitly plot the lines for connecting the paired video and caption for 50 samples.}
    \label{fig:visualization_latent_space_vsonar}
\end{figure*}

To qualitatively assess the effectiveness of our alignment, we visualize the latent spaces of video and SONAR embeddings before and after each stage of aligning \PE to \sonar using t-SNE (\autoref{fig:visualization_latent_space_vsonar}). 
After each stage of alignment, we observe a better clustering structure where video embeddings and their corresponding \sonar embeddings lie in the closer proximity, indicating that the alignment successfully reduces modality gaps in the latent space and supports a shared semantic representation across modalities, thereby validating the alignment strategy.

\section{Analysis in Cross-modal Drift for \vsonar and \vlcm}

In this section, we investigate the potential happening of the cross-modal drift in \vsonar and \vlcm's decoding process. We aim to answer the question whether the generated language by \sonar decoder and \vlcm faithfully represents the same semantic region of the embedding space in \vsonar. We conduct three analysis on PE-Video testing set to empirically prove that \vsonar and \vlcm suffered a minimum semantic drift during decoding when they handle embeddings from various modalities.

\begin{table}[h!]
\centering
\begin{tabular}{lllllll}
\toprule
                     & \multicolumn{2}{c}{\textbf{Semantic Sim.}} & \multicolumn{4}{c}{\textbf{Round-trip Retrieval}} \\ 
                     & Cosine               & Dist.               & R@1         & R@5        & R@10       & MRR       \\ \midrule
Groundtruth & 0.666                & 0.197               & 87.00\%     & 95.90\%    & 97.10\%    & 0.9084    \\
\sonar Decoder        & 0.689               & 0.175               & 82.50\%     & 97.00\%    & 98.70\%    & 0.8883    \\
\vlcm                  & 0.562               & 0.219               & 82.30\%     & 96.70\%    & 97.90\%    & 0.8867  \\ \bottomrule 
\end{tabular}
\caption{Left: semantic similarities and distances between video embedding and the ground-truth captions, generated captions from \sonar Decoder and \vlcm. Right: results for the round-trip retrieval ablation study where we use the ground-truth captions, \sonar Decoder and \vlcm's generated captions to retrieve the videos.}
\label{tab:semantic-drift-analysis}
\end{table}

\paragraph{Embedding-Level Semantic Fidelity.} For each video embedding $v$, we first compare its similarity and distance to: (i) ground-truth caption embedding $t_{gt}$, (ii) SONAR Decoder caption embedding ($t_{sonar}$, and (iii) LCM caption embedding $t_{lcm}$. This analysis provides a direct comparison for the generated captions from \sonar decoder and \vlcm with the ground-truth captions.

As indicated in \autoref{tab:semantic-drift-analysis}, we find that SONAR-decoded captions show nearly identical cosine similarity or distance compared to the groundtruth, indicating negligible cross-modal drift. vLCM captions show a slightly larger deviation; we attribute this to vLCM’s instruction-following training which introduces stylistic paraphrasing, rather than semantic drift.

\paragraph{Ablation with the Round-trip Retrieval.} We further conduct an ablation study with the round-trip retrieval where we use three groups of captions (Groundtruth, \vsonar Decoder, and \vlcm) as queries to retrieve the source videos in PE-Video. In \autoref{tab:contrastive-statistics-analysis}, we find that Captions decoded by SONAR or LCM retrieve the correct video with extremely high accuracy. Notably, LCM is within 0.2\% of SONAR on R@1. If cross-modal drift were substantial, retrieval accuracy would drop sharply; instead, it remains high, confirming its semantic preservation.

\begin{figure}
    \centering
    \includegraphics[width=0.9\linewidth]{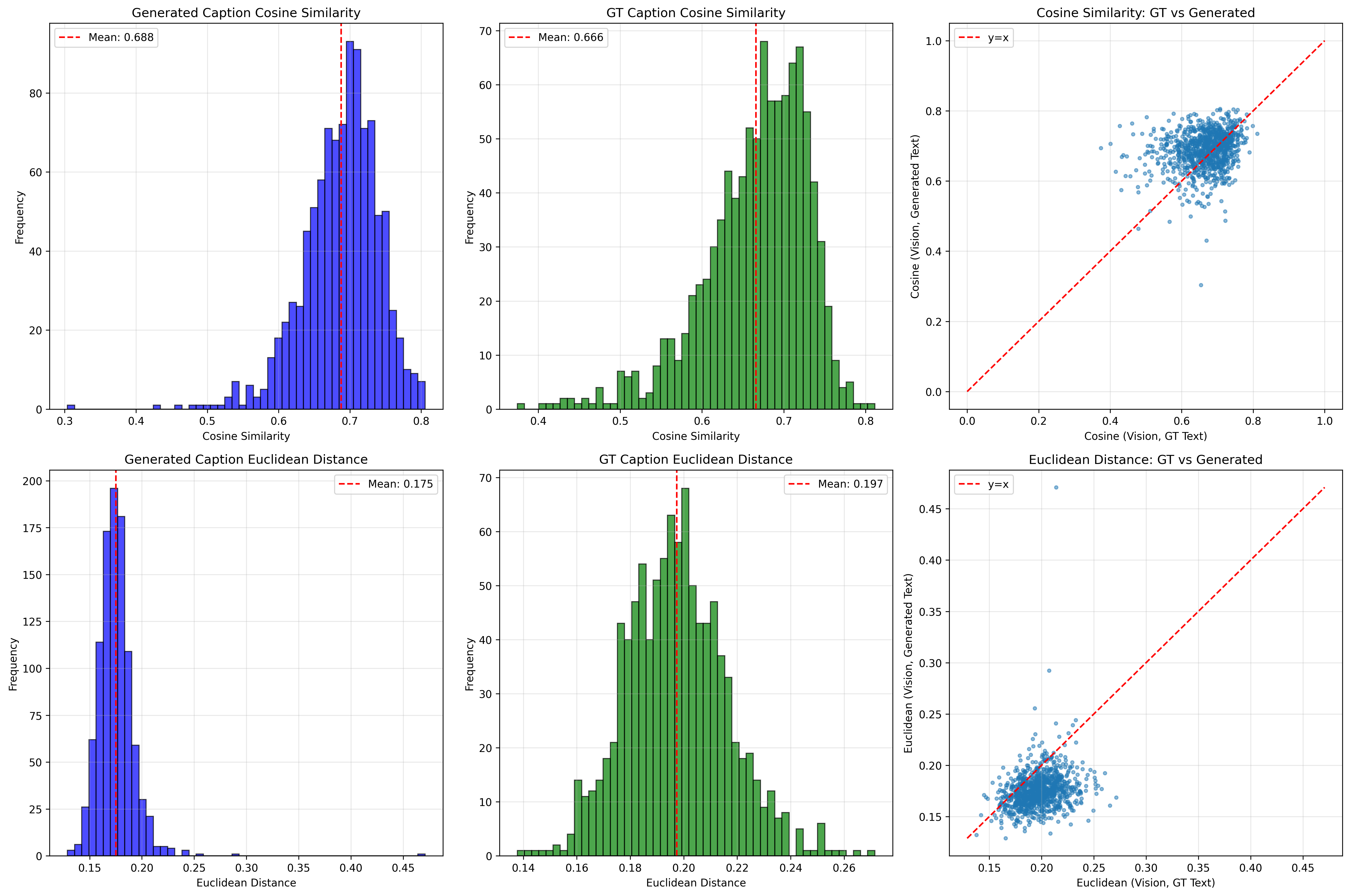}
    \caption{vSONAR's visualization for cross-modal semantic drift.}
    \label{fig:vsonar-cross-modal-drift}
\end{figure}

\begin{figure}
    \centering
    \includegraphics[width=0.9\linewidth]{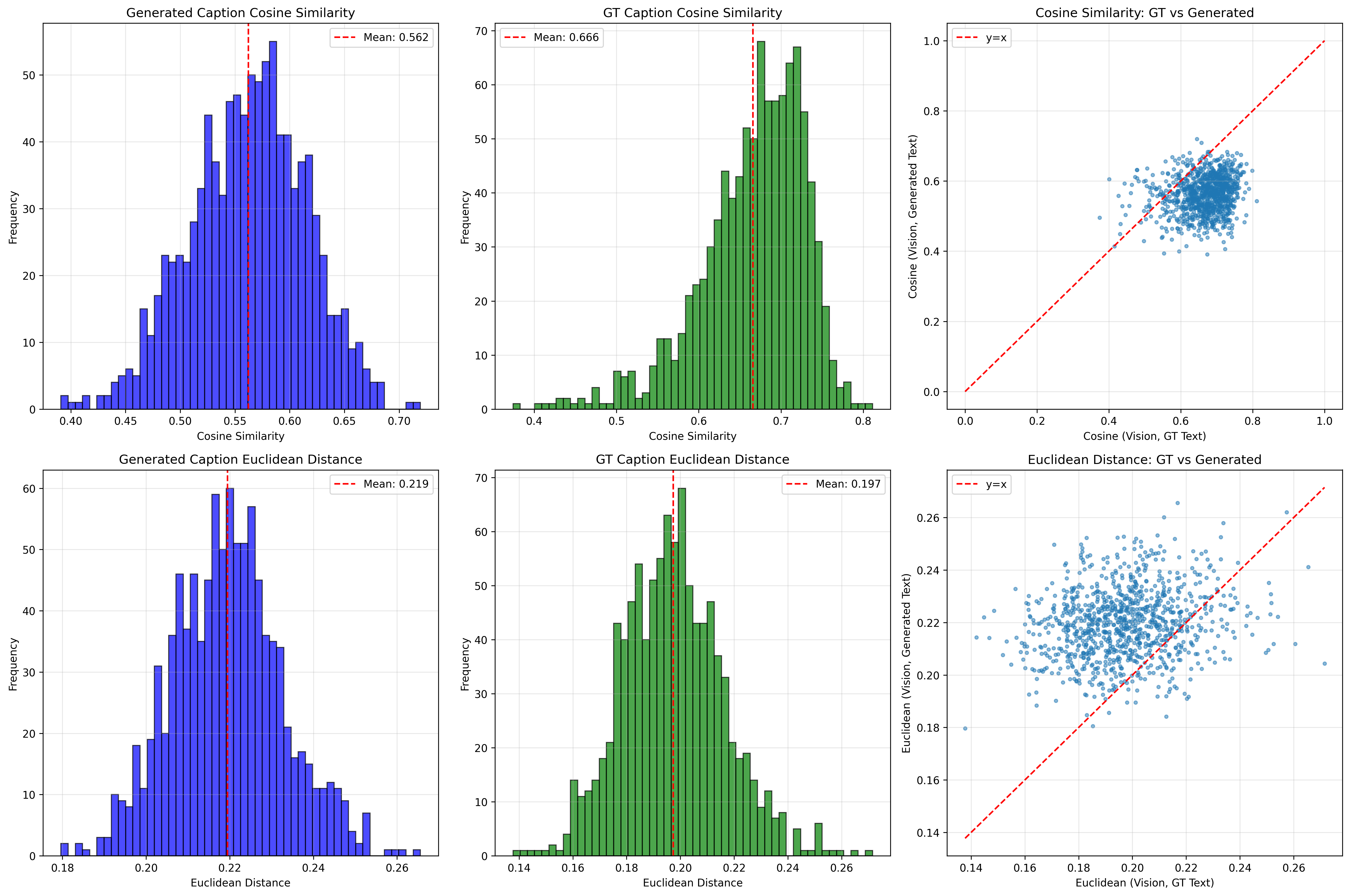}
    \caption{vLCM's visualization for cross-modal semantic drift.}
    \label{fig:vlcm-cross-modal-drift}
\end{figure}

\paragraph{Visualizing the Cross-modal Drift.} Finally, we plot the similarity between vision embeddings and the ground-truth caption embeddings versus the embeddings for \vsonar and \lcm captions in \autoref{fig:vlcm-cross-modal-drift} and \autoref{fig:vsonar-cross-modal-drift}. SONAR-decoded captions show nearly identical (or slightly better) cosine similarity/distance compared to ground truth, indicating negligible cross-modal drift. vLCM captions show a slightly larger deviation; we attribute this to vLCM’s instruction-following training which introduces stylistic paraphrasing, rather than semantic drift (verified in the next experiment). The points cluster also is generally around the $y=x$ line, directly showing no significant systematic semantic shift.

\section{Qualitative Case}
\label{appendix:qualitative_case}

\subsection{Image Qualitative Cases for \vlcm}

\begin{figure*}[t]
    \centering
    \adjustbox{max width=\textwidth}{
    \begin{tabular}{>{\centering\arraybackslash}m{0.25\textwidth} 
                    >{\arraybackslash}m{0.2\textwidth} 
                    >{\arraybackslash}m{0.5\textwidth}}
        \toprule
        \textbf{Vision Input} & \textbf{Prompt} & \textbf{Outputs} \\
        \midrule
        \includegraphics[width=\linewidth]{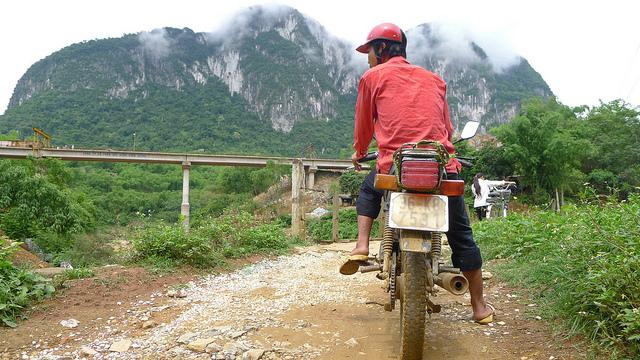} 
        & Write a short and informative description that highlights the main subjects and actions occurring in the given image.
        \texttt{<IMAGE>}
        & \textbf{\lcm}: \textit{``The video shows a man riding a red-colored bike on a red-colored motorbike riding on a black mountain, with a motorcycle and a rug, and a red man standing behind a mountain with a bridge and a grass and a roof.''} 
        
        \textbf{\vlcm}: \textit{``A man riding a red motorcycle on a mountain.''} \\
        \midrule
        \includegraphics[width=\linewidth]{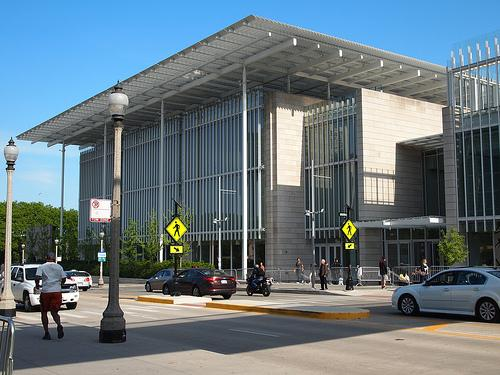} 
        & Compose a descriptive paragraph that brings the image to life through detailed storytelling, connecting the various visual elements into a unified narrative. \texttt{<IMAGE>} 
        & \textbf{\lcm}: \textit{``The vision begins with the bright video building standing in front of a broadly paved building, and a crowd of people walking on the street. As the sun shines over the building, the buildings begin to streak down the street. The vehicles move closer to the street, with the buildings creeping down the street. The vehicles are moving to the street, and there are people standing on the other side of the pavement. Eventually, the vision ends with the vehicles moving off onto the street, and the buildings slide down the street.''} 
        
        \textbf{\vlcm}: \textit{``A large building is walking near the corner of a street. There are white windows in front of the building. There is a large grey pole in front of the building. There are people walking on the street in front of the pole.''} \\

        \bottomrule
    \end{tabular}
    }
    \caption{Qualitative examples for image captioning task in M3IT benchmark. Each row shows the input image, the given prompt, and the generated model output.}
    \label{fig:vlm_qualitative1}
\end{figure*}

\begin{figure*}[t]
    \centering
    \adjustbox{max width=\textwidth}{
    \begin{tabular}{>{\centering\arraybackslash}m{0.25\textwidth} 
                    >{\arraybackslash}m{0.4\textwidth} 
                    >{\arraybackslash}m{0.3\textwidth}}
        \toprule
        \textbf{Vision Input} & \textbf{Prompt} & \textbf{Outputs} \\
        \midrule
        \includegraphics[width=\linewidth]{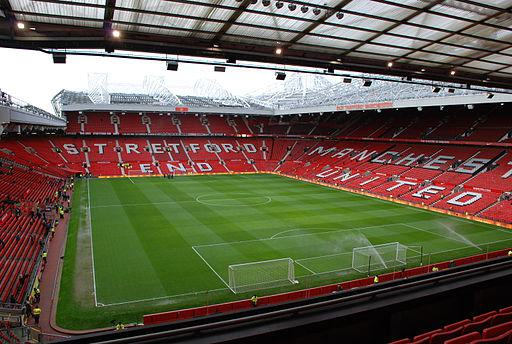} 
        & Observe the image closely, recognize the named entities present, and use your knowledge to answer the question in relation to those entities. \texttt{<IMAGE>} Who plays at this  stadium ?

        & \textbf{\lcm}: \textit{``The entities that play at this stadium are Red Stadium.''} 
        
        \textbf{\vlcm}: \textit{``The team that plays at this stadium is Manchester United.''} \\
        \midrule
        \includegraphics[width=\linewidth]{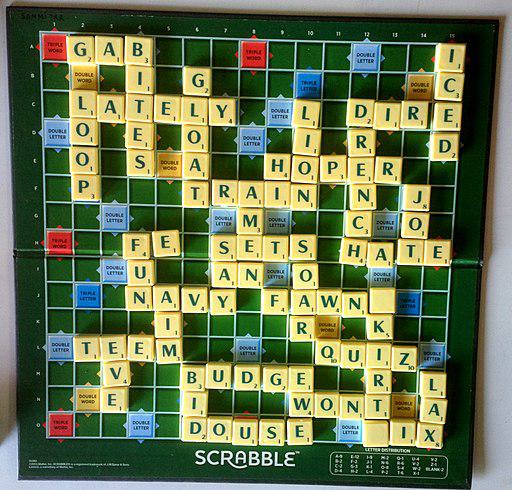} 
        & Analyze the image to locate any named entities and provide an appropriate response to the question. Start of vision. <IMAGE> End of vision. In this game, how many points is a 'V' tile worth?

        & \textbf{\lcm}: \textit{``It is not specified in the information provided that how many points a 'V' tile is worth.''} 
        
        \textbf{\vlcm}: \textit{``A 'V' tile is worth 4 points.''} \\
        \midrule
        \includegraphics[width=\linewidth]{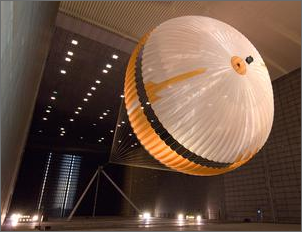} 
        & Carefully review the question and the related image/text context, identify the precise answer from the multiple-choice selections, and provide a detailed, logical explanation to support your choice, illustrating the multi-jumbo reasoning journey.
        
        \texttt{<IMAGE>}
        
        Question: What of the following could Gordon's test show?
        
        [...]
        
        Options:
        
        (A) if the spacecraft was damaged when using a parachute with a 1 m outlet going 200 km per hour.
        
        (B) How stable was a parachute with an 1 m vent at 200 km per hour.
        
        (C) whether a parachute with a 1 m vent would sway too much at 400 km/h.
        
        & \textbf{\lcm}: \textit{``(B) how stable was a parachute with a 1 m vent at 200 km per hour.''} 
        
        \textbf{\vlcm}: \textit{``The answer is (B) how stable was a parachute with a 1 m air leak was at 200 km per hour.''} \\
        \midrule
        \includegraphics[width=\linewidth]{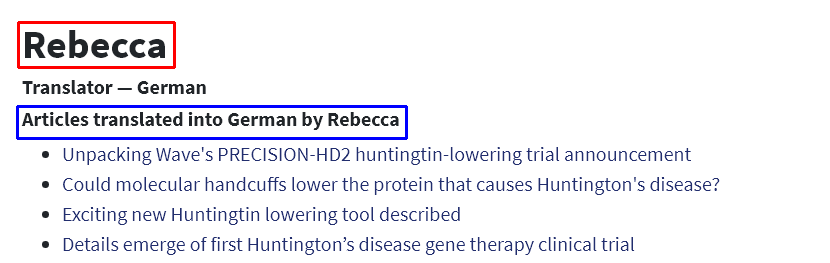} 
        & Please respond to the query by referring to the picture and its limited entities. \texttt{<IMAGE>} What is the name of the translator?  
        & \textbf{\lcm}: \textit{``The name of the translator is not provided.''} 
        
        \textbf{\vlcm}: \textit{``The name of the translator is Rebecca Herting.''} \\

        \bottomrule
    \end{tabular}
    }
    \caption{Qualitative examples for image question answering tasks. Each row shows the input image, the given prompt, and the generated model output.}
    \label{fig:vlm_qualitative2}
\end{figure*}

We present the qualitative cases for image captioning in \autoref{fig:vlm_qualitative1} and image question answering in \autoref{fig:vlm_qualitative2}.

\subsection{Video Qualitative Cases for \vlcm}

\begin{figure}[t] %
    \centering
    \adjustbox{max width=\textwidth}{
    \begin{tabular}{>{\centering\arraybackslash}m{1\textwidth} 
                    }
        \toprule
        \begin{minipage}{\linewidth}
            \centering
            \includegraphics[width=0.24\linewidth]{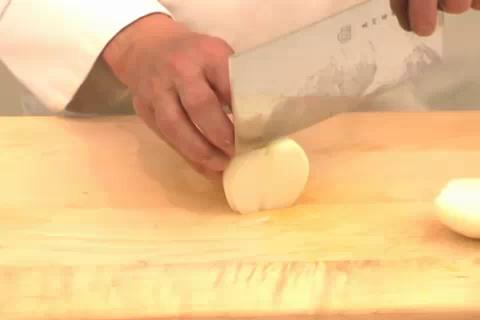}
            \includegraphics[width=0.24\linewidth]{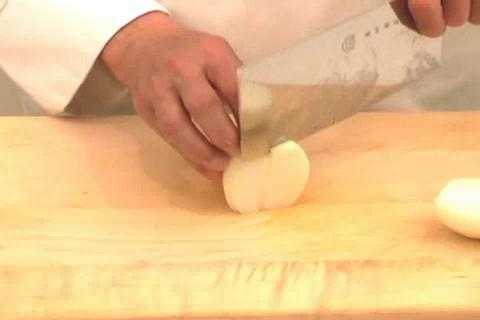}
            \includegraphics[width=0.24\linewidth]{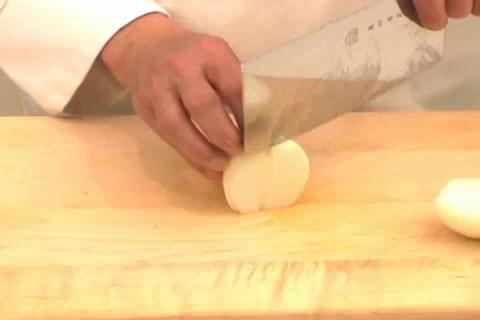}
            \includegraphics[width=0.24\linewidth]{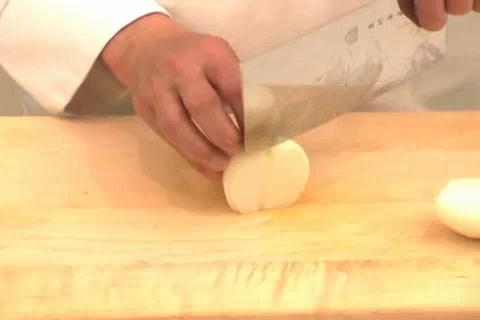}
            
            \includegraphics[width=0.24\linewidth]{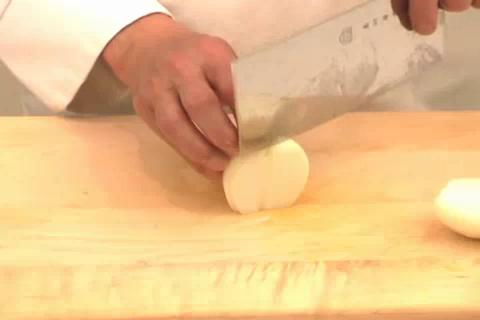}
            \includegraphics[width=0.24\linewidth]{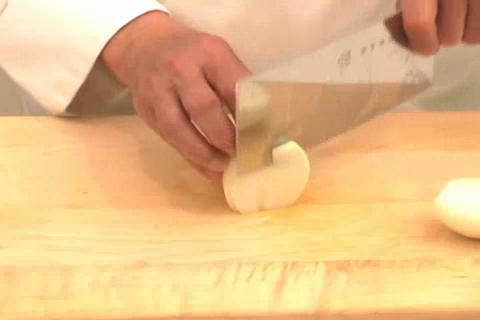}
            \includegraphics[width=0.24\linewidth]{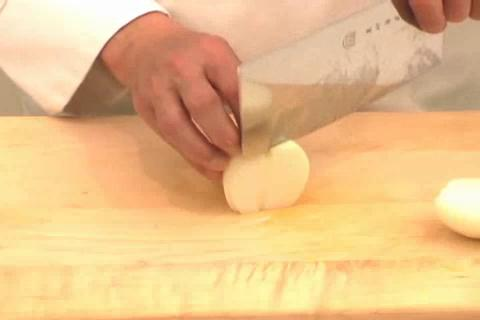}
            \includegraphics[width=0.24\linewidth]{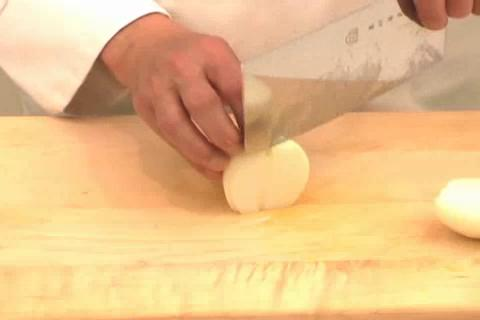}
        \end{minipage} \\[1em]
        \vspace{10pt}

        \begin{minipage}{\linewidth}
            \raggedright
            \textbf{Prompt:} You will be shown a video and asked a question about what you saw. Your answer should be a brief sentence that fits the video's contents. \texttt{<VIDEO>} what is a man slicing? \\
            \textbf{\lcm:} \textit{A man is cutting an onion with a knife.} \\
            \textbf{\vlcm:} \textit{The man in the video is slicing an onion.}
        \end{minipage} \\

        \midrule
        \begin{minipage}{\linewidth}
            \centering
            \includegraphics[width=0.24\linewidth]{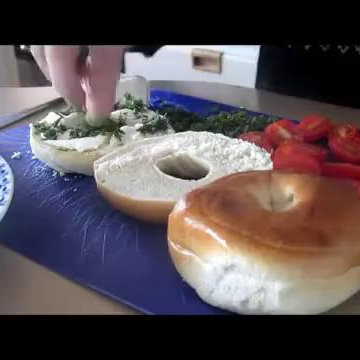}
            \includegraphics[width=0.24\linewidth]{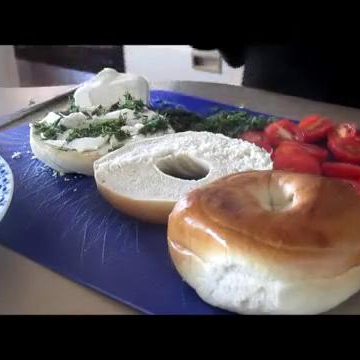}
            \includegraphics[width=0.24\linewidth]{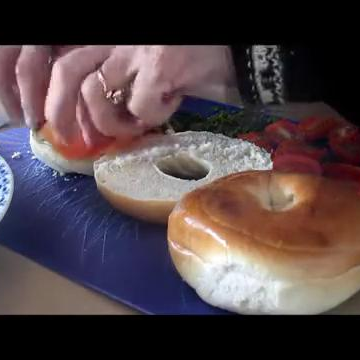}
            \includegraphics[width=0.24\linewidth]{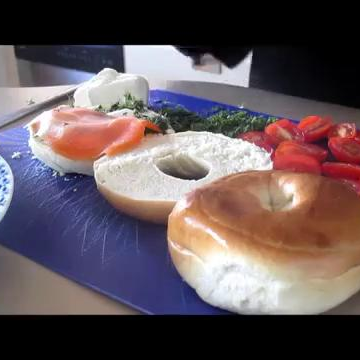}
            
            \includegraphics[width=0.24\linewidth]{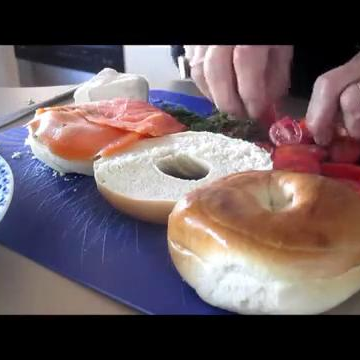}
            \includegraphics[width=0.24\linewidth]{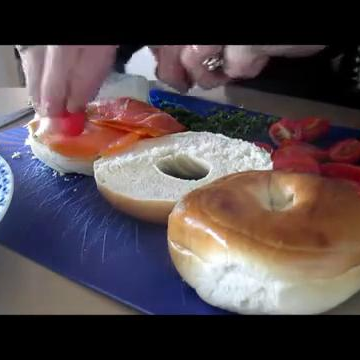}
            \includegraphics[width=0.24\linewidth]{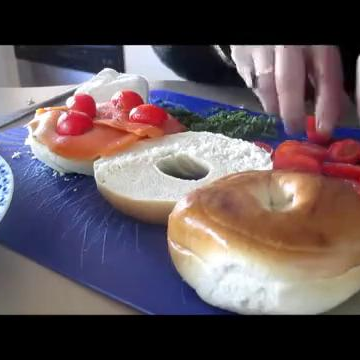}
            \includegraphics[width=0.24\linewidth]{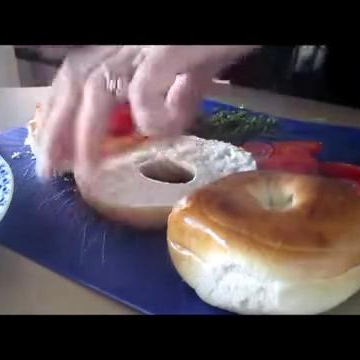}
        \end{minipage} \\[1em]
        \vspace{10pt}

        \begin{minipage}{\linewidth}
            \raggedright
            \textbf{Prompt:} Use your observation skills to answer the given question with the help of a short video clip. Start of vision. <VIDEO> End of vision. What type of bread is shown?
            \\
            \textbf{\lcm:} \textit{The type of bread shown is a sandwich slice.} \\
            \textbf{\vlcm:} \textit{The bread shown in the video is a bagel.}
        \end{minipage} \\
        \midrule

        \begin{minipage}{\linewidth}
            \centering
            \includegraphics[width=0.23\linewidth]{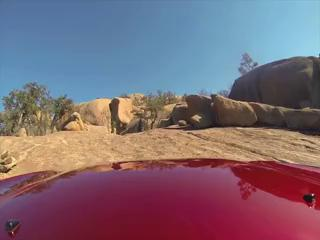}
            \includegraphics[width=0.23\linewidth]{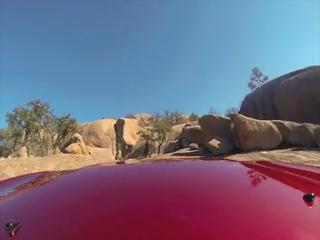}
            \includegraphics[width=0.23\linewidth]{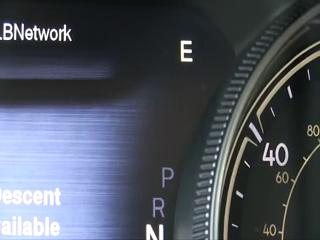}
            \includegraphics[width=0.23\linewidth]{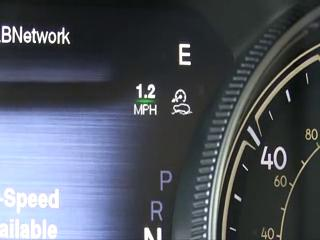}
            
        \end{minipage} \\[1em]
        \vspace{10pt}

        \begin{minipage}{\linewidth}
            \raggedright
            \textbf{Prompt:} Watch the video and create a caption based on the visual cues. \texttt{<VIDEO>} \\
            \textbf{\lcm:} \textit{[End of Response.]} \\
            \textbf{\vlcm:} \textit{someone is driving a red car}
        \end{minipage} \\

        \bottomrule
    \end{tabular}
    }
    \caption{Qualitative examples for video captioning and question answering tasks. The frames are presented from left to right, and top to the bottom.}
    \label{fig:qual_video_block}
\end{figure}

We present the qualitative cases for video captioning and question answering task in \autoref{fig:qual_video_block}.

\subsection{Qualitative Cases for Video Captioning by \vsonar}

\begin{figure*}[t]
    \centering
    \setlength{\tabcolsep}{1pt} %
    
    \begin{tabular}{cccccccc}
        \includegraphics[width=0.24\linewidth]{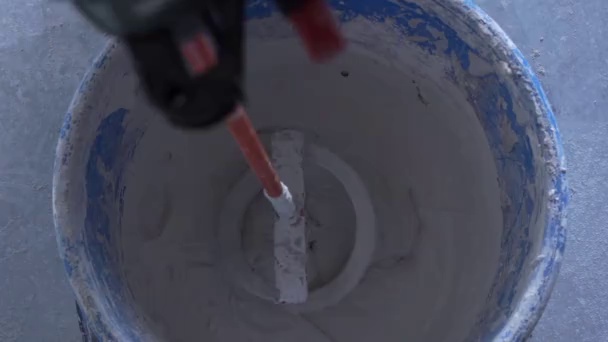} &
        \includegraphics[width=0.24\linewidth]{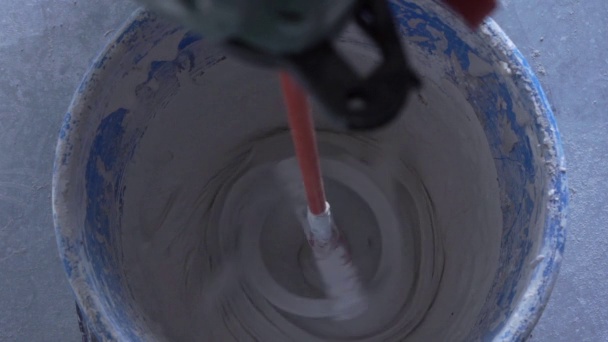} &
        \includegraphics[width=0.24\linewidth]{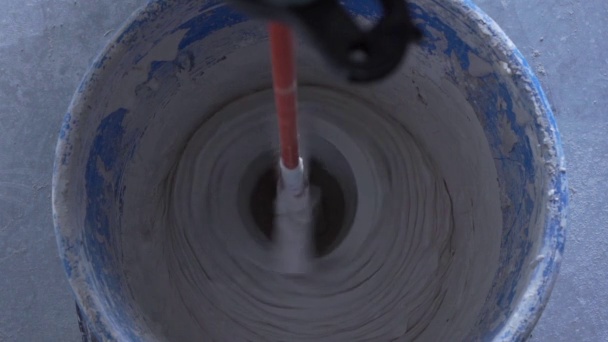} &
        \includegraphics[width=0.24\linewidth]{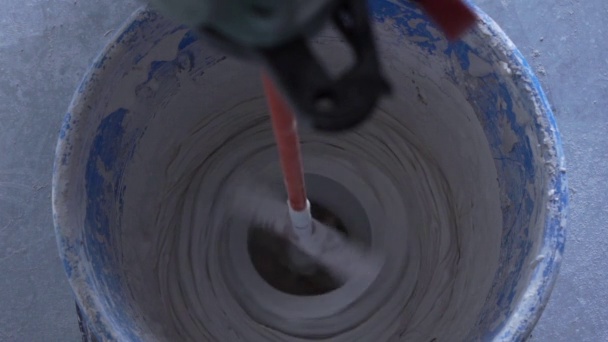} & \\
        \scriptsize $t=1$ & \scriptsize $t=2$ & \scriptsize $t=3$ & \scriptsize $t=4$ & \\
        \includegraphics[width=0.24\linewidth]{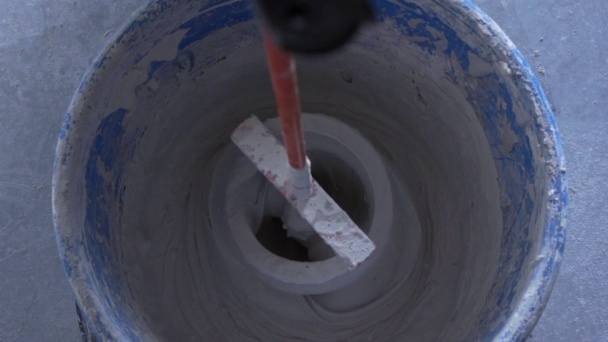} &
        \includegraphics[width=0.24\linewidth]{qualitative-figures/PE-Video_vid_6_idx6/frame_04_idx_506.jpg} &
        \includegraphics[width=0.24\linewidth]{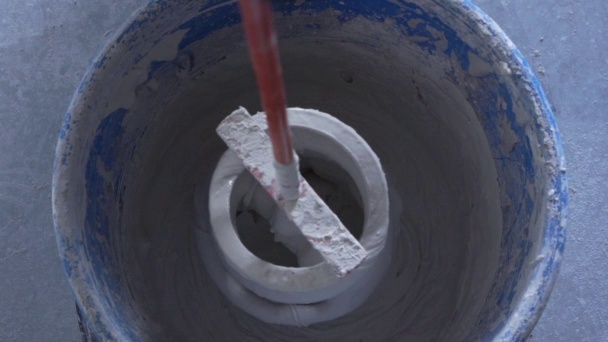} &
        \includegraphics[width=0.24\linewidth]{qualitative-figures/PE-Video_vid_6_idx6/frame_05_idx_632.jpg} \\
        \scriptsize $t=5$ & \scriptsize $t=6$ & \scriptsize $t=7$ & \scriptsize $t=8$ \\
    \end{tabular}
    
    \vspace{2mm}

    \renewcommand{\arraystretch}{1.3} %
    \small
    \begin{tabularx}{\linewidth}{l X}
        \toprule
        \textbf{Models} & \textbf{Video Captions} \\
        \midrule
        
        \rowcolor{deepmindblue}
        \textbf{\vsonar (Ours)} & 
        The video shows a \textcolor{goodgreen}{\textbf{blue mixer}} stirring a \textcolor{goodgreen}{\textbf{white powder}} in a round bucket. The bucket is moving with a rotating motion and is attached to a blue cylinder. The mixer is moving slowly in the middle, creating a white powder. \\
        
        PLM-3B & 
        In the background, there is a bucket, a stick, and a grey surface. The sound of the mixer is audible. \textit{[Vague; mentions audio]} \\
        
        Qwen2.5-VL-3B-Instruct & 
        A close-up shot of a blue bucket filled with white paint. A \textcolor{badred}{\textbf{red-handled paintbrush}} is dipped into the paint and then lifted out, leaving a trail of paint behind it. The brush is then lowered back into the paint and the process is \textcolor{badred}{\textbf{repeated several times}}. The camera remains stationary. \\
        
        InternVL2.5-1B & 
        A \textcolor{badred}{\textbf{yellow hand}} reaches into the white plastic bucket, grabbing the red cylindrical object, which is a tool for \textcolor{badred}{\textbf{pouring cement}}. \\
        
        \bottomrule
    \end{tabularx}

    \caption{\textbf{Qualitative comparison on fine-grained motion understanding in PE-Video.} 
    While state-of-the-art VLMs (Qwen-2.5, InternVL-2.5) suffer from hallucinations or captioning erors (highlighted in \textcolor{badred}{\textbf{red}}) such as a ``paintbrush'' or ``yellow hand,'' and PLM remains vague, \textbf{SONAR} accurately captures the mechanics of the mixer and the blue cylinder attachment (highlighted in \textcolor{goodgreen}{\textbf{green}}).}
    \label{fig:qualitative_pe_video_6}
\end{figure*}

\begin{figure*}[t]
    \centering
    \setlength{\tabcolsep}{1pt} %
    
    \begin{tabular}{cccccccc}
        \includegraphics[width=0.24\linewidth]{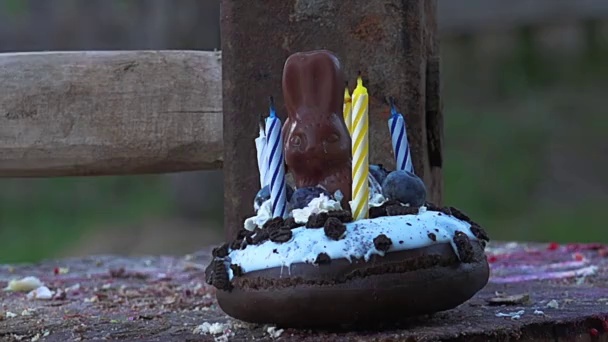} &
        \includegraphics[width=0.24\linewidth]{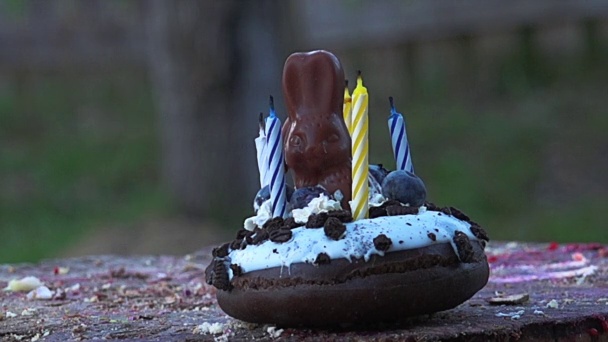} &
        \includegraphics[width=0.24\linewidth]{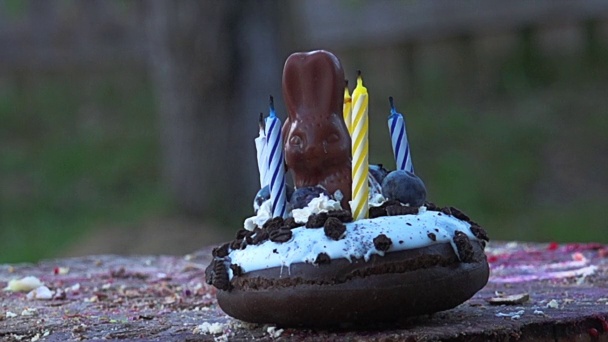} &
        \includegraphics[width=0.24\linewidth]{qualitative-figures/PE-Video_vid_9_idx9/frame_02_idx_147.jpg} & \\
        \scriptsize $t=1$ & \scriptsize $t=2$ & \scriptsize $t=3$ & \scriptsize $t=4$ & \\
        \includegraphics[width=0.24\linewidth]{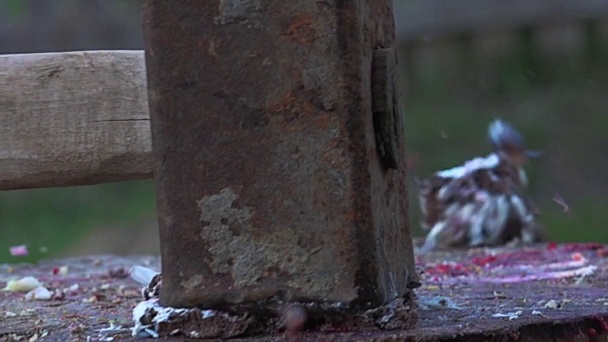} &
        \includegraphics[width=0.24\linewidth]{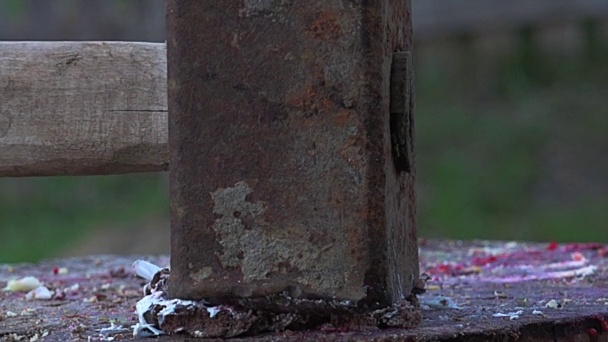} &
        \includegraphics[width=0.24\linewidth]{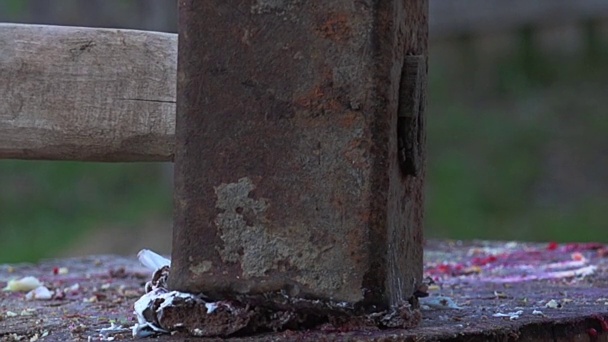} &
        \includegraphics[width=0.24\linewidth]{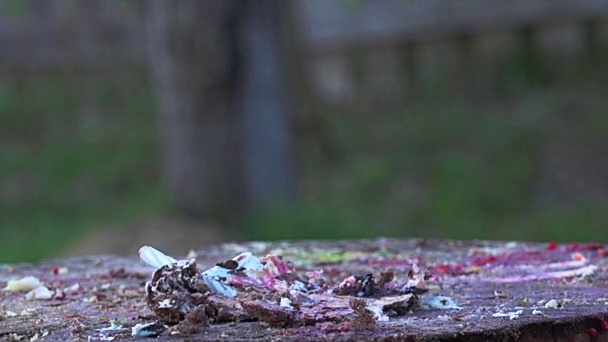} \\
        \scriptsize $t=5$ & \scriptsize $t=6$ & \scriptsize $t=7$ & \scriptsize $t=8$ \\
    \end{tabular}
    
    \vspace{2mm}

    \renewcommand{\arraystretch}{1.3} %
    \small
    \begin{tabularx}{\linewidth}{l X}
        \toprule
        \textbf{Models} & \textbf{Video Captions} \\
        \midrule
        
        \rowcolor{deepmindblue}
        \textbf{\vsonar (Ours)} & 
        The video shows a cake with a candy bar on top. The frame zooms in on a bunny and a chocolate bar, then the candy bar is removed from the frame. The video ends with the bunny lying on the ground, and a camera zooms in on the cake. \\
        
        PLM-3B & 
        In the background, \textcolor{badred}{\textbf{there is a brown surface, a brown object, a brown wall, a green grass surface, and a brown object}}. The sound of the music is audible. \textit{[Vague; mentions audio]} \\
        
        Qwen2.5-VL-3B-Instruct & 
        \textcolor{badred}{\textbf{A chocolate donut}} with white frosting and blueberries is placed on a wooden surface. The donut has three candles on it, two blue and one yellow. The donut is lit up by the candles, and \textcolor{badred}{\textbf{the flames are visible}}. The donut is surrounded by crumbs and other debris. The background is blurred, but it appears to be an outdoor setting with grass and trees. \\
        
        InternVL2.5-1B & 
        The small cake on the table is seen being violently hit by a heavy club-like object from which \textcolor{badred}{\textbf{the smoke begins to rise}}. The cake is shown up close on the table \textcolor{badred}{\textbf{surrounded by paper}}. After the cake is violently hit, the hammer used is placed back onto the table and continues to hit the marble coaster.
         \\
        
        \bottomrule
    \end{tabularx}

    \caption{\textbf{Qualitative comparison in PE-Video.} 
    We highlight the errors in \textcolor{badred}{\textbf{red}}.}
    \label{fig:qualitative_pe_video_9}
\end{figure*}

\begin{figure*}[t]
    \centering
    \setlength{\tabcolsep}{1pt} %
    
    \begin{tabular}{cccccccc}
        \includegraphics[width=0.24\linewidth]{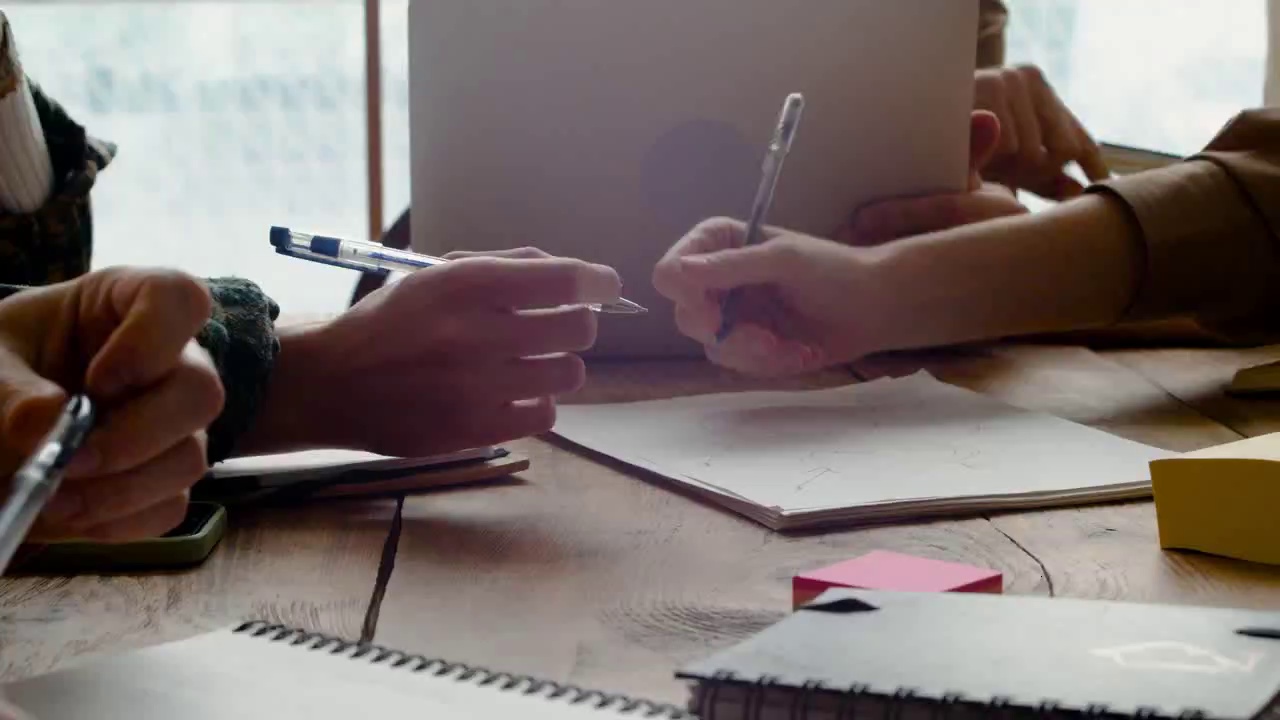} &
        \includegraphics[width=0.24\linewidth]{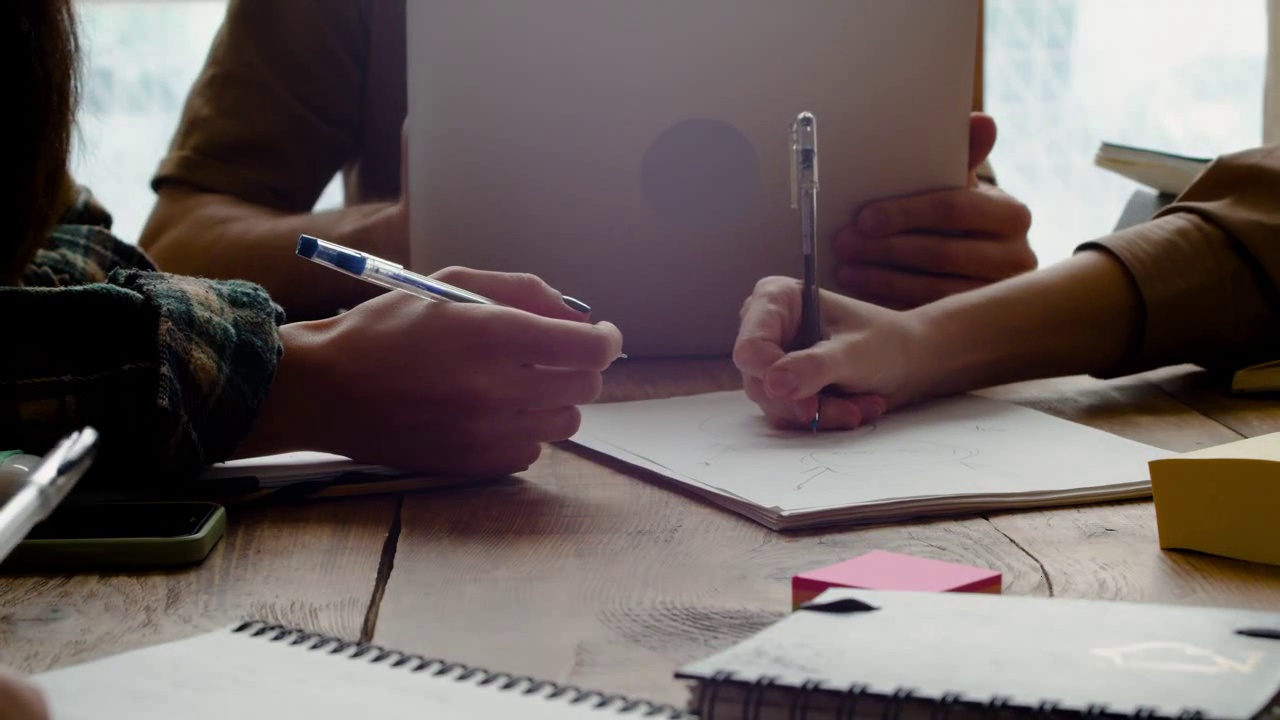} &
        \includegraphics[width=0.24\linewidth]{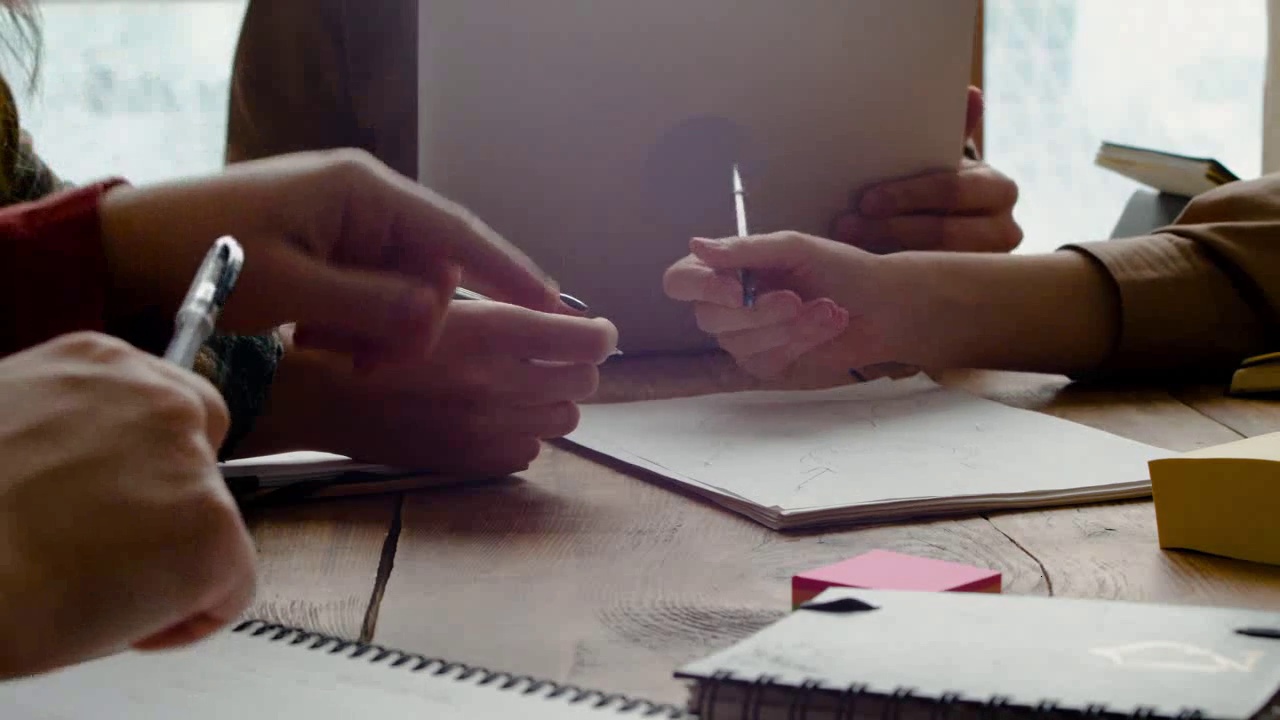} &
        \includegraphics[width=0.24\linewidth]{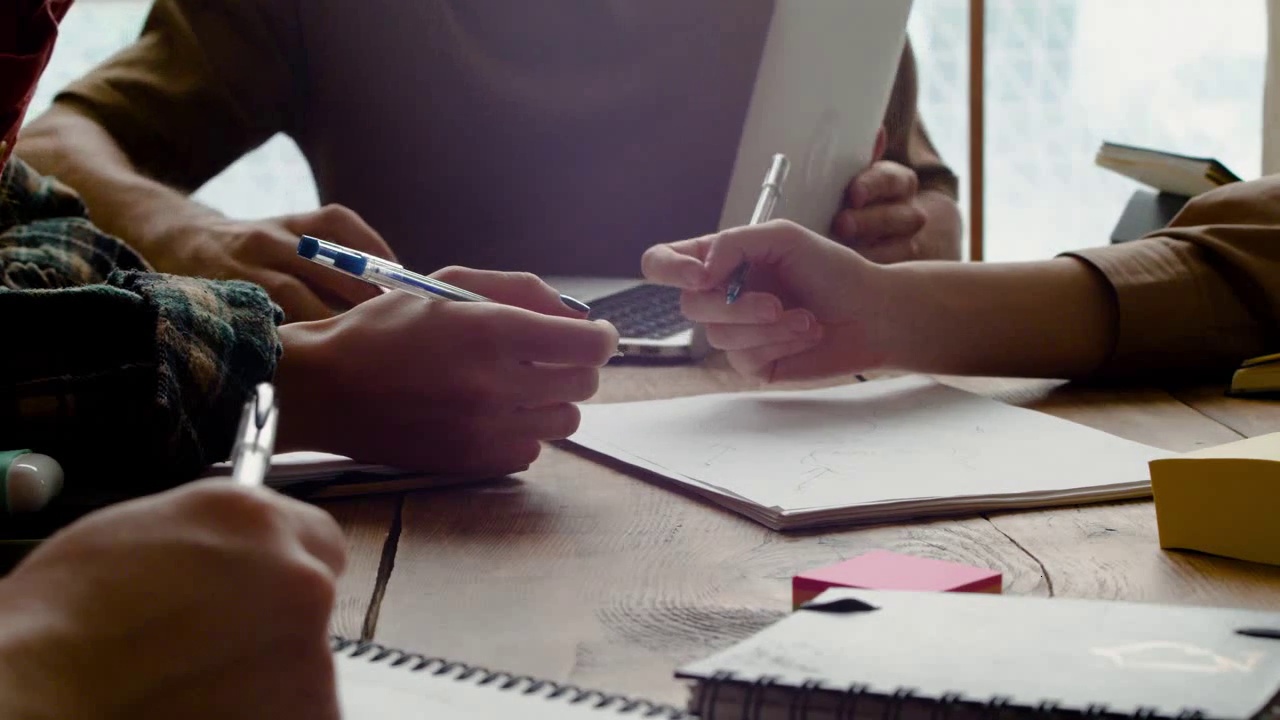} & \\
        \scriptsize $t=1$ & \scriptsize $t=2$ & \scriptsize $t=3$ & \scriptsize $t=4$ & \\
        \includegraphics[width=0.24\linewidth]{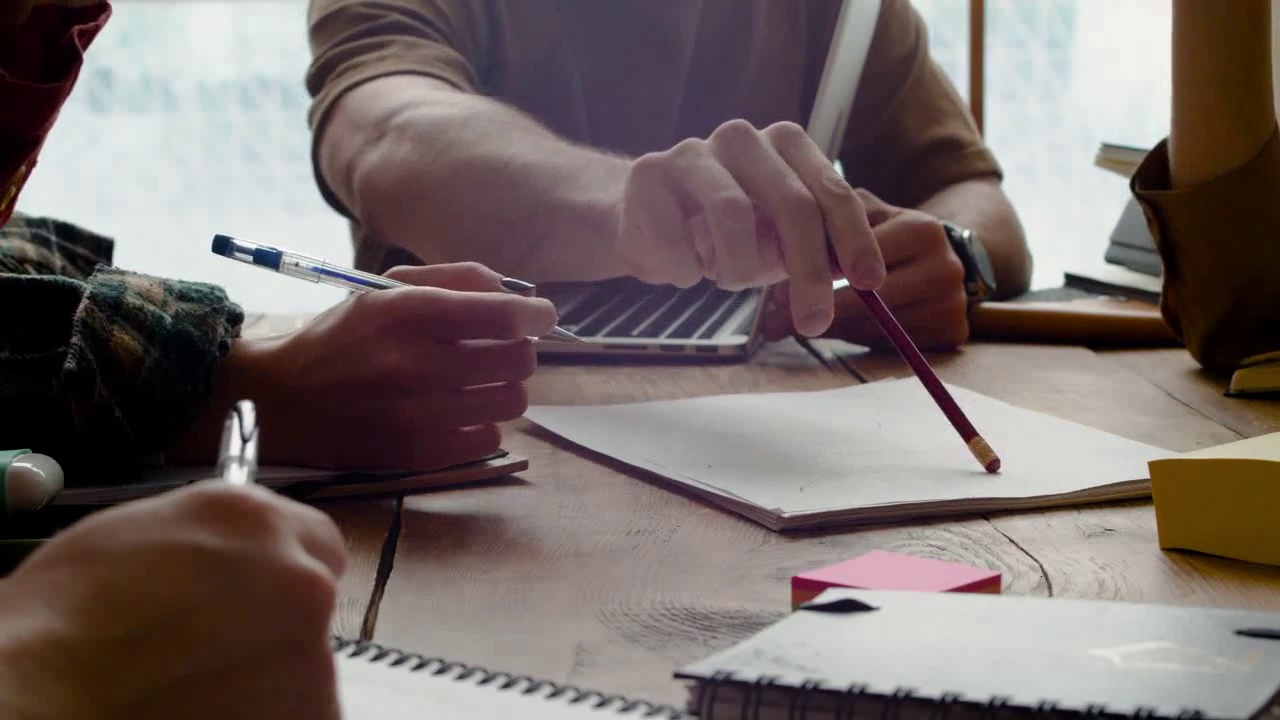} &
        \includegraphics[width=0.24\linewidth]{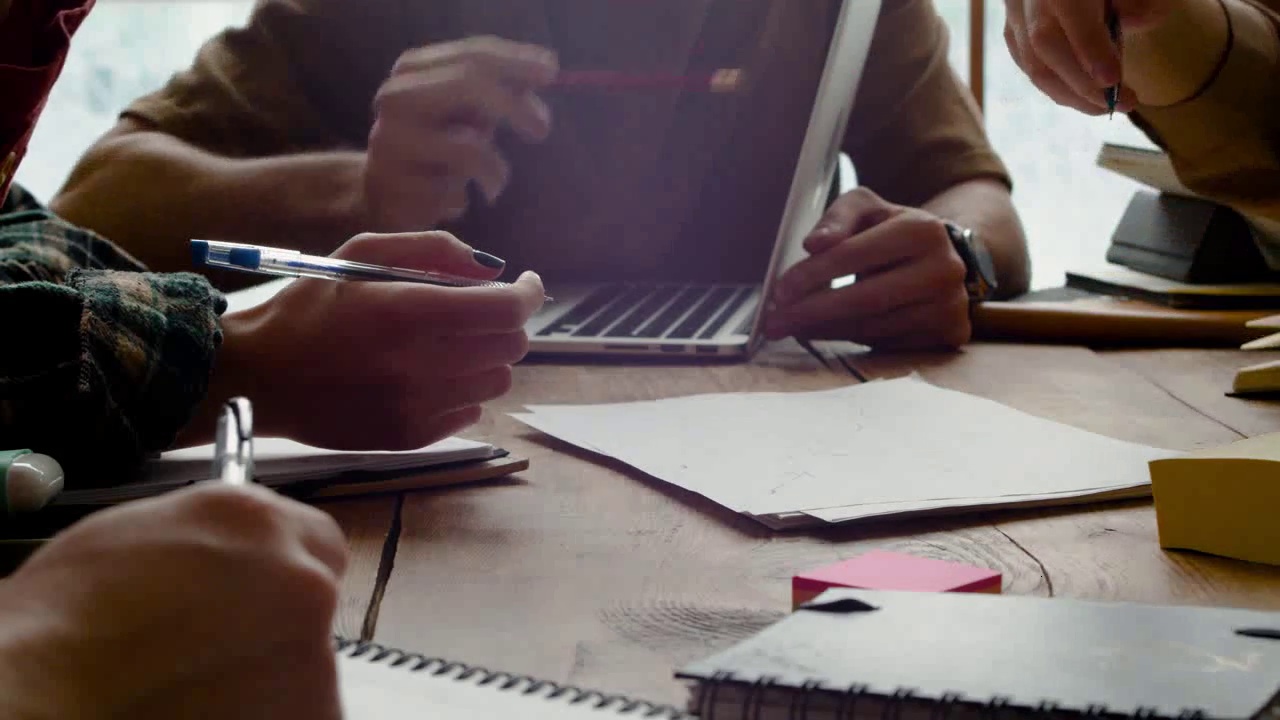} &
        \includegraphics[width=0.24\linewidth]{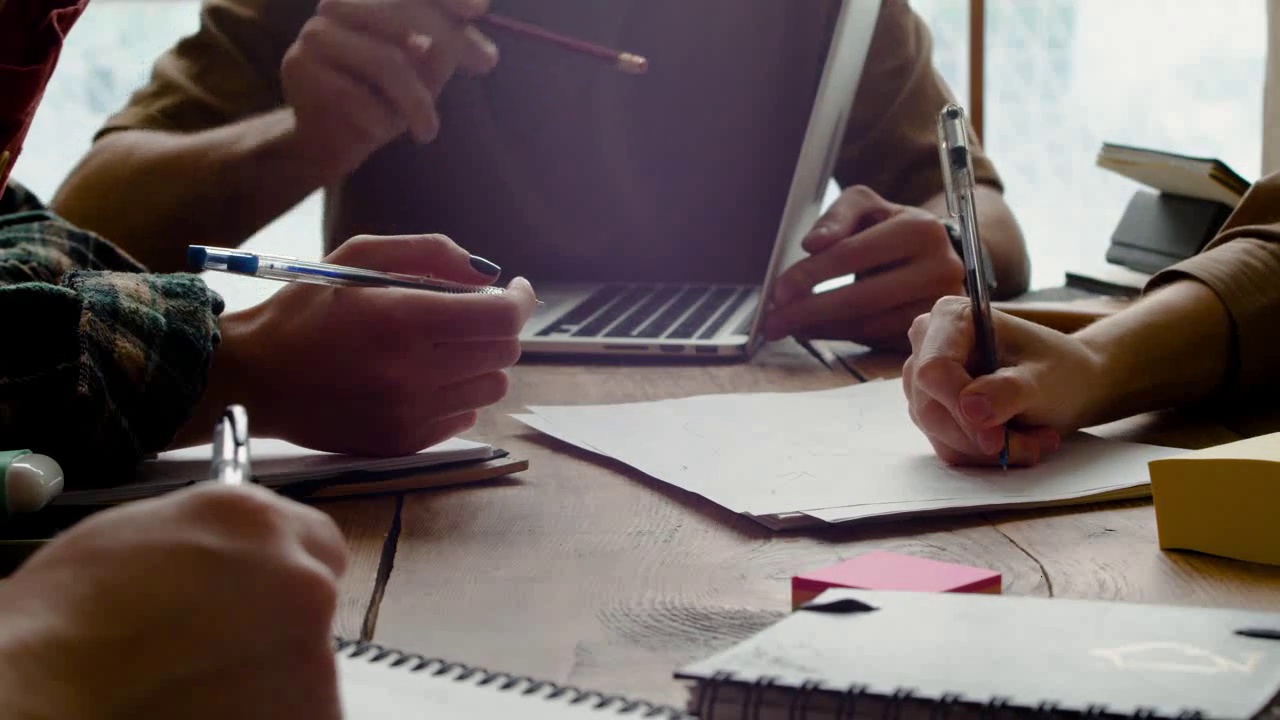} &
        \includegraphics[width=0.24\linewidth]{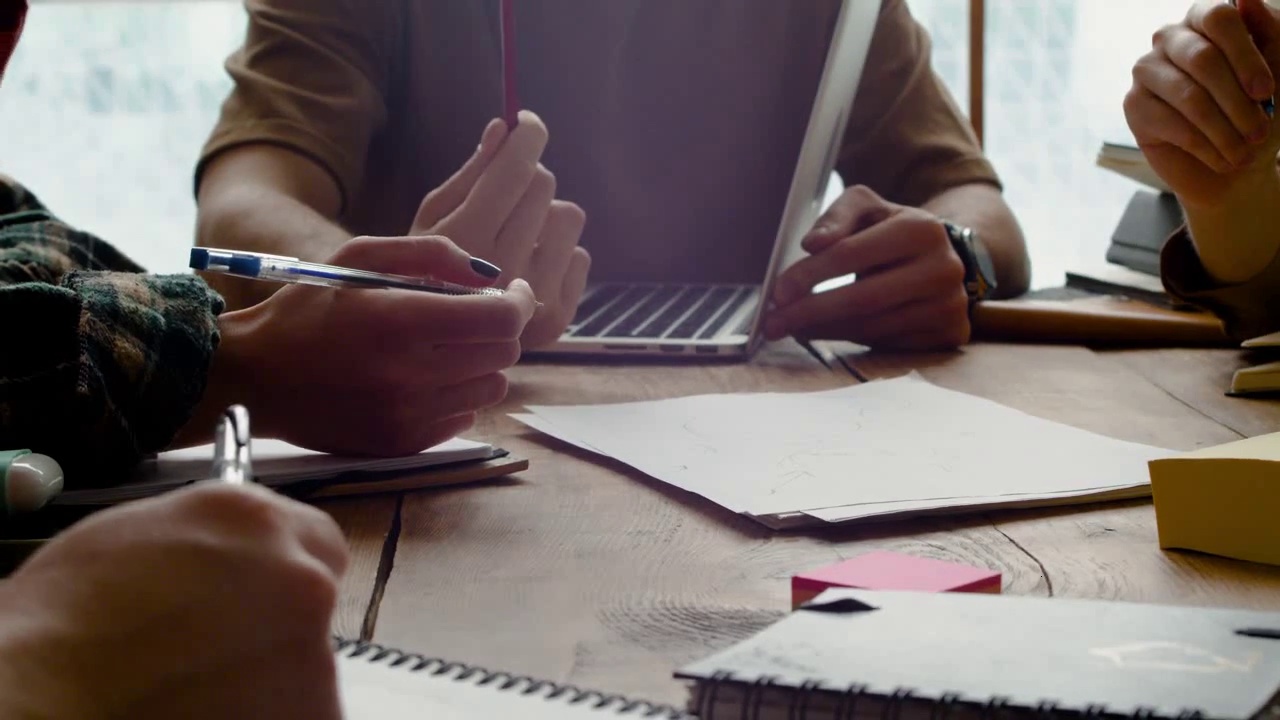} \\
        \scriptsize $t=5$ & \scriptsize $t=6$ & \scriptsize $t=7$ & \scriptsize $t=8$ \\
    \end{tabular}
    
    \vspace{2mm}

    \renewcommand{\arraystretch}{1.3} %
    \small
    \begin{tabularx}{\linewidth}{l X}
        \toprule
        \textbf{Models} & \textbf{Video Captions} \\
        \midrule
        
        \rowcolor{deepmindblue}
        \textbf{\vsonar (Ours)} & 
        The video shows a group of people sitting around a table with notebooks on their hands. The people are writing and drawing on the paper, some are holding a pencil, and others are holding a notebook. The table is set in the foreground with people moving around it, and they are writing on the paper. \\
        
        PLM-3B & 
        The video features a group of people sitting around a table, engaged in a collaborative activity. The individuals are holding pens and pencils, and there are various objects on the table, including a laptop, notebooks, and sticky notes. The scene appears to be a meeting or brainstorming session, with the participants working together to create something. The video is in slow motion, which adds a sense of calmness and focus to the scene. Overall, the video conveys a sense of collaboration, creativity, and productivity.
         \\
        
        Qwen2.5-VL-3B-Instruct & 
        The video shows a group of people sitting around a table, working on a project together. They are using laptops and writing notes on paper. The atmosphere is collaborative and focused, with everyone contributing to the task at hand. The lighting is warm and inviting, creating a cozy and inviting environment. The people are engaged in their work, and the overall mood is positive and productive.
         \\
        
        InternVL2.5-1B & 
        In this video, a team of individuals collaborates to contribute to the design process on a desk.
         \\
        
        \bottomrule
    \end{tabularx}

    \caption{\textbf{Qualitative comparison in DREAM-1K.} }
    \label{fig:qualitative_dream1k_600}
\end{figure*}

\begin{figure*}[t]
    \centering
    \setlength{\tabcolsep}{1pt} %
    
    \begin{tabular}{cccccccc}
        \includegraphics[width=0.24\linewidth]{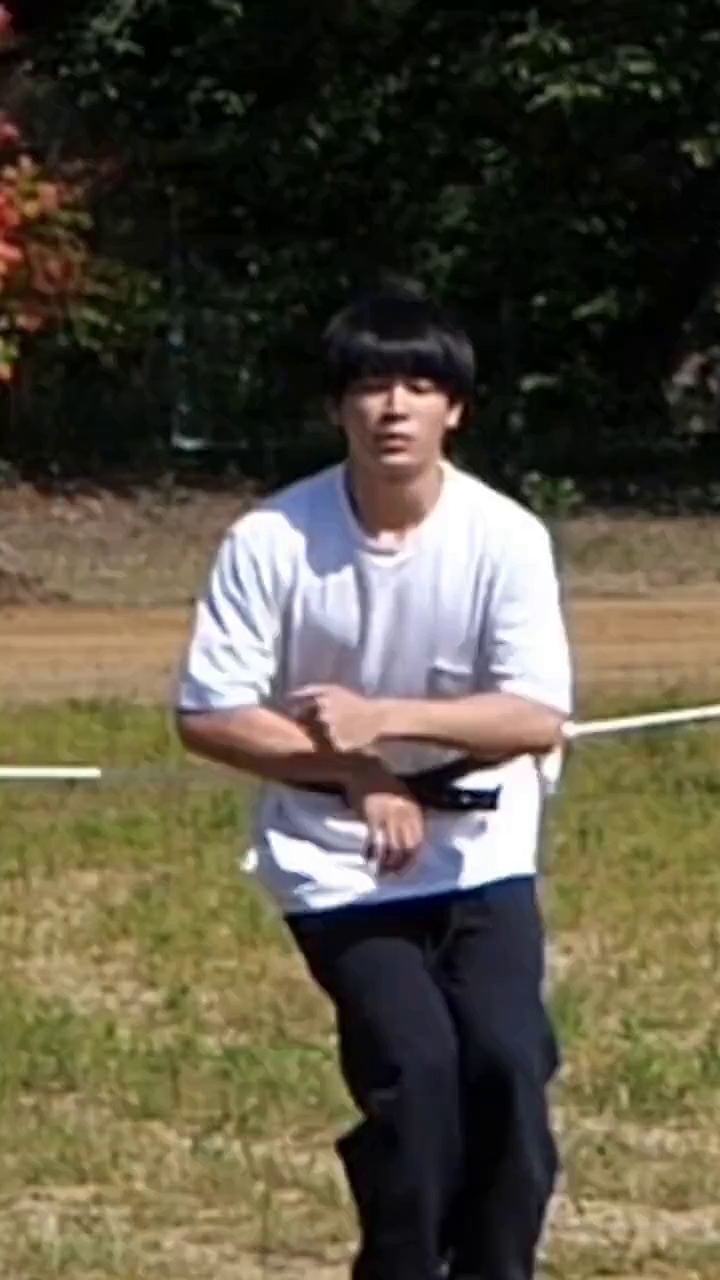} &
        \includegraphics[width=0.24\linewidth]{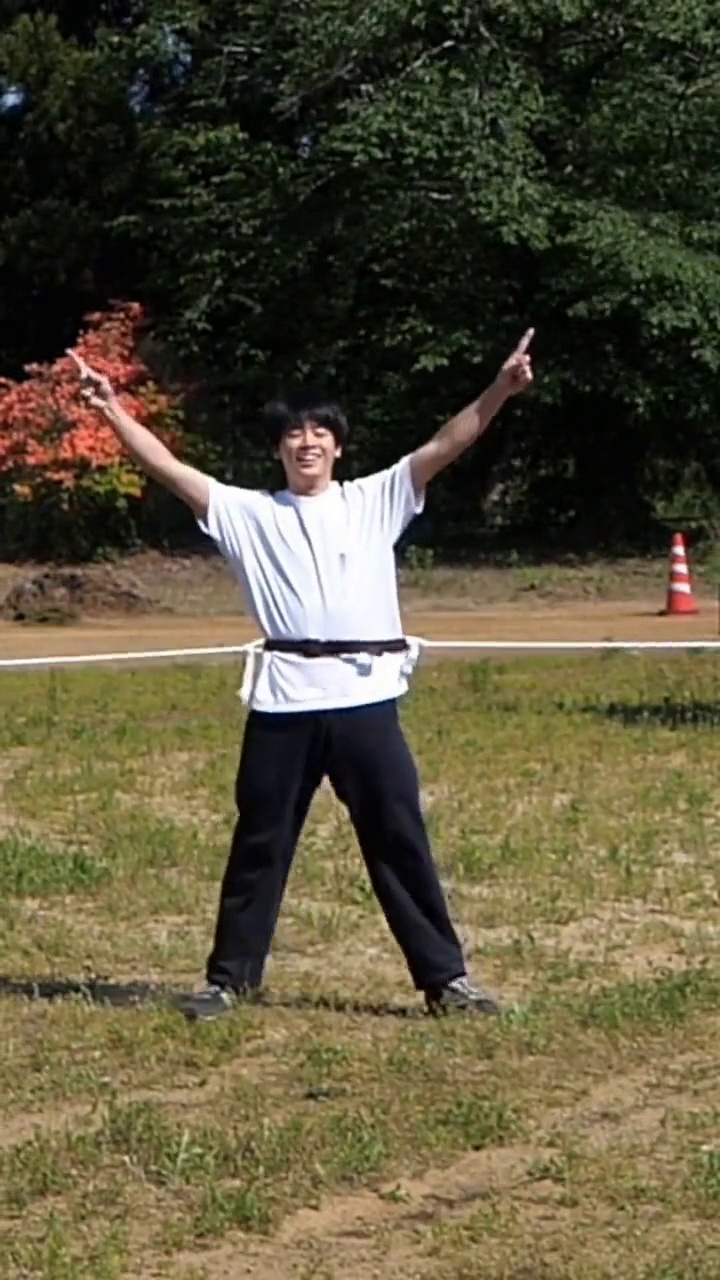} &
        \includegraphics[width=0.24\linewidth]{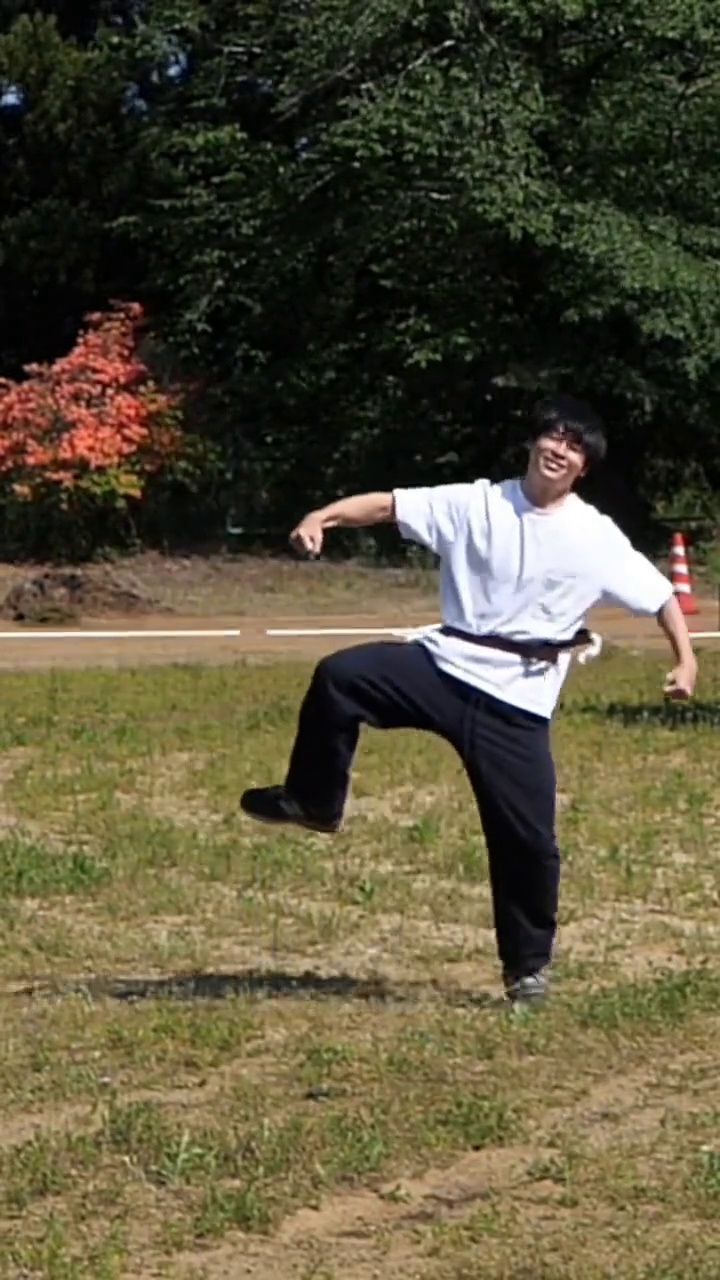} &
        \includegraphics[width=0.24\linewidth]{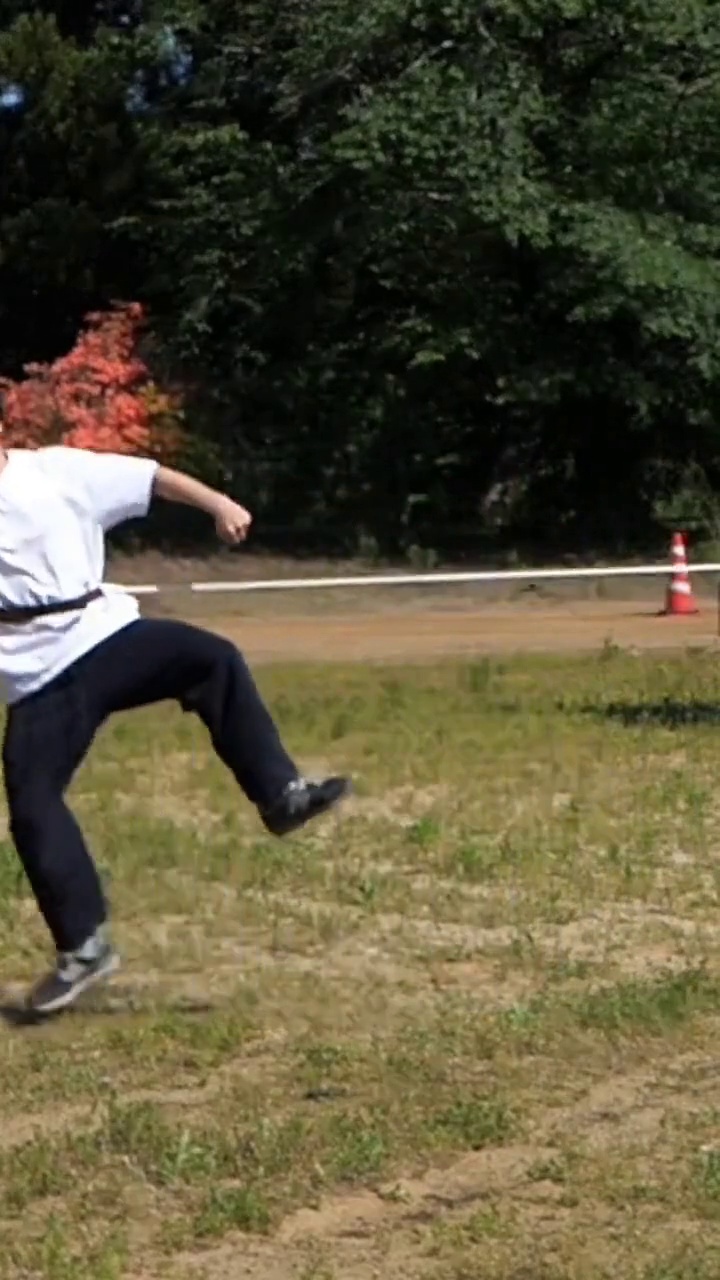} & \\
        \scriptsize $t=1$ & \scriptsize $t=2$ & \scriptsize $t=3$ & \scriptsize $t=4$ & \\
        \includegraphics[width=0.24\linewidth]{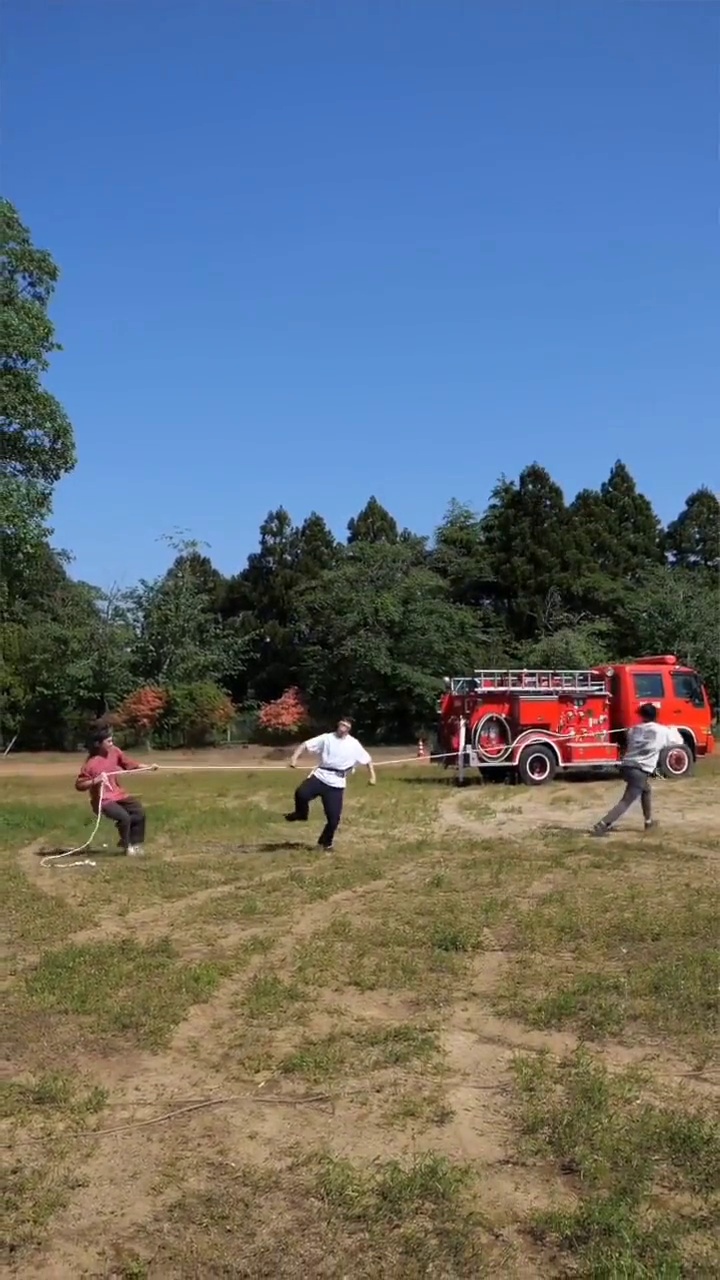} &
        \includegraphics[width=0.24\linewidth]{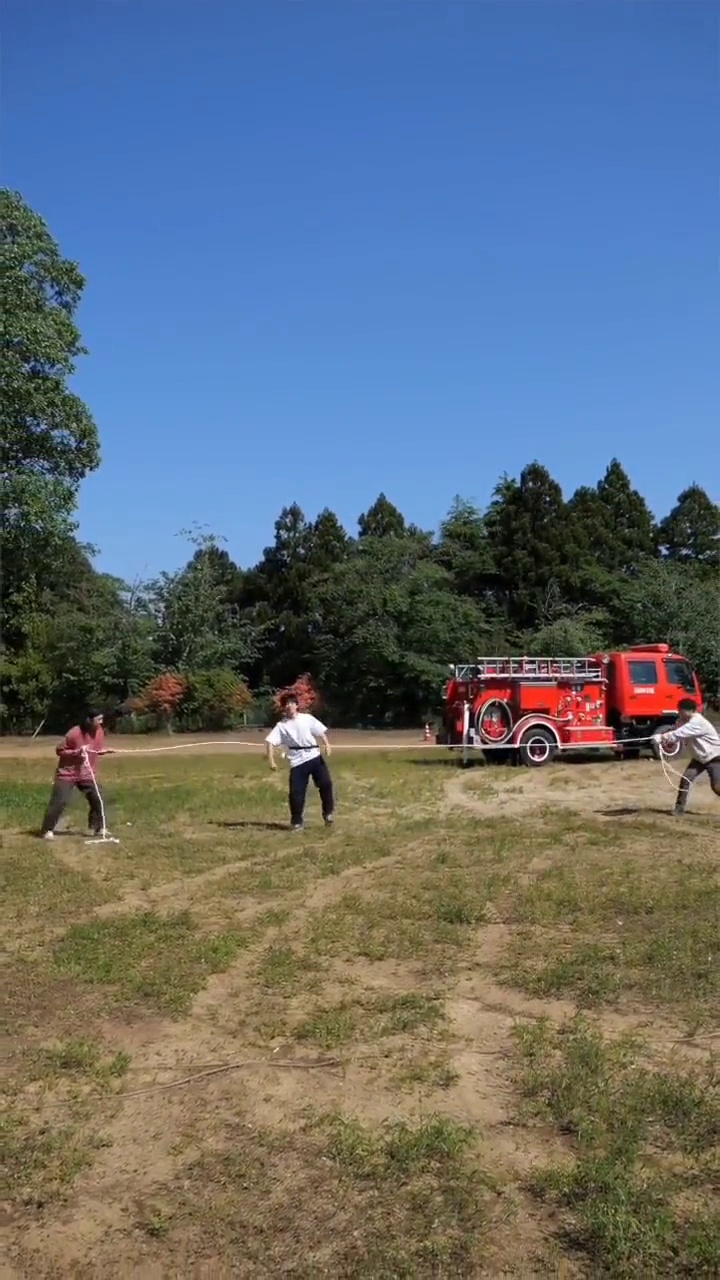} &
        \includegraphics[width=0.24\linewidth]{qualitative-figures/dream1k_vid_500_idx500/frame_05_idx_563.jpg} &
        \includegraphics[width=0.24\linewidth]{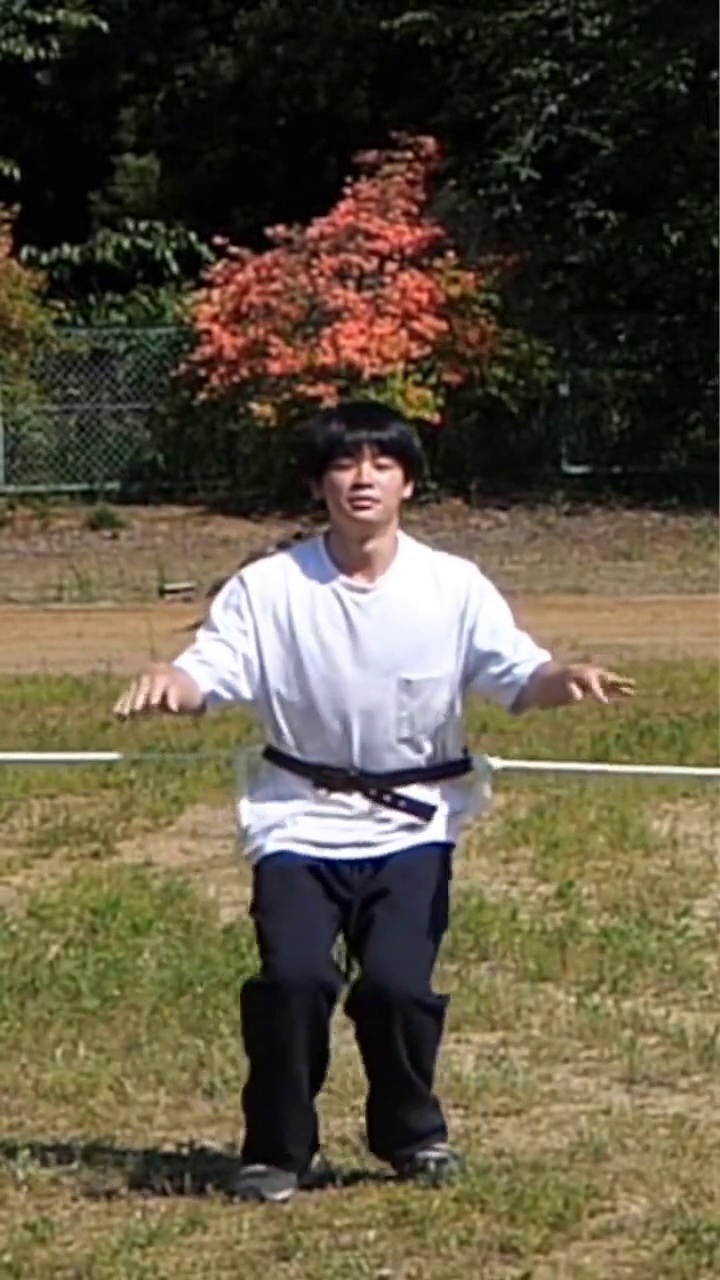} \\
        \scriptsize $t=5$ & \scriptsize $t=6$ & \scriptsize $t=7$ & \scriptsize $t=8$ \\
    \end{tabular}
    
    \vspace{2mm}

    \renewcommand{\arraystretch}{1.3} %
    \small
    \begin{tabularx}{\linewidth}{l X}
        \toprule
        \textbf{Models} & \textbf{Video Captions} \\
        \midrule
        
        \rowcolor{deepmindblue}
        \textbf{\vsonar (Ours)} & 
        The video shows a group of men dancing in a white shirt. A man is standing in the field with a rope in his hands and a red truck behind him. The man is pulling the rope and the truck is moving towards the field. \\
        
        PLM-3B & 
        A man wearing a white t-shirt and black pants is standing on the grass surface and then starts running and kicking the white rope. Another man wearing a red t-shirt and black pants is standing on the grass surface and then starts running and kicking the white rope. The third man wearing a white t-shirt and black pants is standing on the grass surface and then starts running and kicking the white rope. In the background, there is a grass surface, trees, a red vehicle, a white rope, and the sky. The voice of the people is audible. \\
        
        Qwen2.5-VL-3B-Instruct & 
        The video shows a man standing in a field, wearing a white t-shirt and black pants. He is holding a rope and appears to be playing a game of tug of war. The background features trees and a clear blue sky. \\
        
        InternVL2.5-1B & 
        A performer in a white shirt and black belt balances on a long rope in a park, with a red vehicle and a lush red tree in the background. The performer changes their stance, eventually falling off the rope and landing on the ground, then quickly gets back up ready for the next try or battle.
         \\
        
        \bottomrule
    \end{tabularx}

    \caption{\textbf{Qualitative comparison in DREAM-1K.}}
    \label{fig:qualitative_dream1k_500}
\end{figure*}

In addition to the qualitative cases for LCM. We present the qualitative cases for video captioning with \vsonar compared to other VLMs in PVD-Bench and DREAM-1K from \autoref{fig:qualitative_pe_video_6} to \autoref{fig:qualitative_dream1k_500}.

\subsection{Qualitative Cases for Video Captioning by \vsonar}

\begin{figure*}[t]
    \centering
    \setlength{\tabcolsep}{1pt} %
    
    \begin{tabular}{c}
        \includegraphics[width=0.9\linewidth]{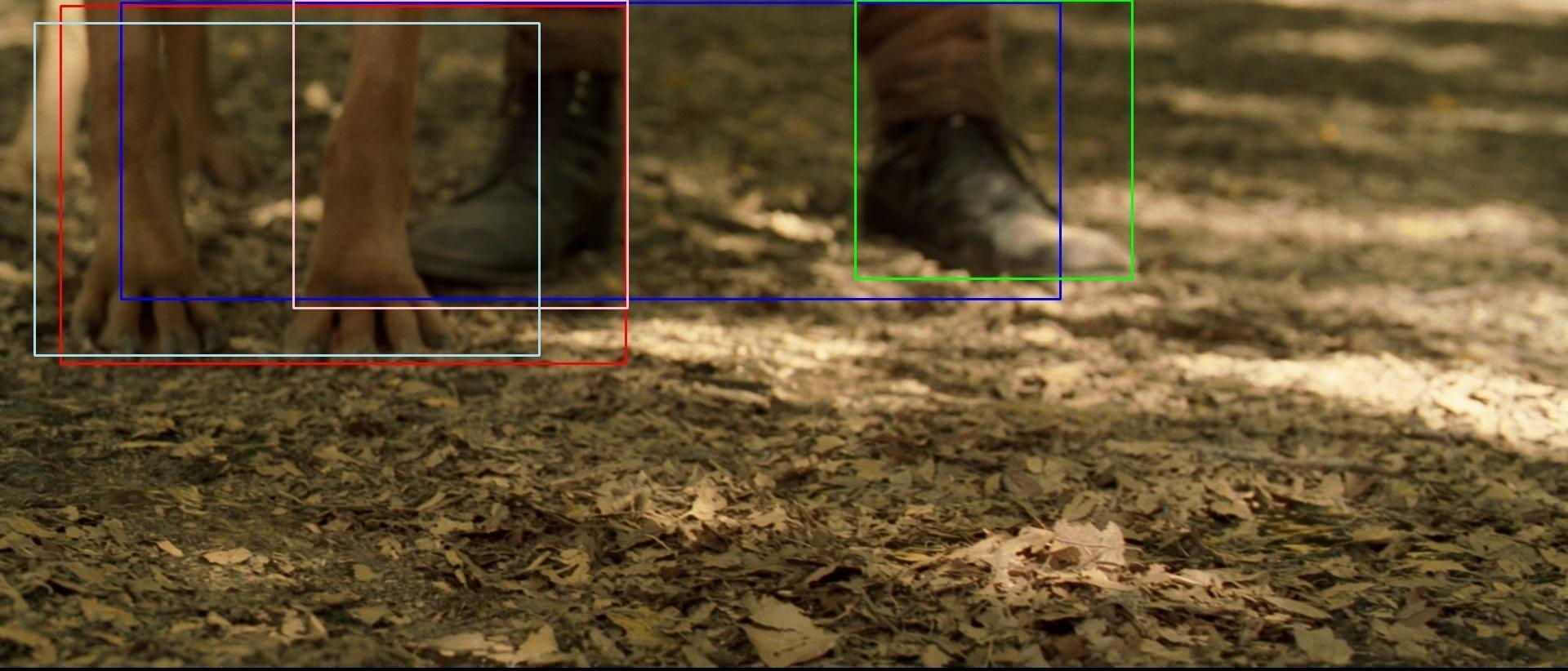}
    \end{tabular}
    
    \vspace{2mm}

    \renewcommand{\arraystretch}{1.3} %
    \small
    \begin{tabularx}{\linewidth}{l X}
        \toprule
        \textbf{Models} & \textbf{} \\
        \midrule
        
        \rowcolor{deepmindblue}
        \textbf{Prompt} & 
         Given an image and some bounded objects, answer the question and explains why the answer is right. 
         
         Objects: The 0-th object is person in the red color box. The 1-th object is person in the blue color box. The 2-th object is person in the green color box. The 3-th object is person in the pink color box. The 4-th object is dog in the light blue color box. 
         
         Question: What is the 3-th object doing with the 4-th object ?   \\
        
        PLM-8B & 
        Young children enjoy taking their pets for a walk outside. \\
        
        Qwen2.5-VL-8B-Instruct & 
        Answer (A) the 3-th object is taking the 4-th object for a walk outside.  Rationale: The 3-th object appears to be a child based on the size and posture, and the 4-th object is a dog. The context of the image suggests an outdoor setting with a child and a dog, which aligns with the idea of a child taking a pet for a walk. Dogs often enjoy walks, and it's common for young children to take their pets for walks as part of their daily routine or playtime. The other options do not fit the context of the image as well; there is no indication of flirting, looking at a computer, or setting up anything. \\
        
        InternVL2.5-8B & 
        the 3-th object was setting up the 4-th object. \\

        v-LCM & 
        The 3rd object is taking the 4th object for a walk outside. 'Cause little kids like to take their pets for a walk.
         \\
        
        \bottomrule
    \end{tabularx}

    \caption{\textbf{Qualitative comparison in Visual-Commonsense-Reasoning (VCR),} which requires grounding and spatial reasoning ability connecting with the commonsense.}
    \label{fig:qualitative_vcr_2}
\end{figure*}

We present the qualitative cases for visual-commonsense-reasoning (VCR) with \vlcm compared to other VLMs in \autoref{fig:qualitative_vcr_2}. The qualitative example demonstrates that the aligned encoders, thought being trained with semantic-level caption, can still capture layout grounding and the spatial reasoning ability.

\end{document}